\documentclass[10pt,twocolumn,letterpaper]{article}

\usepackage[pagenumbers]{cvpr} 

\usepackage{graphicx}
\usepackage{amsmath}
\usepackage{amssymb}
\usepackage{booktabs}
\usepackage{multirow}

\usepackage[pagebackref,breaklinks,colorlinks]{hyperref}

\usepackage[capitalize]{cleveref}
\crefname{section}{Sec.}{Secs.}
\Crefname{section}{Section}{Sections}
\Crefname{table}{Table}{Tables}
\crefname{table}{Tab.}{Tabs.}

\def\cvprPaperID{7405} 
\def\confName{CVPR}
\def\confYear{2023}

\begin{document}

\title{ReDirTrans: Latent-to-Latent Translation for Gaze and Head Redirection}

\author{
    Shiwei Jin $^{1}$,  
    Zhen Wang $^{2}$, 
    Lei Wang $^{2}$, 
    Ning Bi $^{2}$, 
    Truong Nguyen $^{1}$ \\
    $^{1}$ ECE Dept. UC San Diego, $^{2}$ Qualcomm Technologies, Inc. \\
    {\tt\small \{sjin, tqn001\}@eng.ucsd.edu, \{zhewang, wlei, nbi\}@qti.qualcomm.com}
}

\maketitle

\begin{abstract}
Learning-based gaze estimation methods require large amounts of training data with accurate gaze annotations. 
Facing such demanding requirements of gaze data collection and annotation, several image synthesis methods were proposed, which successfully redirected gaze directions precisely given the assigned conditions. 
However, these methods focused on changing gaze directions of the images that only include eyes or restricted ranges of faces with low resolution (less than $128\times128$) to largely reduce interference from other attributes such as hairs, which limits application scenarios. 
To cope with this limitation, we proposed a portable network, called ReDirTrans, achieving latent-to-latent translation for redirecting gaze directions and head orientations in an interpretable manner. 
ReDirTrans projects input latent vectors into aimed-attribute embeddings only and redirects these embeddings with assigned pitch and yaw values. 
Then both the initial and edited embeddings are projected back (deprojected) to the initial latent space as residuals to modify the input latent vectors by subtraction and addition, representing old status removal and new status addition.
The projection of aimed attributes only and subtraction-addition operations for status replacement essentially mitigate impacts on other attributes and the distribution of latent vectors. 
Thus, by combining ReDirTrans with a pretrained fixed e4e-StyleGAN pair, we created ReDirTrans-GAN, which enables accurately redirecting gaze in full-face images with $1024\times1024$ resolution while preserving other attributes such as identity, expression, and hairstyle.
Furthermore, we presented improvements for the downstream learning-based gaze estimation task, using redirected samples as dataset augmentation. 
\end{abstract}
  
\section{Introduction}
\label{sec:introduction}
Gaze is a crucial non-verbal cue that conveys attention and awareness in interactions. 
Its potential applications include mental health assessment \cite{andrist2014conversational, huang2016stressclick}, social attitudes analysis \cite{kaisler2016trusting}, human-computer interaction \cite{feit2017toward}, automotive assistance \cite{schwehr2017driver}, AR/VR \cite{burova2020utilizing, sitzmann2018saliency}. 
However, developing a robust unified learning-based gaze estimation model requires large amounts of data from multiple subjects with precise gaze annotations \cite{zheng2020self, yu2020unsupervised}.
Collecting and annotating such an appropriate dataset is complex and expensive. 
To overcome this challenge, several methods have been proposed to redirect gaze directions \cite{wood2018gazedirector, yu2019improving, he2019photo, zheng2020self, yu2020unsupervised} in real images with assigned directional values to obtain and augment training data.  
Some works focused on generating eye images with new gaze directions by either 1) estimating warping maps \cite{yu2019improving, yu2020unsupervised} to interpolate pixel values or 2) using encoder-generator pairs to generate redirected eye images \cite{wood2018gazedirector, he2019photo}. 

ST-ED \cite{zheng2020self} was the first work to extend high-accuracy gaze redirection from eye images to face images. 
By disentangling several attributes, including person-specific appearance, it can explicitly control gaze directions and head orientations. 
However, due to the design of the encoder-decoder structure and limited ability to maintain appearance features by a $1\times1024$ projected appearance embedding, ST-ED generates low-resolution ($128\times128$) images with restricted face range (no hair area), which narrows the application ranges and scenarios of gaze redirection. 

As for latent space manipulation for face editing tasks, large amounts of works \cite{shen2020interfacegan, tov2021designing, alaluf2021restyle, abdal2021styleflow,  haas2022tensor, alaluf2022hyperstyle} were proposed to modify latent vectors in predefined latent spaces ($W$ \cite{karras2019style}, $W^+$ \cite{abdal2019image2stylegan} and $S$ \cite{wu2021stylespace}).
Latent vectors in these latent spaces can work with StyleGAN \cite{karras2019style, karras2020analyzing} to generate high-quality and high-fidelity face images with desired attribute editing. 
Among these methods, Wu \textit{et.al} \cite{wu2021stylespace} proposed the latent space $S$ working with StyleGAN, which achieved only one degree-of-freedom gaze redirection by modifying a certain channel of latent vectors in $S$ by an uninterpreted value instead of pitch and yaw values of gaze directions. 

Considering these, we proposed a new method, called ReDirTrans, to achieve latent-to-latent translation for redirecting gaze directions and head orientations in high-resolution full-face images based on assigned directional values.
Specifically, we designed a framework to project input latent vectors from a latent space into the aimed-attribute-only embedding space for an interpretable redirection process. 
This embedding space consists of estimated pseudo conditions and embeddings of aimed attributes, where conditions describe deviations from the canonical status and embeddings are the `carriers' of the conditions. 
In this embedding space, all transformations are implemented by rotation matrices multiplication built from pitch and yaw values, which can make the redirection process more interpretable and consistent. 
After the redirection process, the original embeddings and redirected ones are both decoded back to the initial latent space as the residuals to modify the input latent vectors by subtraction and addition operations. 
These operations represent removing the old state and adding a new one, respectively. 
ReDirTrans only focuses on transforming embeddings of aimed attributes and achieves status replacement by the residuals outputted from weight-sharing deprojectors.
ReDirTrans does not project or deproject other attributes with information loss; and it does not affect the distribution of input latent vectors. 
Thus ReDirTrans can also work in a predefined feature space with a fixed pretrained encoder-generator pair for the redirection task in desired-resolution images. 

In summary, our contributions are as follows:
\vspace{-0.2cm}
\begin{itemize}
  \itemsep 0em 
  \item A latent-to-latent framework, \textit{ReDirTrans}, which projects latent vectors to an embedding space for an interpretable redirection process on aimed attributes and maintains other attributes, including appearance, in initial latent space with no information loss caused by projection-deprojection processes. 
  \item A portable framework that can seamlessly integrate into a pretrained GAN inversion pipeline for high-accuracy redirection of gaze directions and head orientations, without the need for any parameter tuning of the encoder-generator pairs.
  \item A layer-wise architecture with learnable parameters that works with the fixed pretrained StyleGAN and achieves redirection tasks in high-resolution full-face images through \textit{ReDirTrans-GAN}. 
\end{itemize}

\section{Related Works}
\label{sec:related_works}

\textbf{Gaze and Head Redirection.}
\label{subsec:redirection}
Methods for redirecting gaze directions can be broadly classified into two categories: warping-based methods and generator-based methods. 
Deepwarp \cite{ganin2016deepwarp, kononenko2017photorealistic} presented a deep network to learn warping maps between pairs of eye images with different gaze directions, which required large amounts of data with annotations. 
Yu \textit{et al.} \cite{yu2019improving} utilized a pretrained gaze estimator and synthetic eye images to reduce the reliance on annotated real data. 
Yu \textit{et al.} \cite{yu2020unsupervised} further extended the warping-based methods in an unsupervised manner by adding a gaze representation learning network.  
As for the generator-based methods, He \textit{et al.} \cite{he2019photo} developed a GAN-based network for generating eye images with new gaze directions. 
FAZE \cite{park2019few} proposed an encoder-decoder architecture to transform eye images into latent vectors for redirection with rotation matrix multiplication, and then decode the edited ones back to the synthetic images with new gaze directions. 
ST-ED \cite{zheng2020self} further extended the encoder-decoder pipeline from gaze redirection only to both head and gaze redirection over full face images by disentangling latent vectors, and achieving precise redirection performance. 
However, ST-ED generates images with a restricted face range (no hair area) with a size of $128\times 128$.
We further improve the redirection task by covering the full face range with $1024\times 1024$ resolution. 

\textbf{Latent Space Manipulation.}
\label{subsec:manipulation}
Numerous methods investigated the latent space working with StyleGAN \cite{karras2019style, karras2020analyzing} to achieve semantic editing in image space due to its meaningful and highly disentangled properties.
As for the supervised methods, InterFaceGAN \cite{shen2020interfacegan} determined hyperplanes for the corresponding facial attribute editing based on provided labels. 
StyleFlow \cite{abdal2021styleflow} proposed mapping a sample from a prior distribution to a latent distribution conditioned on the target attributes estimated by pretrained attribute classifiers. 
Given the unsupervised methods, GANSpace \cite{harkonen2020ganspace}, SeFa \cite{shen2021closed} and TensorGAN \cite{haas2022tensor} leveraged principal components analysis, eigenvector decomposition and higher-order singular value decomposition to discover semantic directions in latent space, respectively. 
Other self-supervised methods proposed mixing of latent codes from other samples for local editing \cite{collins2020editing, chong2021retrieve}, or incorporating the language model CLIP \cite{radford2021learning} for text-driven editing \cite{patashnik2021styleclip}. 

\textbf{Domain Adaptation for Gaze Estimation.}
\label{subsec:adaptation}
Domain gaps among different datasets restrict the application range of pretrained gaze estimation models. 
To narrow the gaps, a few domain adaptation approaches \cite{wood20163d, wood2016learning} were proposed for the generic regression task.
SimGAN \cite{shrivastava2017learning} proposed an unsupervised domain adaptation method for narrowing the gaps between real and synthetic eye images. 
HGM \cite{wang2018hierarchical} designed a unified 3D eyeball model for eye image synthesis and cross-dataset gaze estimation.
PnP-GA \cite{liu2021generalizing} presented a gaze adaptation framework for generalizing gaze estimation in new domains based on collaborative learning. 
Qin \textit{et al.} \cite{qin2022learning} utilized 3D face reconstruction to rotate head orientations together with changed eye gaze accordingly to enlarge overlapping gaze distributions among datasets. 
These adaptation methods typically rely on restricted face or eye images to alleviate interference from untargeted attributes. 
Our work incorporates the redirection task in a predefined meaningful feature space with controllable attributes to achieve high-resolution and full-face redirection. 
\begin{figure*}  
\begin{center}  
    \includegraphics[width=0.77\linewidth]{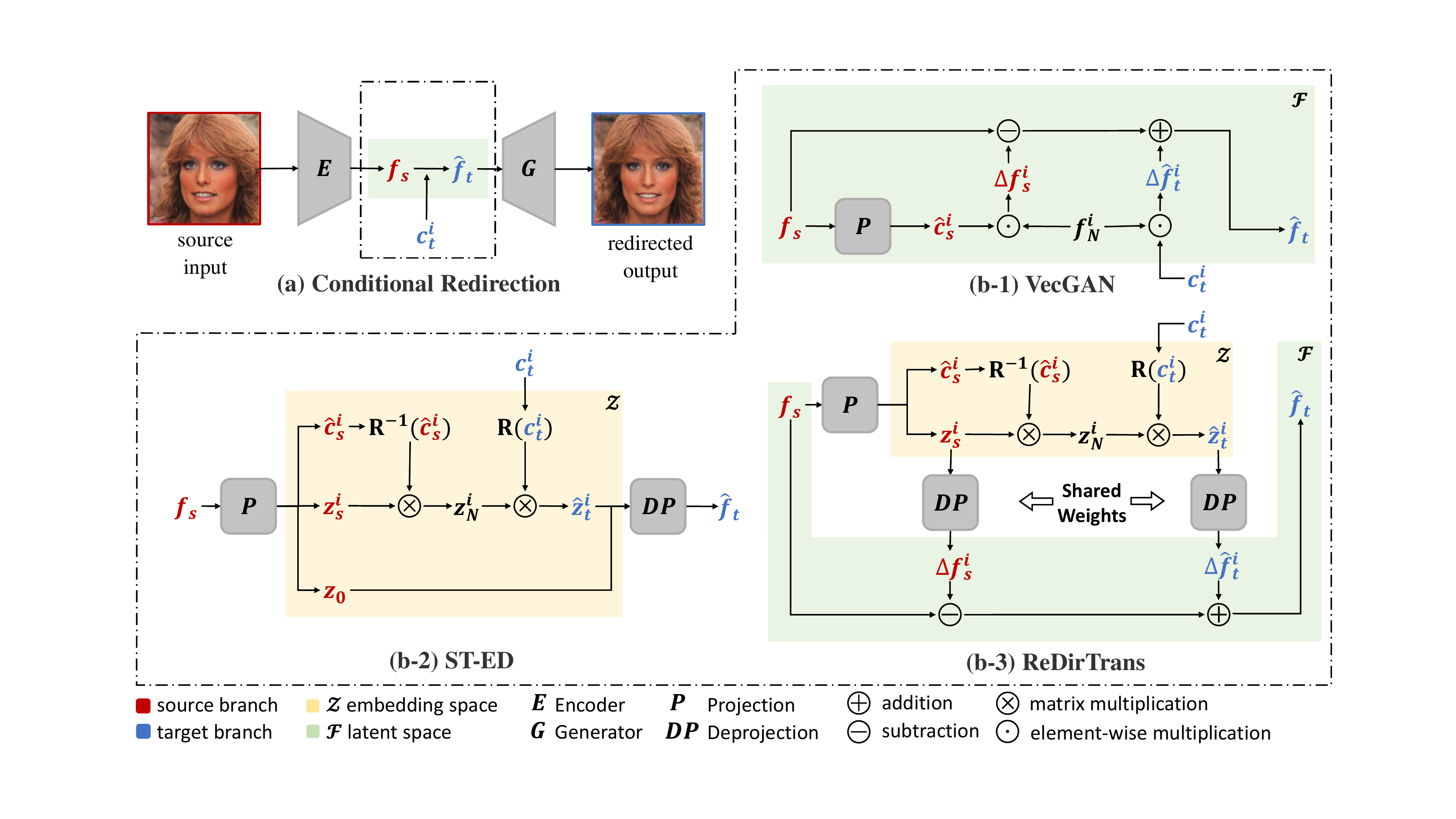}
    \caption{Conditional redirection pipeline and comparison among different redirectors. (a) We first encode the input image into the latent vector. Given the provided conditions, we modify the latent vector and send it to a generator for image synthesis with only aimed attribute redirection. (b) We compared our proposed redirector with two state-of-the-art methods by omitting common modules (basic encoders and decoders/generators) and focusing on the unique components: (b-1) VecGAN achieves editing in feature space $\mathcal{F}$ given projected conditions from latent vectors with a global direction $f^i_N$. (b-2) ST-ED projects the latent vector into conditions and embeddings of aimed attributes, and one appearance-related high dimensional embedding $z_0$ in embedding space $\mathcal{Z}$. After interpretable redirection process in space $\mathcal{Z}$, all embeddings are concatenated and projected back to space $\mathcal{F}$. (b-3) Our proposed ReDirTrans projects the latent vector into conditions and embeddings of aimed attributes only. After an interpretable redirection process, both original and redirected embeddings are deprojected back to initial space $\mathcal{F}$ as residuals. These residuals modify the input latent vector by subtraction and addition operations, which represent the initial status removal and the new status addition, respectively. This approach efficiently reduces effects on other attributes (especially the appearance related information) with fewer parameters than ST-ED.} 
    \label{fig_redirectors}
\end{center} 
\end{figure*}

\section{Method}
\label{sec:method}
\subsection{Problem Statements}
\label{subsec:problem_statements}
Our goal is to train a conditional latent-to-latent translation module for face editing with physical meaning, and it can work either with a trainable or fixed encoder-generator pair. 
This editing module first transforms input latent vectors from encoded feature space $\mathcal{F}$ to an embedding space $\mathcal{Z}$ for redirection in an interpretable manner. 
Then it deprojects the original and redirected embeddings back to the initial feature space $\mathcal{F}$ for editing input latent vectors. 
The edited latent vectors are fed into a generator for image synthesis with the desired status of aimed facial attributes. 
The previous GAN-based work \cite{shen2020interfacegan, haas2022tensor, dalva2022vecgan} achieved a certain facial attribute editing with a global latent residual multiplied by a scalar without physical meaning to describe the relative deviation from the original status.
To make the whole process interpretable and achieve redirection directly based on the new gaze directions or head poses, we follow the assumption proposed by \cite{zheng2020self}, where the dimension of an embedding is decided by the corresponding attribute's degree of the freedom (DoF) and redirection process is achieved by the rotation matrices multiplication.
Thus the transformation equivariant mappings can be achieved between the embedding space $\mathcal{Z}$ and image space. 
To be specific, normalized gazes or head poses can be represented by a two-dimensional embedding with the pitch and yaw as the controllable conditions. 
The embeddings can be edited (multiplied) by the rotation matrices built from the pitch and yaw for achieving redirection (rotation) of aimed attributes in image space accordingly through our proposed redirector. 

\begin{figure*}  
\begin{center}  
    \includegraphics[width=0.77\linewidth]{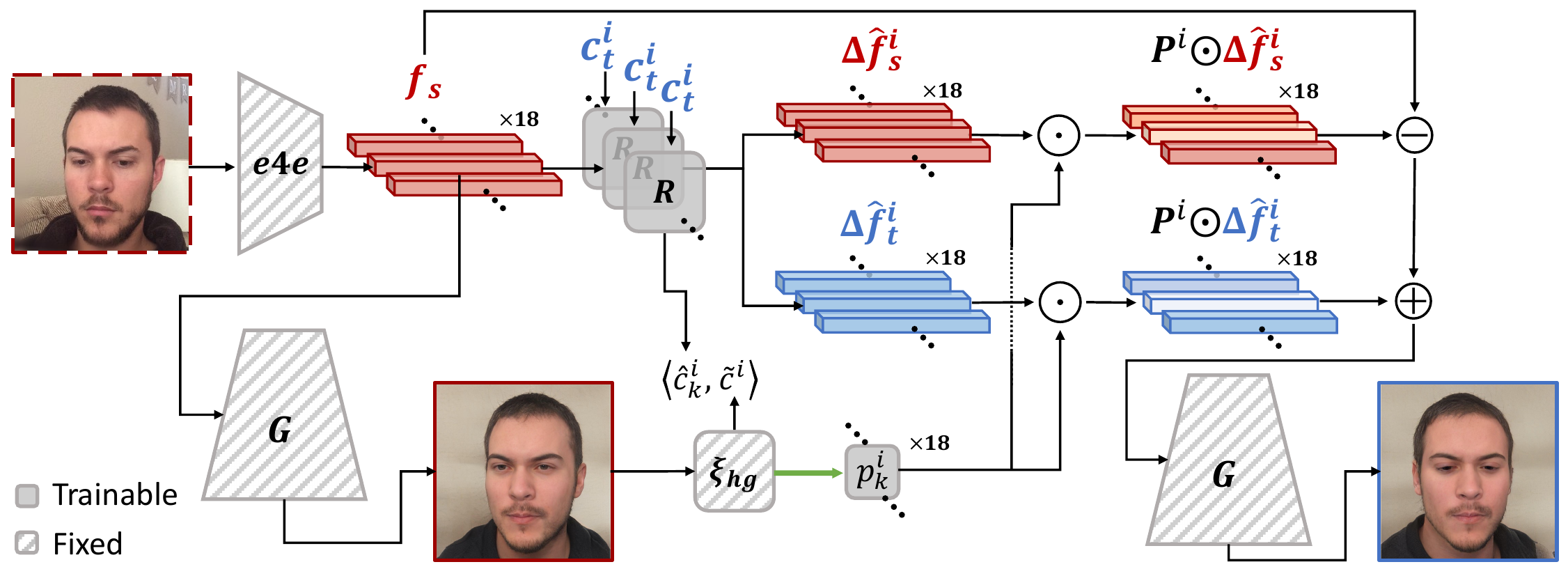}
    \caption{ReDirTrans-GAN: Layer-wise ReDirTrans with fixed e4e and StyleGAN. Given the multi-layer representation of latent vectors in $W^+ \subseteq \mathbb{R}^{18\times512}$, we feed each layer into an independent ReDirTrans for redirection task given the provided target condition $c_t^i$, where $i$ represents a certain attribute. $\Tilde{c}^i$ denotes the pseudo condition estimated directly from the inverted image by a pretrained HeadGazeNet $\xi_{hg}$ \cite{zheng2020self}. We calculate errors between estimated conditions $\hat c^i_k, k\in[1,18]$ from multiple ReDirTrans and $\Tilde{c}^i$ for supervising the trainable weights learning (green arrow) based on the Layer-wise Weights Loss described in Eq. \ref{eq:loss_7} to decide which layers should contribute more to a certain attribute redirection. Given the estimated weights and initial latent vectors $f_s$, we can acquire the final disentangled latent vector $\hat f_{t, d}$ based on Eq. \ref{eq:33} for redirected samples synthesis.} 
    \label{fig_redir+GAN}
\end{center} 
\end{figure*}

\subsection{Redirector Architecture}
\label{subsec:redirect_arch}
ST-ED is one of the state-of-the-art architectures for gaze and head poses redirection over face images \cite{zheng2020self} shown in Fig. \ref{fig_redirectors} (b-2).
ST-ED projects the input latent vector $f$ to non-varying embeddings $z^0$ and $M$ varying ones with corresponding estimated conditions $\{(z^i, \hat{c}^i)|i\in[1, M]\}$, where $\hat{c}^i$ describes the estimated amount of deviation from the canonical status of the attribute $i$, and it can be compared with the ground truth $c^i$ for the learning of conditions from latent vectors. 
The non-varying embedding $z^0$ defines subject's appearance, whose dimension is much larger (over twenty times larger in ST-ED) than other varying embeddings.
It is inefficient to project input latent vectors into a high-dimensional embedding to maintain non-varying information such as identity, hairstyle, etc. 
Thus, we propose a new redirector architecture, called \textit{ReDirTrans}, shown in Fig. \ref{fig_redirectors} (b-3), which transforms the source latent vector $f_{s}$ to the embeddings of aimed attributes through the projector \textbf{\textit{P}} and redirects them given the newly provided target conditions $c_{t}$. 
Then we deproject both original embeddings $\mathbf{z_{s}}$ and redirected embeddings $\mathbf{\hat z_{t}}$ back to the feature space $\mathcal{F}$ through the weights-sharing deprojectors \textbf{\textit{DP}} to acquire latent residuals.
These residuals contain source and target status of aimed attributes, denoted as $\Delta f^i_{s}$ and $\Delta \hat f^i_{t}$, respectively. 
Inspired by addition and subtraction \cite{shen2020interfacegan, dalva2022vecgan} for face editing in feature space $\mathcal{F}$, the edited latent vector is
\begin{equation}
\label{eq:1}
    \hat f_{t} = f_s + \sum_{i=1}^M(- \Delta f^i_{s} + \Delta \hat f^i_{t}), i \in [1, M],
\end{equation}
where the subtraction means removing source status and the addition indicates bringing in new status. 
The projector \textbf{\textit{P}} ensures that the dimension of embeddings can be customized based on the degrees of freedom of desired attributes, and the transformations can be interpretable with physical meanings. 
The deprojector \textbf{\textit{DP}} enables the original and edited features in the same feature space, allowing ReDirTrans to be compatible with pretrained encoder-generator pairs that are typically trained together without intermediate (editing) modules. 
ReDirTrans reduces parameters by skipping projection (compression) and deprojection (decompression) of the features that are not relevant to the desired attributes, but vital for final image synthesis. 

\subsection{Predefined Feature Space}
\label{subsec:predefined_space}
Except for the \textit{trainable} encoder-decoder (or -generator) pair to learn a specific feature space for redirection task as ST-ED did, ReDirTrans can also work in the predefined feature space to coordinate with \textit{fixed}, \textit{pretrained} encoder-generator pairs. 
For our implementation, we chose the $W^+ \in \mathbb{R}^{18\times512}$ feature space \cite{abdal2019image2stylegan}, which allows us to utilize StyleGAN \cite{karras2020analyzing} for generating high-quality, high-fidelity face images.
We refer to this implementation as \textit{ReDirTrans-GAN}.  
Considering multi-layer representation of the latent vector \cite{abdal2019image2stylegan} and its semantic disentangled property between different layers \cite{abdal2021styleflow, harkonen2020ganspace} in $W^+$ space, we proposed layer-wise redirectors, shown in Fig. \ref{fig_redir+GAN}, rather than using a single ReDirTrans to process all ($18$) layers of the latent vector. 
To largely reduce the interference between different layers during redirection, we assume that if one attribute's condition can be estimated from certain layers with less errors than the others, then we can `modify' these certain layers with higher weights $p_k^i, k\in[1, 18]$ than others to achieve redirection of the corresponding attribute $i$ only.
$\boldsymbol{P}^i=[p^i_1, \cdots, p^i_{18}]^T \in \mathbb{R}^{18\times 1}$, as part of network parameters, is trained given the loss function described in Eq. \ref{eq:loss_7}. 
The final disentangled latent vectors after redirection is 
\begin{equation}
\label{eq:33}
\hat f_{t, d} = f_s + \sum_{i=1}^M \boldsymbol{P}^i \odot (- \Delta f^i_{s} + \Delta \hat f^i_{t}), i \in [1, M],
\end{equation}
where $\odot$ means element-wise multiplication and $(- \Delta f^i_{s} + \Delta \hat f^i_{t})\in \mathbb{R}^{18\times 512}$. 
One \textbf{challenge} regarding the predefined feature space comes from the inversion quality. 
There exist attribute differences between input images and inverted results, shown in Fig. \ref{fig_red_comparison} and \ref{fig_rec}, which means that the conditions in source images cannot be estimated from source latent vectors. 
To solve this, instead of using conditions from source images, we utilized estimated conditions from the inverted images, which ensures the correctness and consistence of conditions learning from latent vectors.

\subsection{Training Pipeline}
\label{subsec:train_pipeline}
Given a pair of source and target face images, $I_{s}$ and $I_{t}$ from the same person, we utilize an encoder to first transform $I_{s}$ into the feature space $\mathcal{F}$, denoted as $f_{s}$. 
We further disentangle $f_{s}$ into the gaze-direction-related embedding $z_{s}^1$ and the head-orientation-related embedding $z_{s}^2$ with corresponding estimated conditions: $\hat{c}_{s}^1$ and $\hat{c}_{s}^2$ by the projector \textbf{\textit{P}}.
Then we build rotation matrices using the pitch and yaw from estimated conditions $(\hat{c}_{s}^1, \hat{c}_{s}^2)$ and target conditions $(c_{t}^1, c_{t}^2)$ to normalize embeddings and redirect them to the new status, respectively:  
\begin{equation}
\label{eq:2}
\begin{aligned}
    \text{Normalization: } z_{N}^{i} &= \mathbf{R}^{-1}(\hat{c}^{i}_{s})\cdot z_{s}^{i}, \\
    \text{Redirection: \ \ \ \ \ \ } \hat{z}_{t}^{i} & = \mathbf{R}(c_{t}^{i})\cdot z_{N}^{i},  
\end{aligned}
\end{equation}
where $i\in \{1, 2\}$, representing gaze directions and head orientations, respectively, and $z_{N}^{i}$ denotes the normalized embedding of the corresponding attribute. 
We feed the original embedding $z_{s}^{i}$ and the modified embedding $\hat{z}_{t}^{i}$ into the weights-sharing deprojectors \textbf{\textit{DP}} to transform these embeddings back to the feature space $\mathcal{F}$ as the residuals. 
Given these residuals, we implement subtraction and addition operations over $f_{s}$ as described in Eq. \ref{eq:1} (or Eq. \ref{eq:33}) to acquire the edited latent vector $\hat f_{t}$ (or $\hat f_{t, d}$), which is sent to a generator for synthesizing redirected face image $\hat I_{t}$. $\hat I_{t}$ should have the same gaze direction and head orientation as $I_{t}$.  

\subsection{Learning Objectives}
\label{subsec:learning_objectives}
We supervise the relationship between the generated image $\hat I_{t}$ and the target image $I_{t}$ with several loss functions: pixel-wise reconstruction loss, LPIPS metric \cite{zhang2018unreasonable} and attributes loss by a task-related pretrained model. 
\begin{equation}
\label{eq:loss_1}
    \mathcal{L}_{rec}(\hat I_{t}, I_{t}) = \big|\big|\hat I_{t} - I_{t}\big|\big|_2, 
\end{equation}
\begin{equation}
\label{eq:loss_2}
    \mathcal{L}_{LPIPS}(\hat I_{t}, I_{t}) = \big|\big|\psi(\hat I_{t}) - \psi(I_{t})\big|\big|_2, 
\end{equation}
\begin{equation}
\label{eq:loss_6}
    \mathcal{L}_{att}(\hat I_{t}, I_{t}) = \langle \xi_{hg}(\hat I_{t}), \xi_{hg}(I_{t})\rangle,  
\end{equation}
where $\psi(\cdot)$ denotes the perceptual feature extractor \cite{zhang2018unreasonable}, $\xi_{hg}(\cdot)$ represents the pretrained HeadGazeNet \cite{zheng2020self} to estimate the gaze and head pose from images and $\langle u, v\rangle = \arccos\frac{u \cdot v}{||u||\cdot||v||}$. 



\vspace{-1em}
\paragraph{Identity Loss.}
Identity preservation after redirection is critical for the face editing task. 
Considering this, we calculate the cosine similarity of the identity-related features between the source image and the redirected image: 
\begin{equation}
\label{eq:loss_3}
    \mathcal{L}_{ID}(\hat I_{t}, I_{s}) = 1 - \langle \phi(\hat I_{t}), \phi(I_{s})\rangle, 
\end{equation}
where $\phi(\cdot)$ denotes the pretrained ArcFace \cite{deng2019arcface} model. 


\vspace{-1em}
\paragraph{Label Loss.}
We have ground truth of gaze directions and head orientations, which can guide the conditions learning from the input latent vectors for the normalization step:  
\begin{equation}
\label{eq:loss_4}
    \mathcal{L}_{lab}(\hat c^{i}_{s},  c^{i}_{s}) = \langle \hat c^{i}_{s}, c^{i}_{s}\rangle, \ \ i \in \{1, 2\}. 
\end{equation}

\vspace{-1.5em}
\paragraph{Embedding Loss.}
The normalized embeddings only contain the canonical status of the corresponding attribute after the inverse rotation applied to the original estimated embeddings, shown in Fig. \ref{fig_redirectors}. 
Thus the normalized embeddings given a certain attribute across different samples within batch $B$ should be consistent. 
To reduce the number of possible pairs within a batch, we utilize the first normalized embedding $z_{N, 1}^{i}$ as the basis: 
\begin{equation}
\label{eq:loss_5}
    \mathcal{L}_{emb} = \frac1{B-1} \sum^B_{j=2} \langle z_{N, 1}^{i}, z_{N, j}^{i} \rangle,  \ \ i \in \{1, 2\}. 
\end{equation}



\vspace{-1.5em}
\paragraph{Layer-wise Weights Loss.}
This loss is specifically designed for the $W^+$ space to decide the weights $p_i$ of which layer should contribute more to the aimed attributes editing. 
Firstly, we calculate the layer-wise estimated conditions $\hat c^i_k$ and calculate estimated pseudo labels $\Tilde{c}^i$. 
Secondly, we have layer-wise estimated label errors by $\langle \hat c^i_k, \Tilde{c}^i\rangle$.
Lastly, we calculate the cosine similarity between the reciprocal of label errors and weights of layers as the loss:
\begin{equation}
\label{eq:loss_7}
    \mathcal{L}_{prob} = \langle 
    \{p_k\}, \{\frac1{\langle \hat c^i_k, \Tilde{c}^i\rangle}\}
    \rangle, k\in[1, K], i \in \{1, 2\},  
\end{equation}
where $K$ is the number of layers for editing. 

\vspace{-1em}
\paragraph{Full Loss.}
The combined loss function for supervising the redirection process is: 
\begin{equation}
\label{eq:loss_9}
\begin{aligned}
    \mathcal{L} = \lambda_{r}\mathcal{L}_{rec} &+ \lambda_{L}\mathcal{L}_{LPIPS} + \lambda_{ID}\mathcal{L}_{ID} + \lambda_{a}\mathcal{L}_{att} \\ + \lambda_{l}\mathcal{L}_{lab} &+ 
    \lambda_{e}\mathcal{L}_{emb} + 
    \lambda_{p}\mathcal{L}_{prob},  
\end{aligned}
\end{equation}
where $\mathcal{L}_{LPIPS}$ and $\mathcal{L}_{prob}$ are utilized only when the pretrained StyleGAN is used as the generator. 
\section{Experiments}
\label{sec:exps}

\begin{table}[t]
\begin{tabular}{@{}cccccc@{}}
\toprule
 &
  \begin{tabular}[c]{@{}c@{}}Gaze\\ Redir\end{tabular} &
  \begin{tabular}[c]{@{}c@{}}Head \\ Redir\end{tabular} &
  \begin{tabular}[c]{@{}c@{}}Gaze \\ Induce\end{tabular} &
  \begin{tabular}[c]{@{}c@{}}Head \\ Induce\end{tabular} &
  LPIPS \\ \midrule
StarGAN $^\dagger$ \cite{choi2018stargan}        & 4.602          & 3.989          & 0.755          & 3.067          & 0.257          \\
He \textit{et al.} $^\dagger$ \cite{he2019photo} & 4.617          & 1.392          & 0.560          & 3.925          & 0.223          \\
VecGAN \cite{dalva2022vecgan}         & 2.282          & 0.824          & 0.401          & 2.205          & \textbf{0.197} \\
ST-ED \cite{zheng2020self}            & 2.385          & 0.800          & \textbf{0.384} & 2.187          & 0.208          \\ \midrule
ReDirTrans                            & \textbf{2.163} & \textbf{0.753} & 0.429          & \textbf{2.155} & \textbf{0.197} \\ \bottomrule
\end{tabular}
\caption{Within-dataset quantitative comparison (GazeCapture test subset) between different methods for redirecting head orientations and gaze directions. (Lower is better). \textbf{Head (Gaze) Redir} denotes the redirection accuracy in degree between the redirected image and the target image given head orientations (gaze directions). \textbf{Head (Gaze) Induce} denotes the errors in degree on gaze (head) when we redirect the head (gaze). $^\dagger$ denotes copied results from \cite{zheng2020self}. Other methods are retrained given previous papers.}
\label{tab_1}
\end{table}

\subsection{datasets}
\label{subsec:datasets}
We utilize GazeCapture \cite{krafka2016eye} training subset to train the redirector and assess the performance with its test subset, MPIIFaceGaze and CelebA-HQ \cite{karras2017progressive}. 
Supplementary material provides more information about these datasets. 


\subsection{Evaluation Criteria}
\label{subsec:criteria}
We follow metrics utilized by ST-ED \cite{zheng2020self} to evaluate different redirectors' performance.

\textbf{Redirection Error.} We measure the redirection accuracy in image space by a pre-trained ResNet-50 based \cite{he2016deep} head pose and gaze estimator $\xi'_{hg}$, which is unseen during the training. 
Given the target image $I_{t}$ and the generated one $\hat I_{t}$ redirected by conditions of $I_{t}$, we report the angular error between $\xi'_{hg}(\hat I_{t})$ and $\xi'_{hg}(I_{t})$ as the redirection errors. 

\textbf{Disentanglement Error.} We quantify the disentanglement error by the condition's fluctuation range of one attribute when we redirect the other one. 
The redirection angle $\epsilon$ follows $\mathcal{U}(-0.1\pi, 0.1\pi)$. 
For example, when we redirect the head pose of the generated image $\hat I_{t}$ by $\epsilon$ and generate a new one $\hat I'_{t}$, we calculate the angular error of the estimated gaze directions between $\xi'_{hg}(\hat I_{t})$ and $\xi'_{hg}(I'_{t})$.

\textbf{LPIPS.} LPIPS is able to measure the distortion \cite{zhang2018unreasonable} and image similarity in gaze directions \cite{he2019photo} between images, which is applied to evaluate the redirection performance. 


\subsection{Redirectors in Learnable Latent Space}
\label{subsec:results}
We compared quantitative performance of different redirectors, which were trained along with the trainable encoder-decoder pair designed by ST-ED on $128\times128$ images with restricted face ranges, given the criteria proposed in Sec. \ref{subsec:criteria}.
Table \ref{tab_1} and Table \ref{tab_2} present within-dataset and cross-dataset performance, respectively. 
From these tables, we observe that our proposed ReDirTrans achieved more accurate redirection and better LPIPS compared with other state-of-the-art methods by considering the extra embedding space $\mathcal{Z}$ for redirecting embeddings of aimed attributes only and maintaining other attributes including the appearance-related information in the original latent space $\mathcal{F}$. 
ST-ED \cite{zheng2020self} projected input latent vectors into nine embeddings including the non-varying embedding $z^0$. 
This appearance-related high dimensional embedding $z^0$ requires more parameters than ReDirTrans during projection.
After redirecting the embeddings of aimed attributes, ST-ED deprojected a stack of $z^0$, redirected embeddings, and rest unvaried embeddings of other attributes back to the feature space for decoding. 
\begin{table}[t]
\begin{tabular}{@{}cccccc@{}}
\toprule
 &
  \begin{tabular}[c]{@{}c@{}}Gaze\\ Redir\end{tabular} &
  \begin{tabular}[c]{@{}c@{}}Head \\ Redir\end{tabular} &
  \begin{tabular}[c]{@{}c@{}}Gaze \\ Induce\end{tabular} &
  \begin{tabular}[c]{@{}c@{}}Head \\ Induce\end{tabular} &
  LPIPS \\ \midrule
StarGAN $^\dagger$ \cite{choi2018stargan}        & 4.488          & 3.031          & 0.786          & 2.783          & 0.260          \\
He \textit{et al.} $^\dagger$ \cite{he2019photo} & 5.092	       & 1.372	        & 0.684	         & 3.411	      & 0.241          \\
VecGAN \cite{dalva2022vecgan}         & 2.670          & 1.242          & 0.391          & 1.941          & 0.207          \\
ST-ED \cite{zheng2020self}            & \textbf{2.380} & 1.085          & \textbf{0.371} & \textbf{1.782}  & 0.212          \\ \midrule
ReDirTrans                            & \textbf{2.380} & \textbf{0.985} & 0.391          & \textbf{1.782} & \textbf{0.202} \\ \bottomrule
\end{tabular}
\caption{Cross-dataset quantitative comparison (MPIIFaceGaze) between different methods for redirecting head orientations and gaze directions. (Lower is better). Notations are the same as them in the Table \ref{tab_1}. $^\dagger$ denotes copied results from \cite{zheng2020self}. Other methods are retrained given previous papers.}
\label{tab_2}
\end{table}
\begin{figure}[t]
    \centering
    \begin{subfigure}[t]{0.25\linewidth}
        \begin{minipage}[b]{1\linewidth}
        \includegraphics[width=1\linewidth]{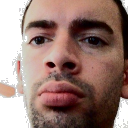}
        \includegraphics[width=1\linewidth]{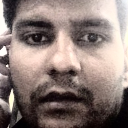}
        \end{minipage}
    \caption{Input}
    \label{fig:input1}
  \end{subfigure}
  \hspace{-0.021\linewidth}
  \centering
    \begin{subfigure}[t]{0.25\linewidth}
        \begin{minipage}[b]{1\linewidth}
        \includegraphics[width=1\linewidth]{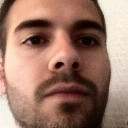}
        \includegraphics[width=1\linewidth]{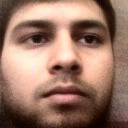}
        \end{minipage}
    \caption{ST-ED}
    \label{fig:e4e_inv1}
  \end{subfigure}
  \hspace{-0.021\linewidth}
  \centering
    \begin{subfigure}[t]{0.25\linewidth}
        \begin{minipage}[b]{1\linewidth}
        \includegraphics[width=1\linewidth]{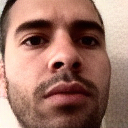}
        \includegraphics[width=1\linewidth]{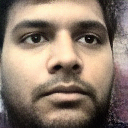}
        \end{minipage}
    \caption{ReDirTrans}
    \label{fig:restyle1}
  \end{subfigure}
  \hspace{-0.021\linewidth}
  \begin{subfigure}[t]{0.25\linewidth}
        \begin{minipage}[b]{1\linewidth}
        \includegraphics[width=1\linewidth]{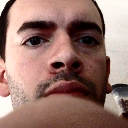}
        \includegraphics[width=1\linewidth]{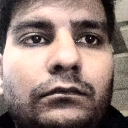}
        \end{minipage}
    \caption{Target}
    \label{fig:input2}
  \end{subfigure}
  \vspace{-0.5em}
  \caption{Qualitative Comparison of ReDirTrans and ST-ED in GazeCapture. ReDirTrans preserves more facial attributes, such as lip thickness and sharpness of the beard.}
  \label{fig_reb_comp}
\end{figure}
This projection-deprojection process of non-varying embedding $z^0$ results in loss of appearance and worse LPIPS, as depicted in Fig. \ref{fig_reb_comp}.   
VecGAN \cite{dalva2022vecgan} was proposed to edit the attributes only within the feature space by addition and subtraction operations. 
Since there is no projection-deprojection process, given the original latent code, LPIPS performance is better than ST-ED. 
However, as no extra embedding space was built for the aimed attributes editing, both redirection accuracy and the disentanglement process were affected. 

\begin{figure*}
    \captionsetup[subfigure]{labelformat=empty}
    \captionsetup[subfigure]{justification=centering}
    \centering
        \begin{minipage}{0.3cm}
        \rotatebox{90}{\scriptsize{ReDirTrans-GAN~~~~~~~~~~~ST-ED~~~~~~~~~ReDirTrans-GAN~~~~~~~~~~ST-ED}}
        \end{minipage}%
    \begin{subfigure}[t]{0.124\linewidth}
        \begin{minipage}{1\linewidth}
        \includegraphics[width=1\linewidth]{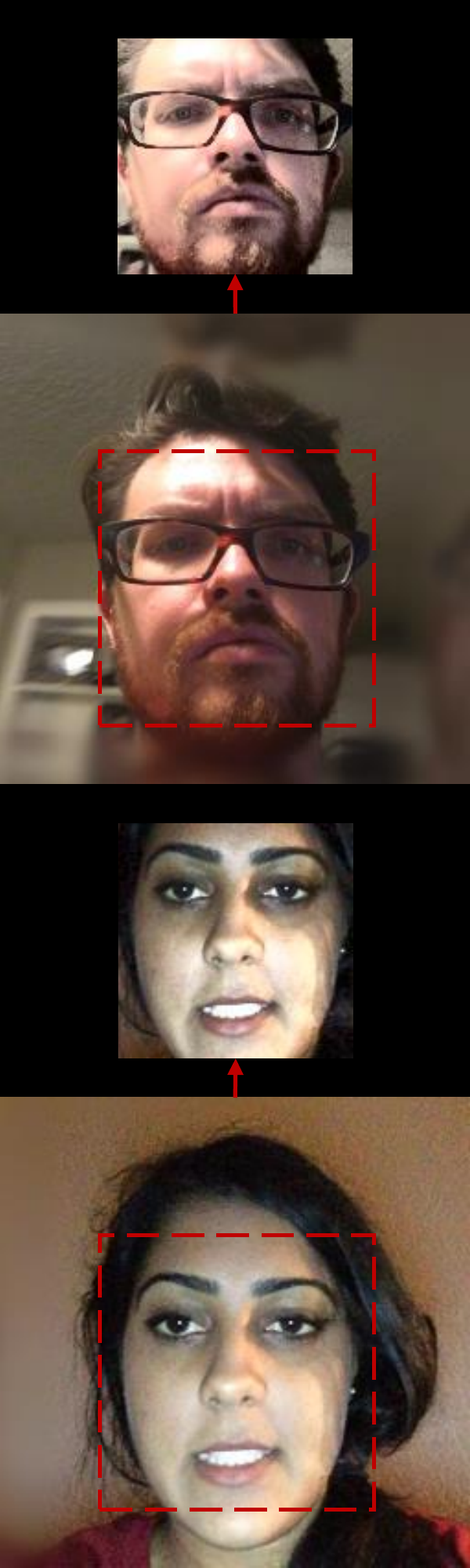}
        \end{minipage}
    \caption{Input}
    \label{fig:comp_input}
  \end{subfigure}
  \hspace{-0.015\linewidth}
  \centering
    \begin{subfigure}[t]{0.124\linewidth}
        \begin{minipage}{1\linewidth}
        \includegraphics[width=1\linewidth]{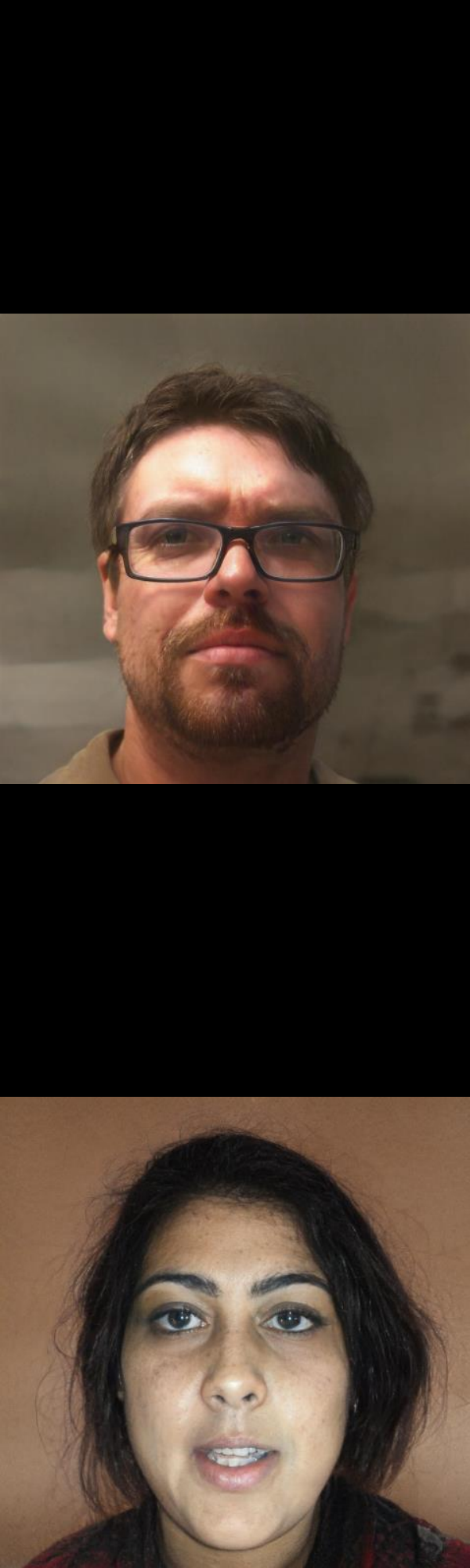}
        \end{minipage}
    \caption{e4e \\ Inversion}
    \label{fig:comp_inver}
  \end{subfigure}
  \hspace{-0.015\linewidth}
  \centering
    \begin{subfigure}[t]{0.124\linewidth}
        \begin{minipage}{1\linewidth}
        \includegraphics[width=1\linewidth]{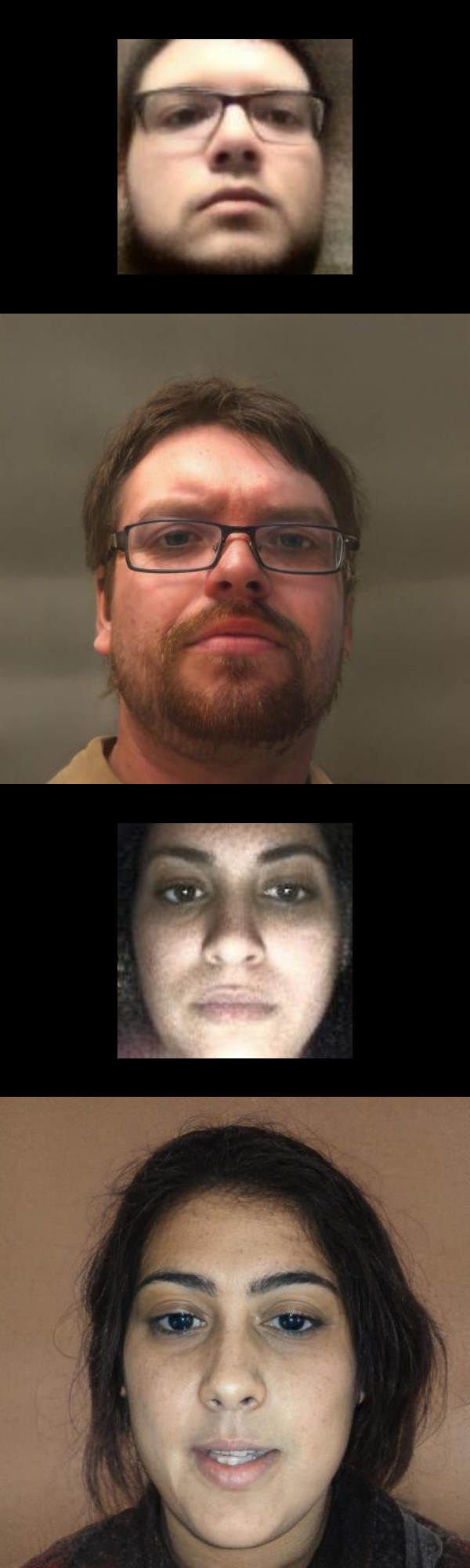}
        \end{minipage}
    \caption{Redirection}
    \label{fig:comp_redir}
  \end{subfigure}
  \hspace{-0.015\linewidth}
  \centering
    \begin{subfigure}[t]{0.124\linewidth}
        \begin{minipage}{1\linewidth}
        \includegraphics[width=1\linewidth]{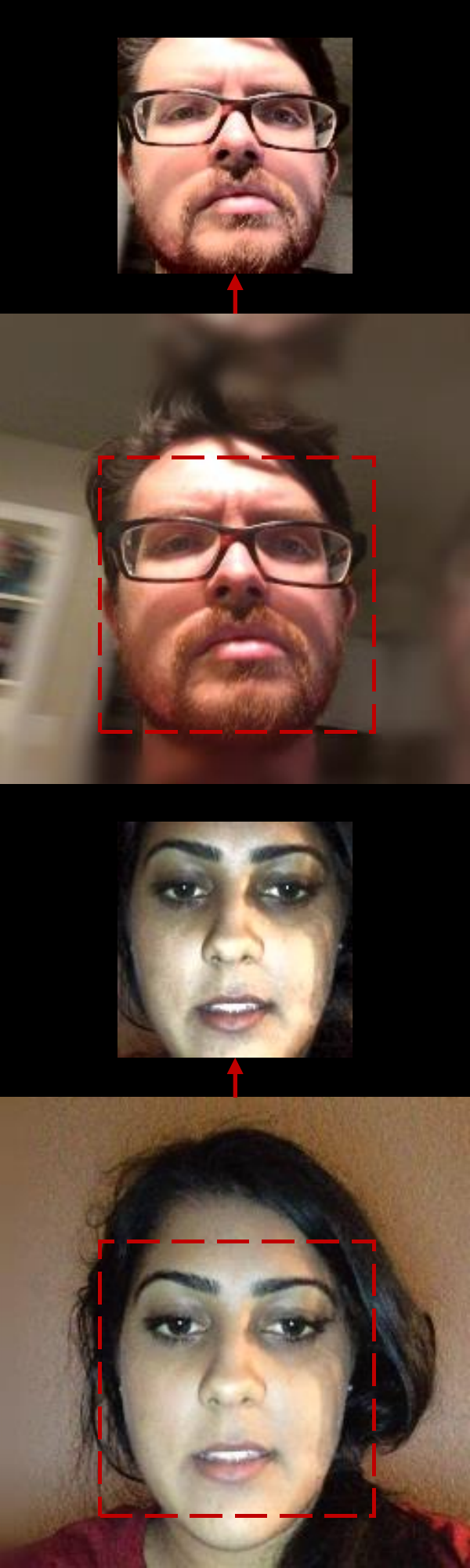}
        \end{minipage}
    \caption{Target}
    \label{fig:comp_tar}
  \end{subfigure}
  \hspace{-0.008\linewidth}
  \centering
    \begin{subfigure}[t]{0.124\linewidth}
        \begin{minipage}{1\linewidth}
        \includegraphics[width=1\linewidth]{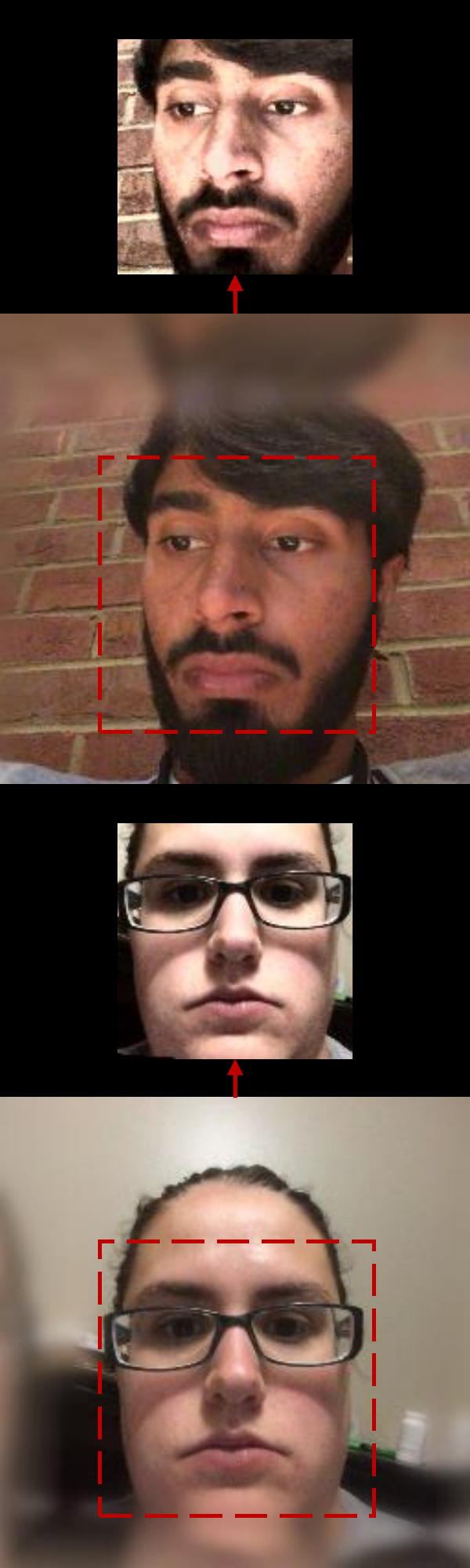}
        \end{minipage}
    \caption{Input}
    \label{fig:comp_input1}
  \end{subfigure}
  \hspace{-0.015\linewidth}
  \centering
    \begin{subfigure}[t]{0.124\linewidth}
        \begin{minipage}{1\linewidth}
        \includegraphics[width=1\linewidth]{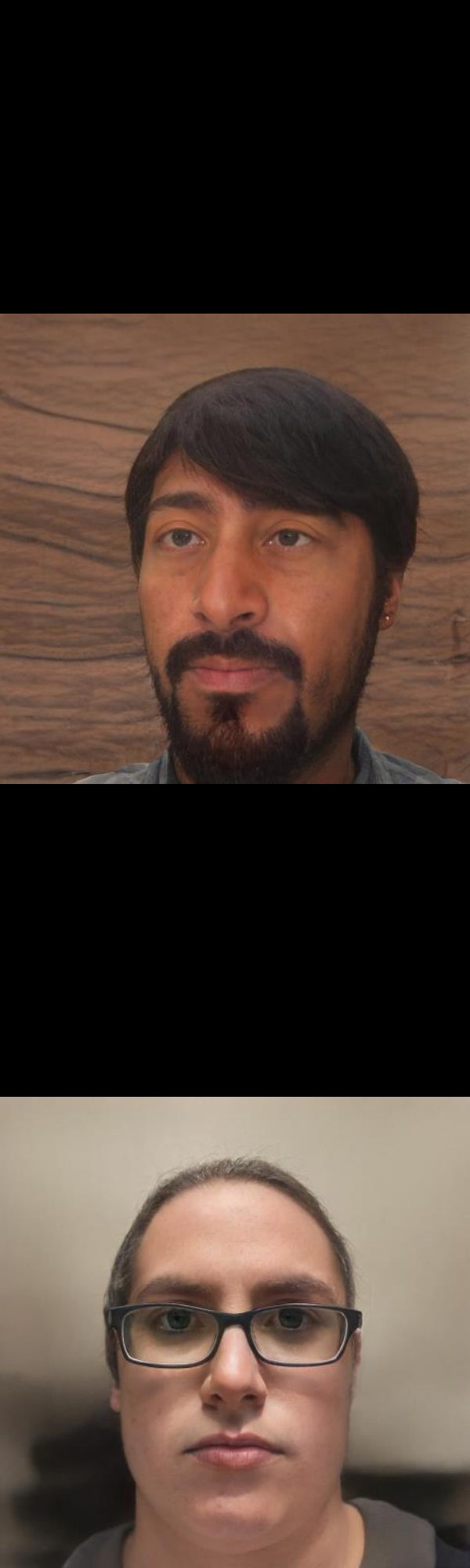}
        \end{minipage}
    \caption{e4e \\ Inversion}
    \label{fig:comp_inver1}
  \end{subfigure}
  \hspace{-0.015\linewidth}
  \centering
    \begin{subfigure}[t]{0.124\linewidth}
        \begin{minipage}{1\linewidth}
        \includegraphics[width=1\linewidth]{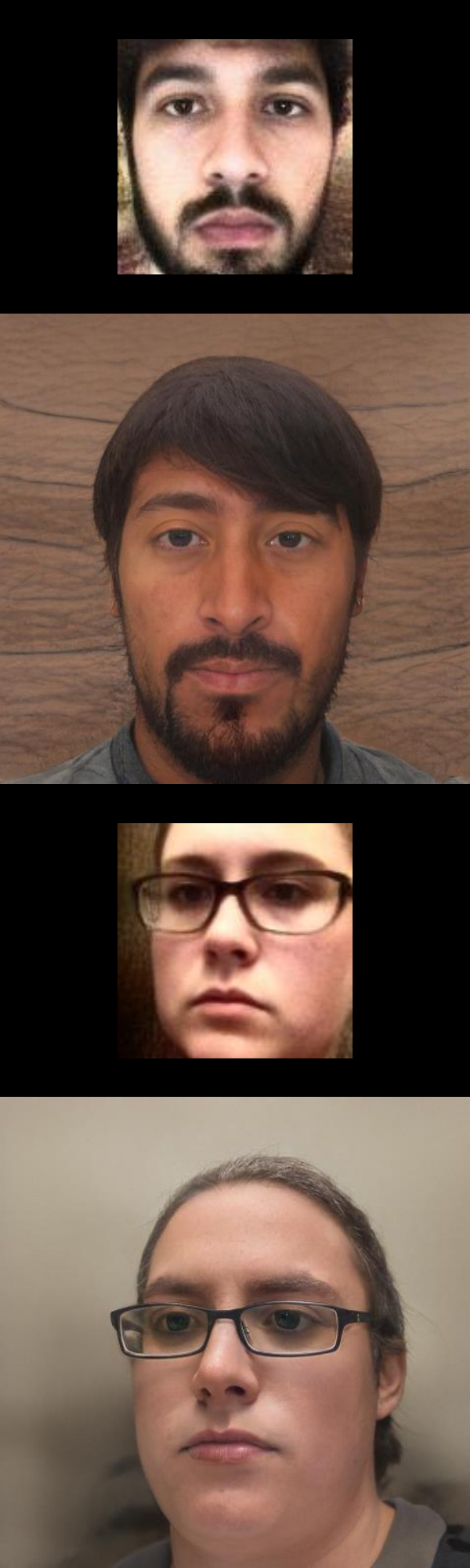}
        \end{minipage}
    \caption{Redirection}
    \label{fig:comp_redir1}
  \end{subfigure}
  \hspace{-0.015\linewidth}
  \centering
    \begin{subfigure}[t]{0.124\linewidth}
        \begin{minipage}{1\linewidth}
        \includegraphics[width=1\linewidth]{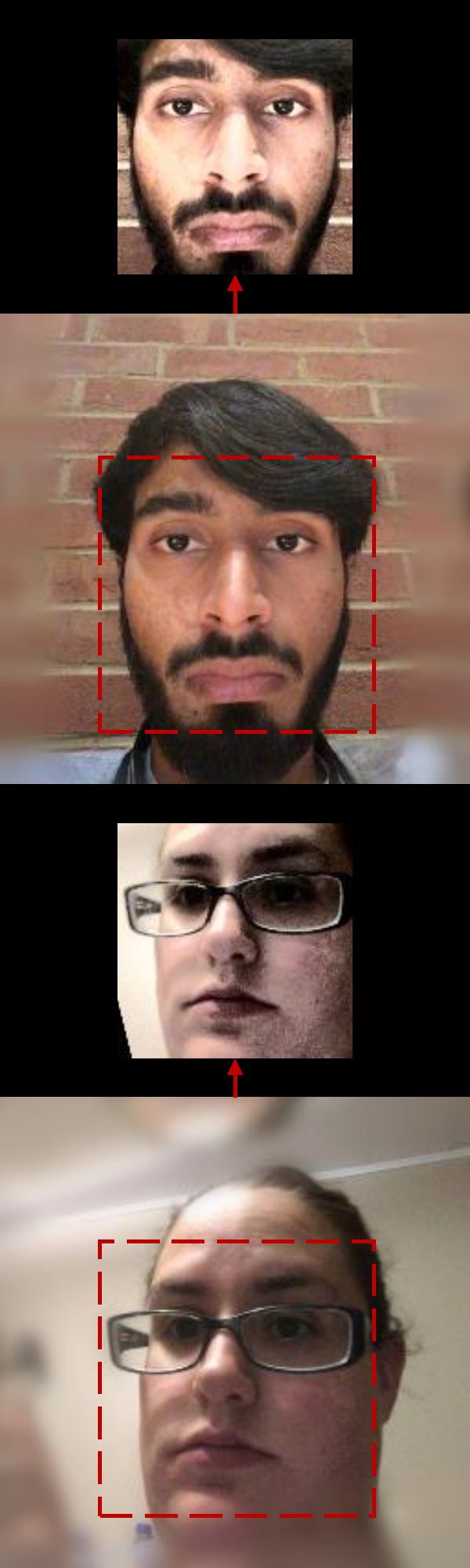}
        \end{minipage}
    \caption{Target}
    \label{fig:tar1}
  \end{subfigure}
  \caption{Qualitatively comparisons between ST-ED and ReDirTrans-GAN. Red boxes represent different face covering ranges.} 
  \label{fig_red_comparison}
\end{figure*}

\subsection{Redirectors in Predefined Latent Space}
\begin{figure*}[t]
    \captionsetup[subfigure]{labelformat=empty}
    \captionsetup[subfigure]{justification=centering}
    \centering
    \begin{subfigure}[t]{0.124\linewidth}
        \begin{minipage}{1\linewidth}
        \includegraphics[width=1\linewidth]{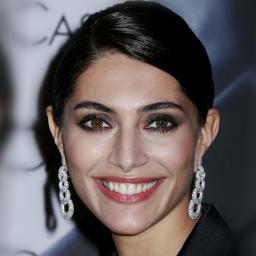}
        \includegraphics[width=1\linewidth]{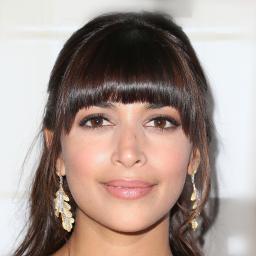}
        \includegraphics[width=1\linewidth]{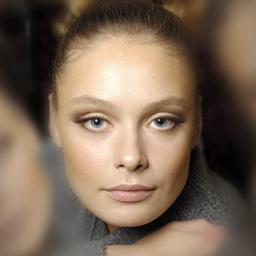}
        \includegraphics[width=1\linewidth]{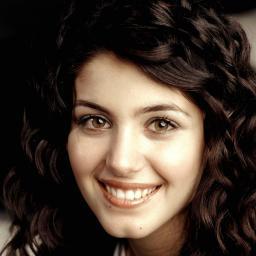}
        \end{minipage}
    \caption{Input}
    \label{fig:input5}
  \end{subfigure}
  \hspace{-0.01\linewidth}
  \centering
    \begin{subfigure}[t]{0.124\linewidth}
        \begin{minipage}{1\linewidth}
        \includegraphics[width=1\linewidth]{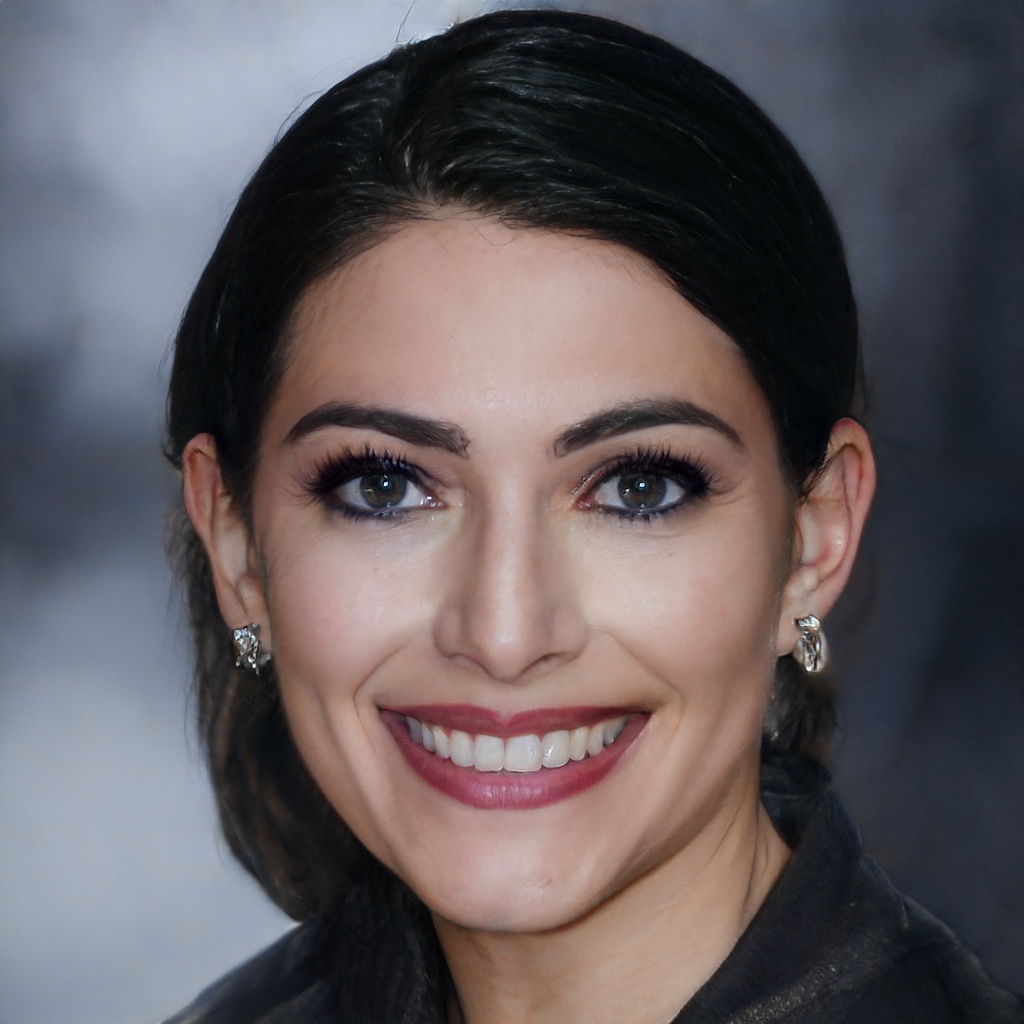}
        \includegraphics[width=1\linewidth]{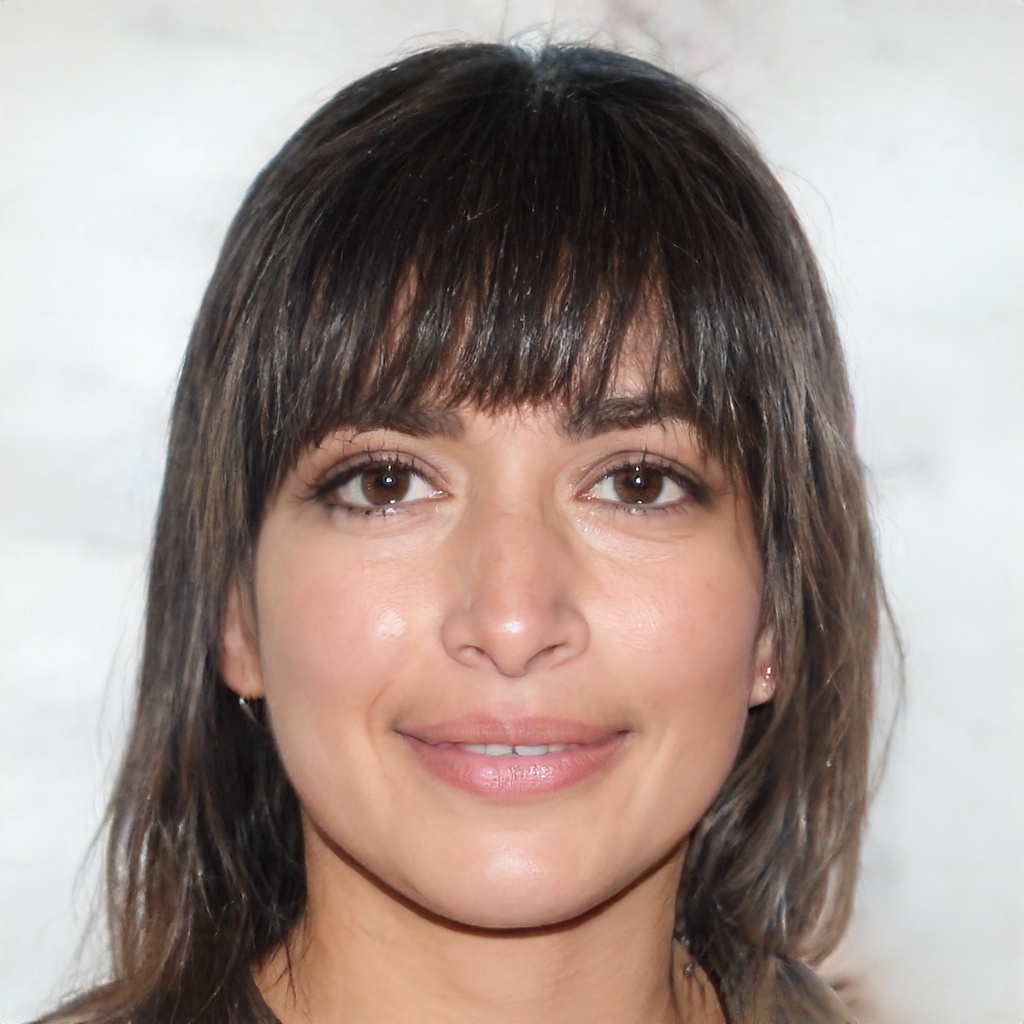}
        \includegraphics[width=1\linewidth]{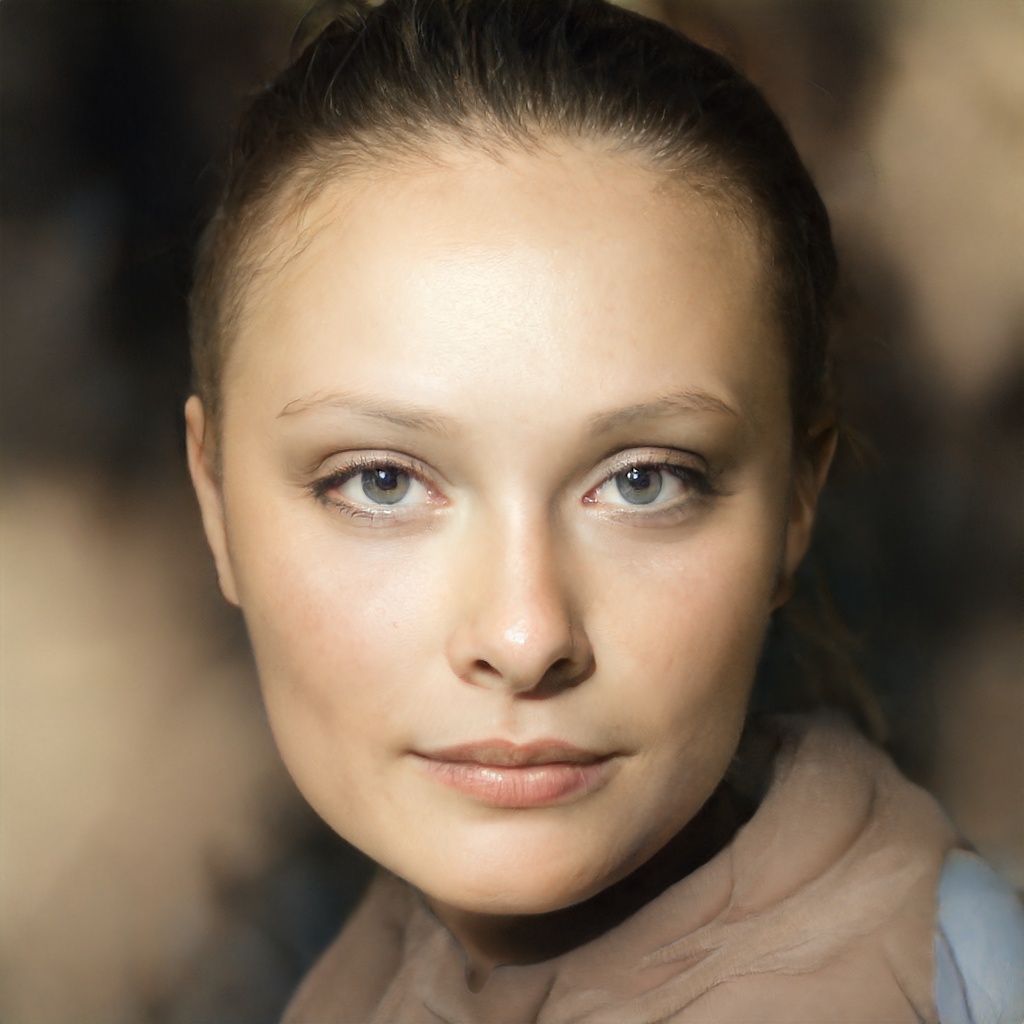}
        \includegraphics[width=1\linewidth]{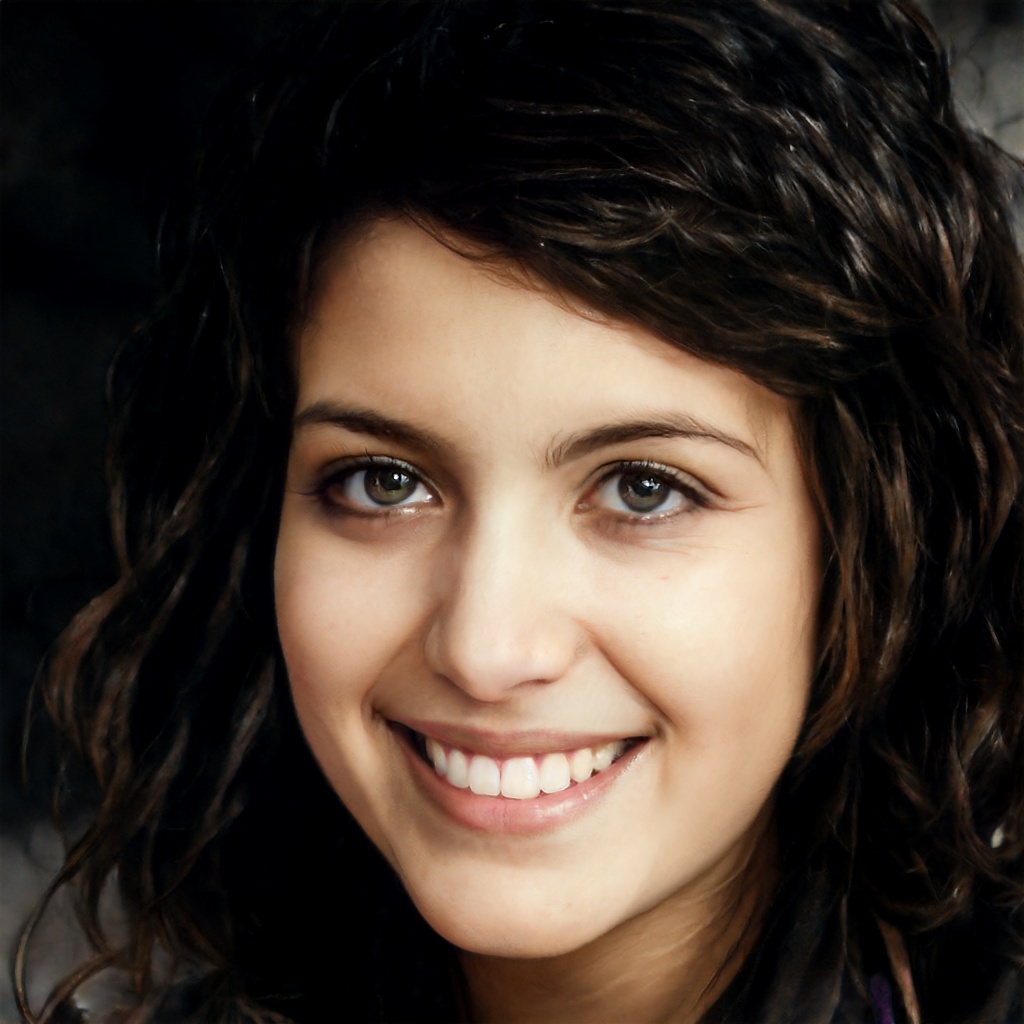}
        \end{minipage}
    \caption{e4e \\ Inversion}
    \label{fig:input6}
  \end{subfigure}
  \hspace{-0.01\linewidth}
  \centering
    \begin{subfigure}[t]{0.124\linewidth}
        \begin{minipage}{1\linewidth}
        \includegraphics[width=1\linewidth]{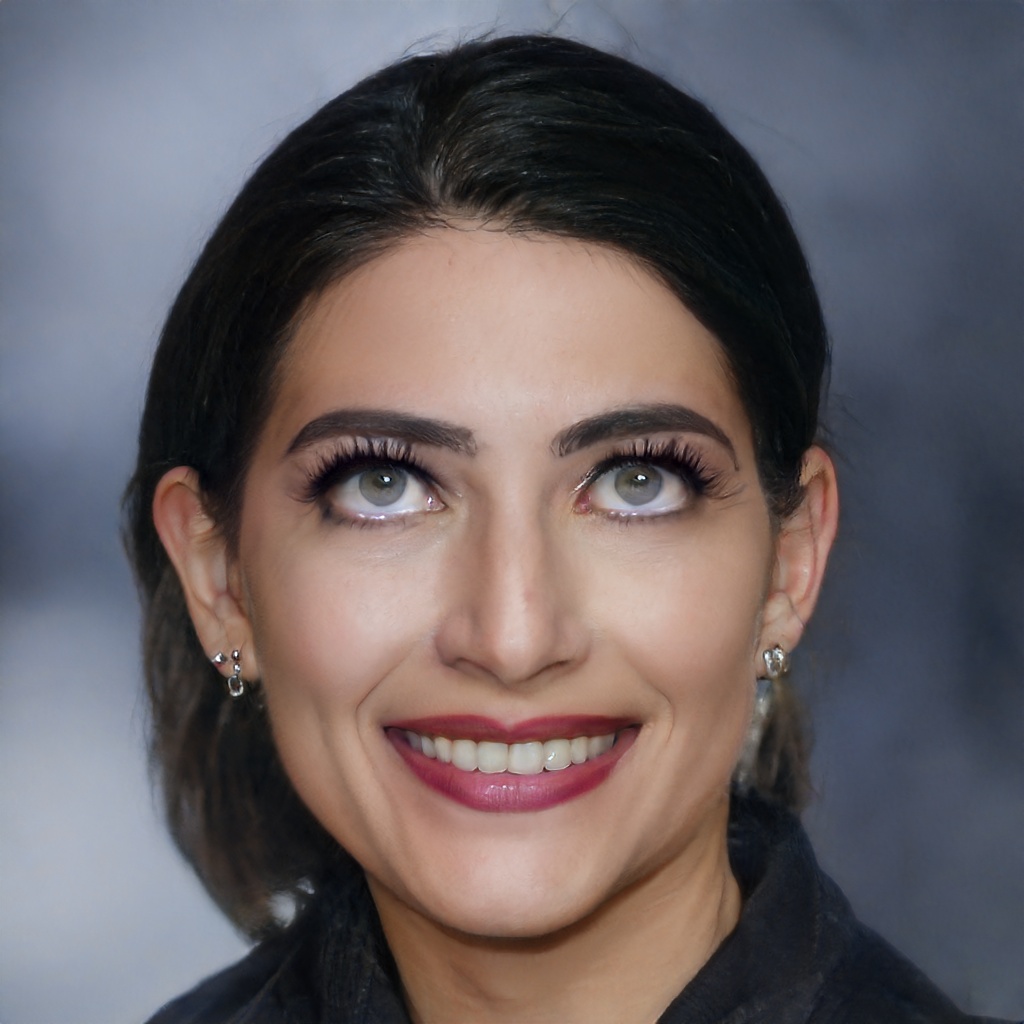}
        \includegraphics[width=1\linewidth]{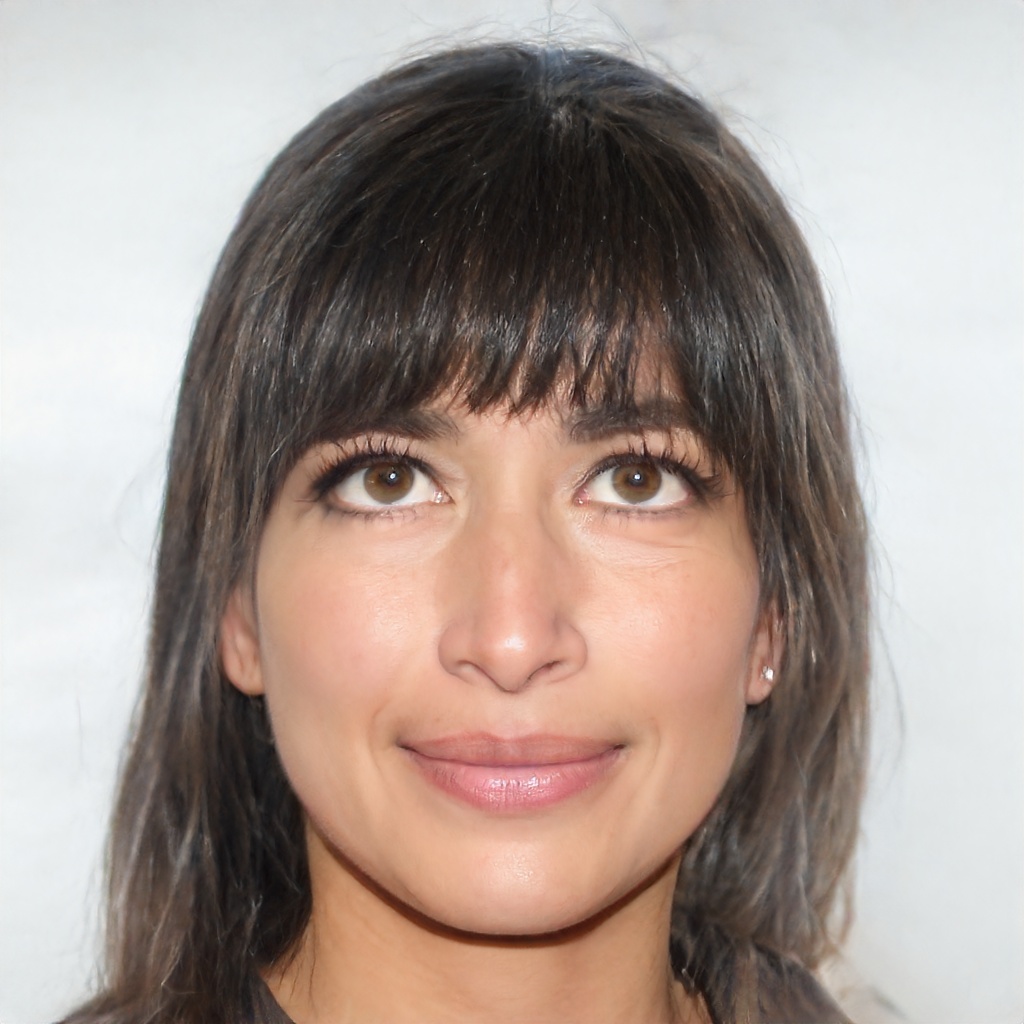}
        \includegraphics[width=1\linewidth]{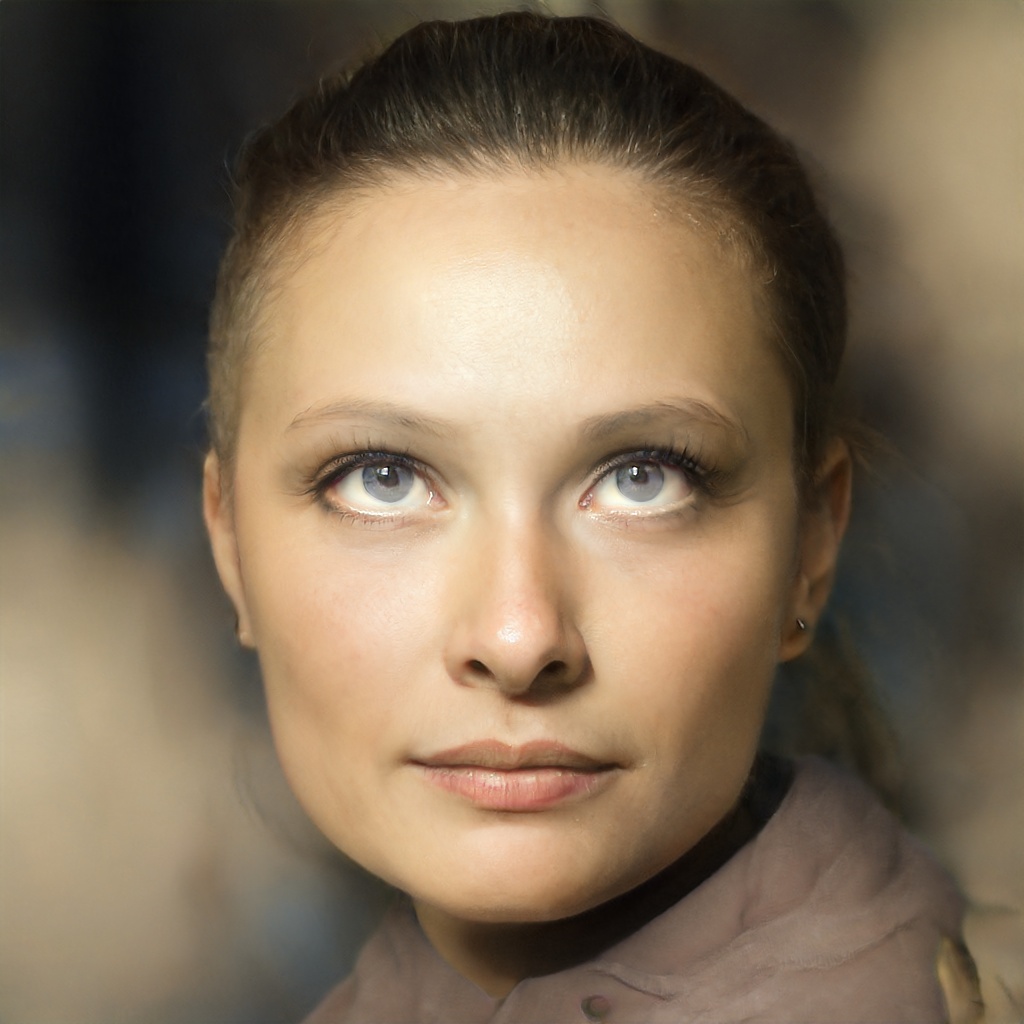}
        \includegraphics[width=1\linewidth]{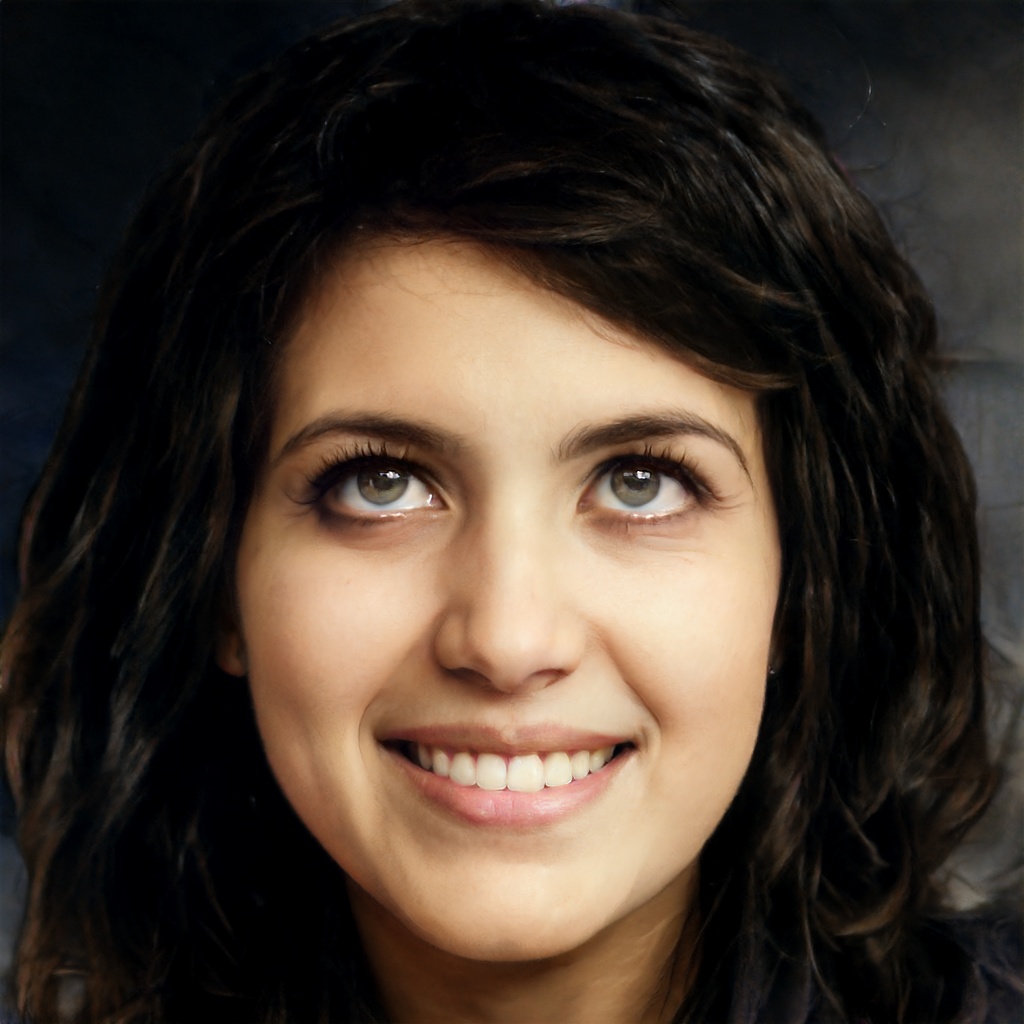}
        \end{minipage}
    \caption{g: ($30^{\circ}$, $0^{\circ}$) \\ h: (\ \ $0^{\circ}$, $0^{\circ}$)}
    \label{fig:input7}
  \end{subfigure}
  \hspace{-0.01\linewidth}
  \centering
    \begin{subfigure}[t]{0.124\linewidth}
        \begin{minipage}{1\linewidth}
        \includegraphics[width=1\linewidth]{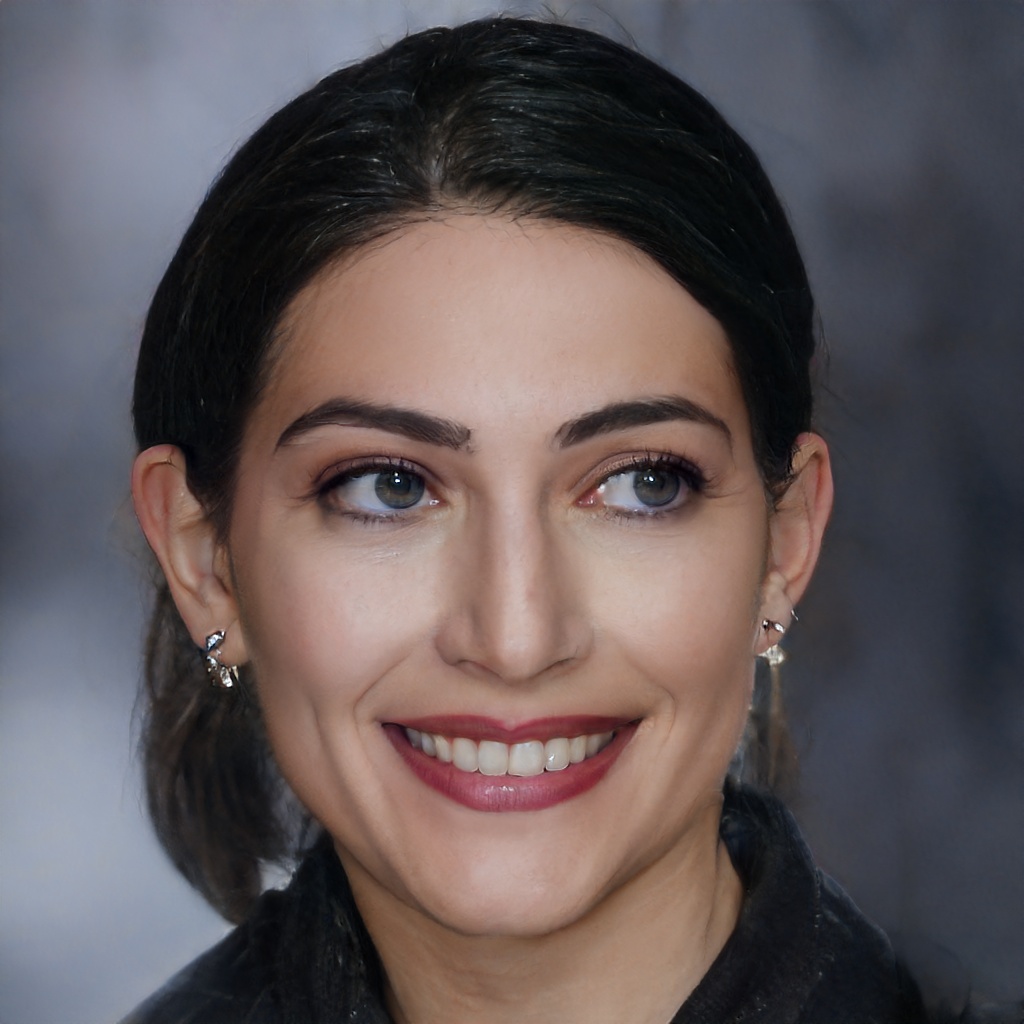}
        \includegraphics[width=1\linewidth]{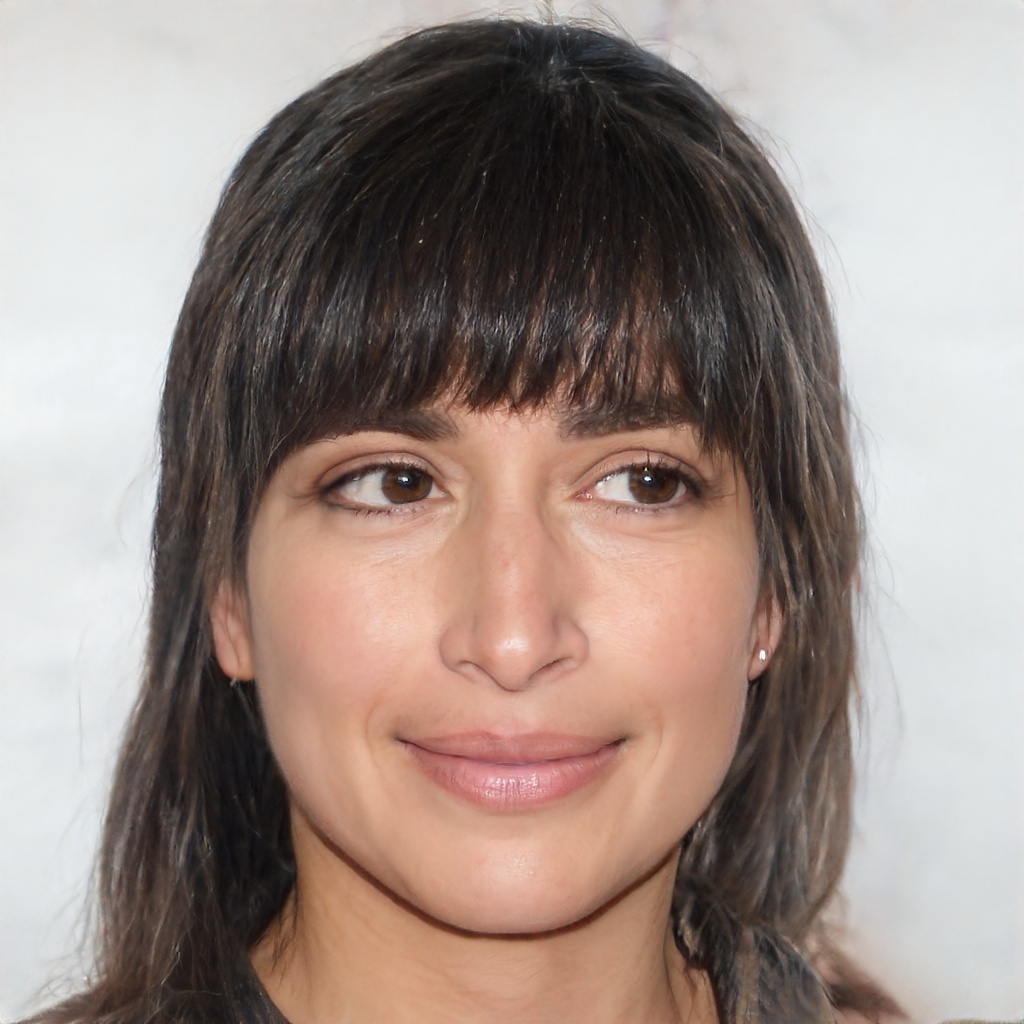}
        \includegraphics[width=1\linewidth]{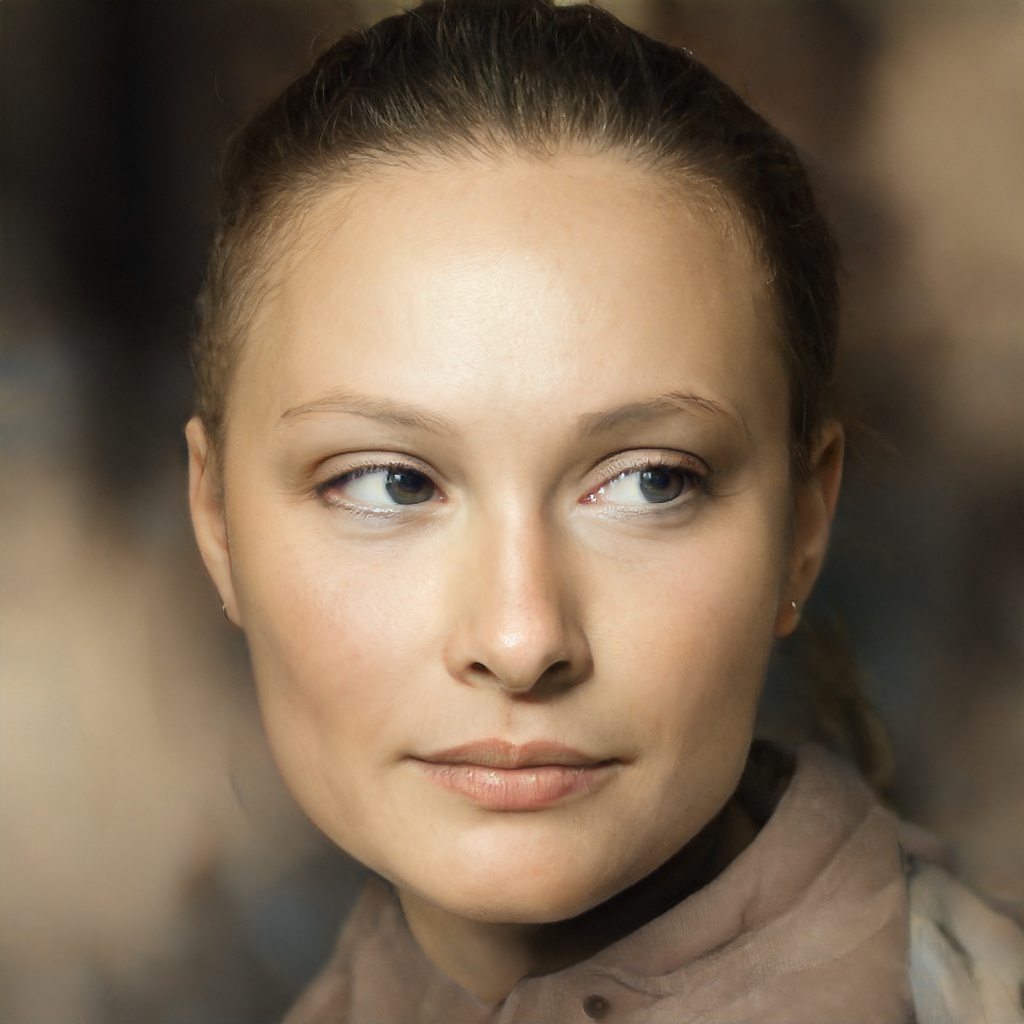}
        \includegraphics[width=1\linewidth]{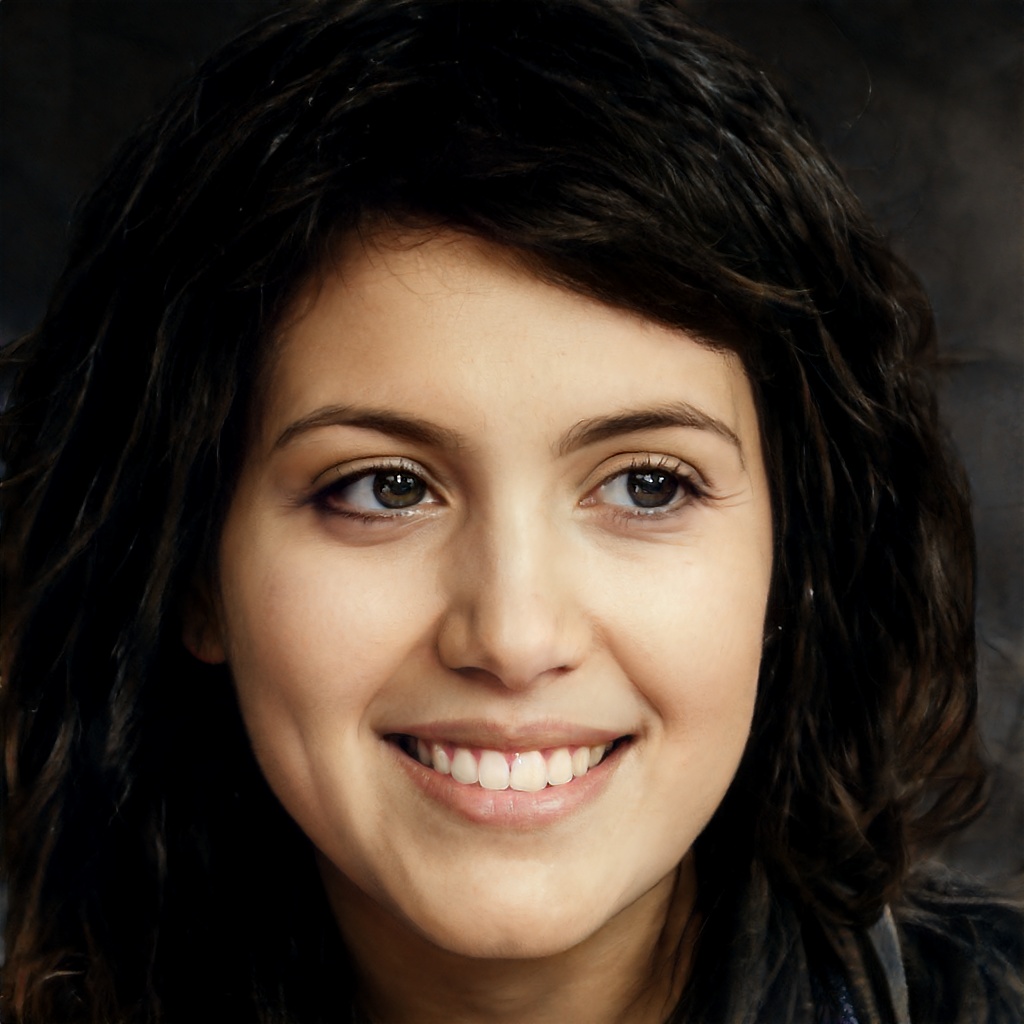}
        \end{minipage}
    \caption{g: ($0^{\circ}$, $-20^{\circ}$) \\ h: ($0^{\circ}$, \ \ \ \ \ $0^{\circ}$)}
    \label{fig:input8}
  \end{subfigure}
  \hspace{-0.01\linewidth}
  \centering
    \begin{subfigure}[t]{0.124\linewidth}
        \begin{minipage}{1\linewidth}
        \includegraphics[width=1\linewidth]{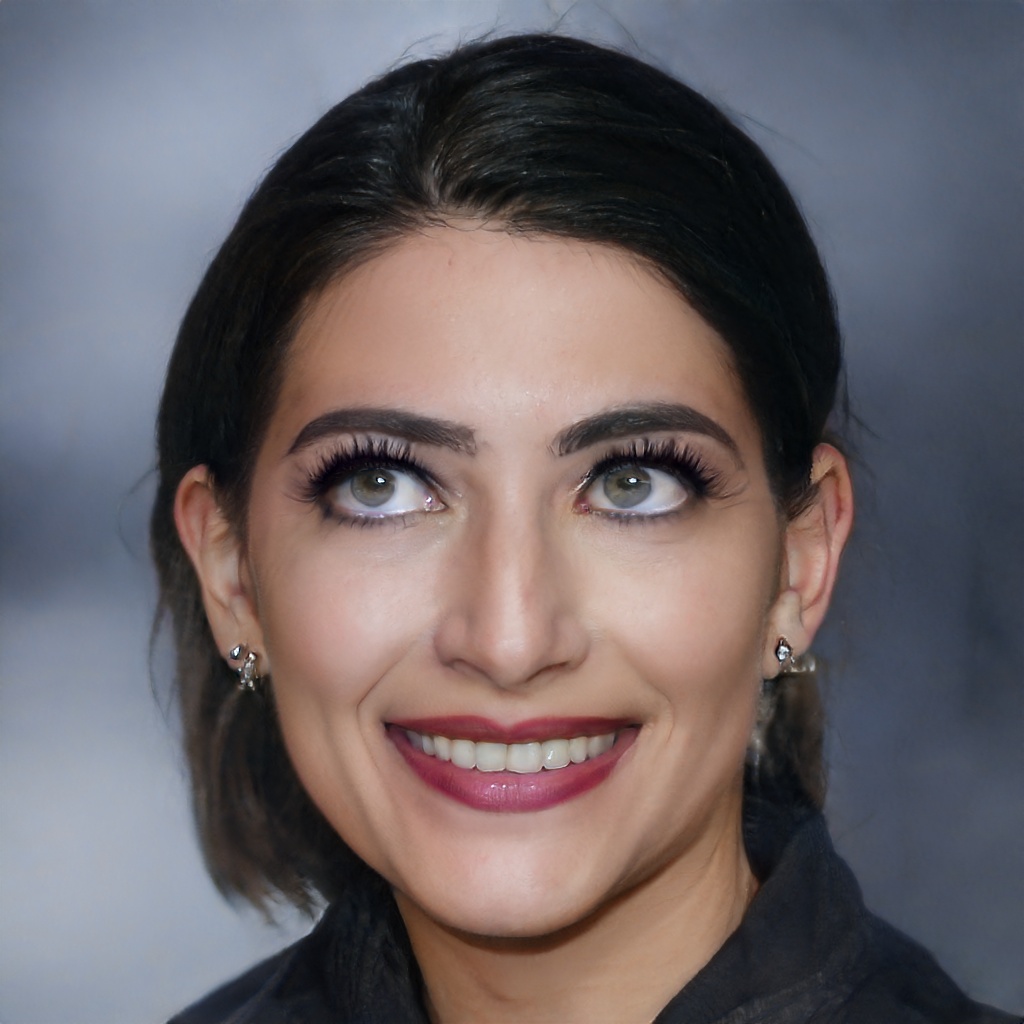}
        \includegraphics[width=1\linewidth]{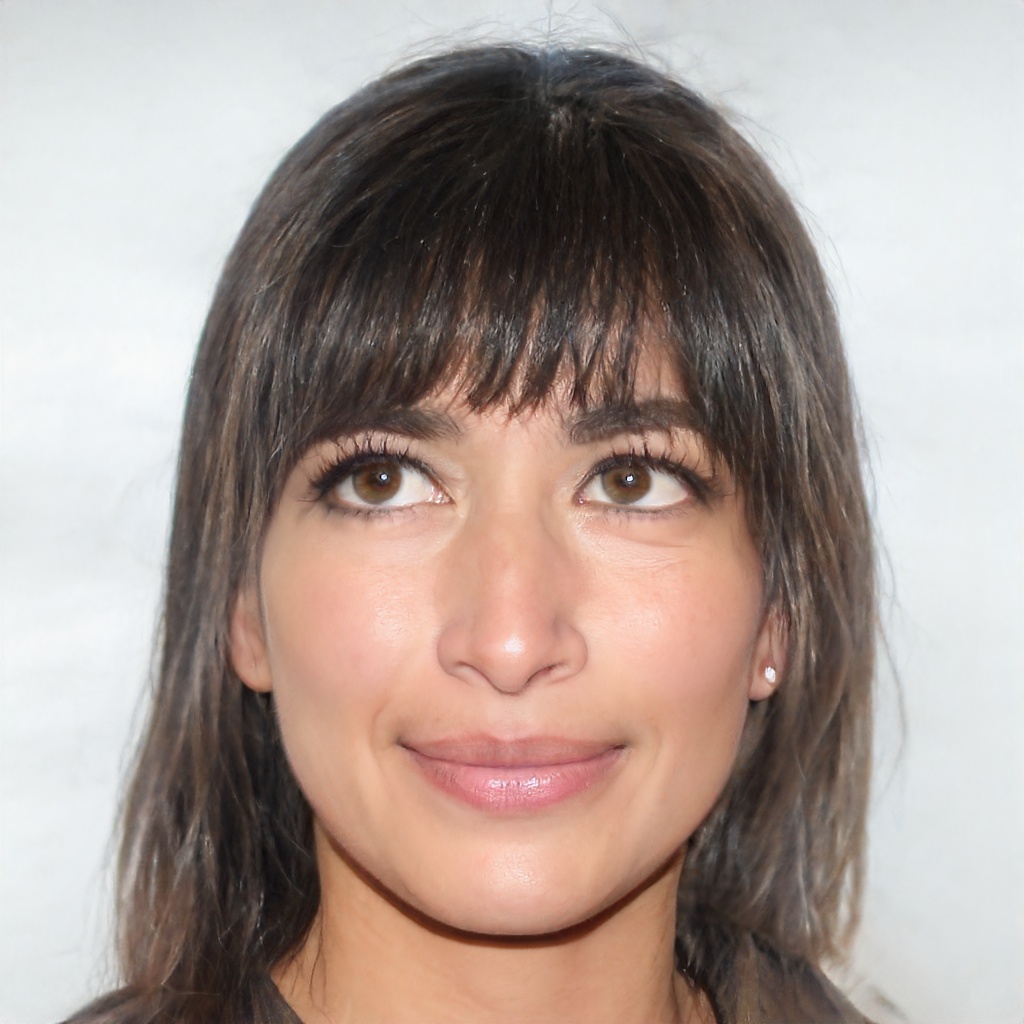}
        \includegraphics[width=1\linewidth]{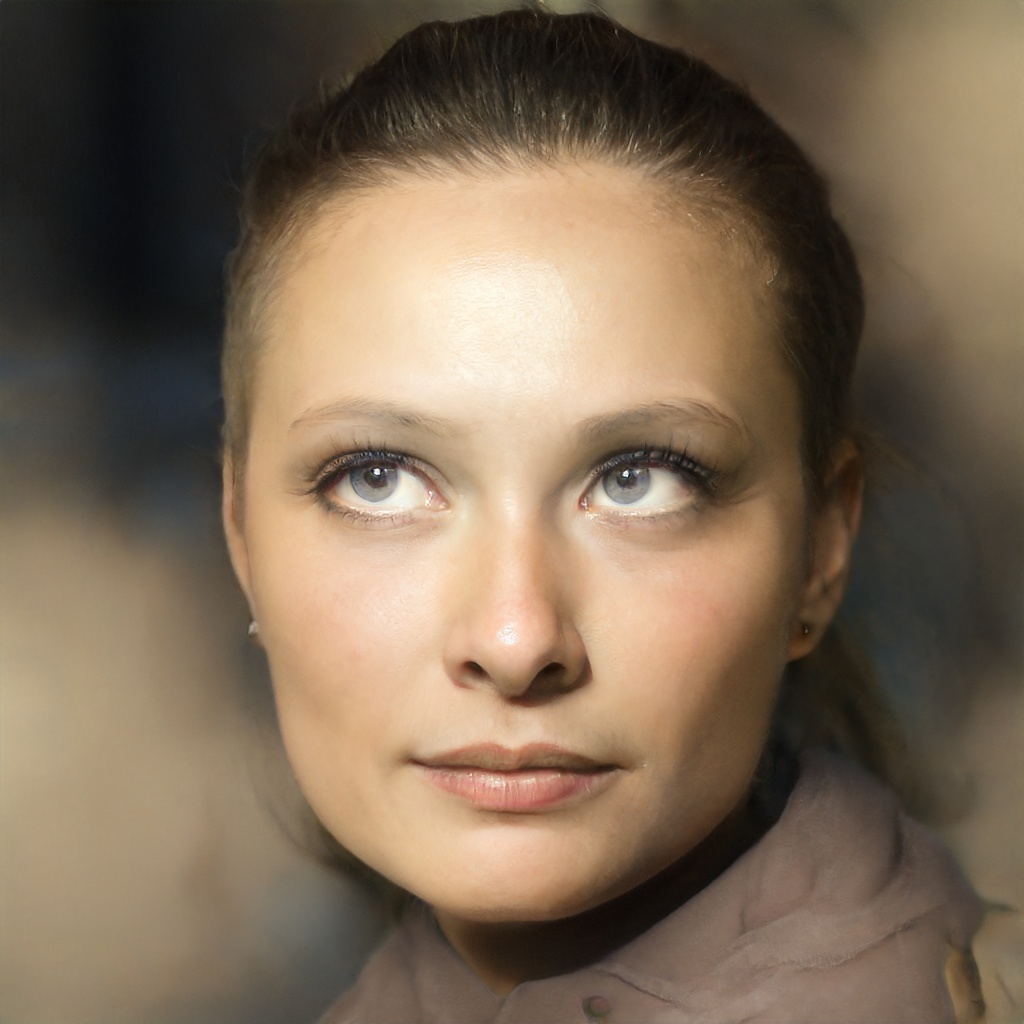}
        \includegraphics[width=1\linewidth]{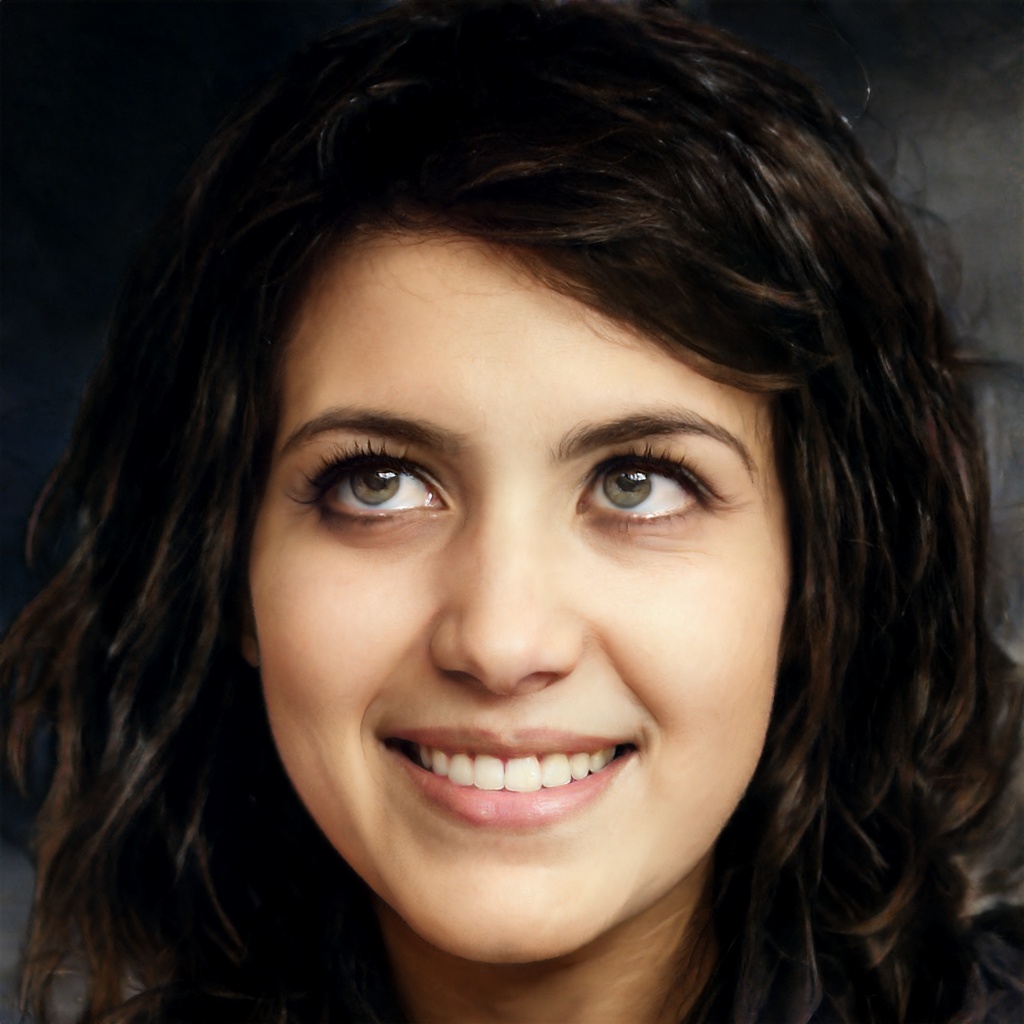}
        \end{minipage}
    \caption{g: ($30^{\circ}$, $20^{\circ}$) \\ h: (\ \ $0^{\circ}$, \ \ $0^{\circ}$)}
    \label{fig:input9}
  \end{subfigure}
  \hspace{-0.01\linewidth}
  \centering
    \begin{subfigure}[t]{0.124\linewidth}
        \begin{minipage}{1\linewidth}
        \includegraphics[width=1\linewidth]{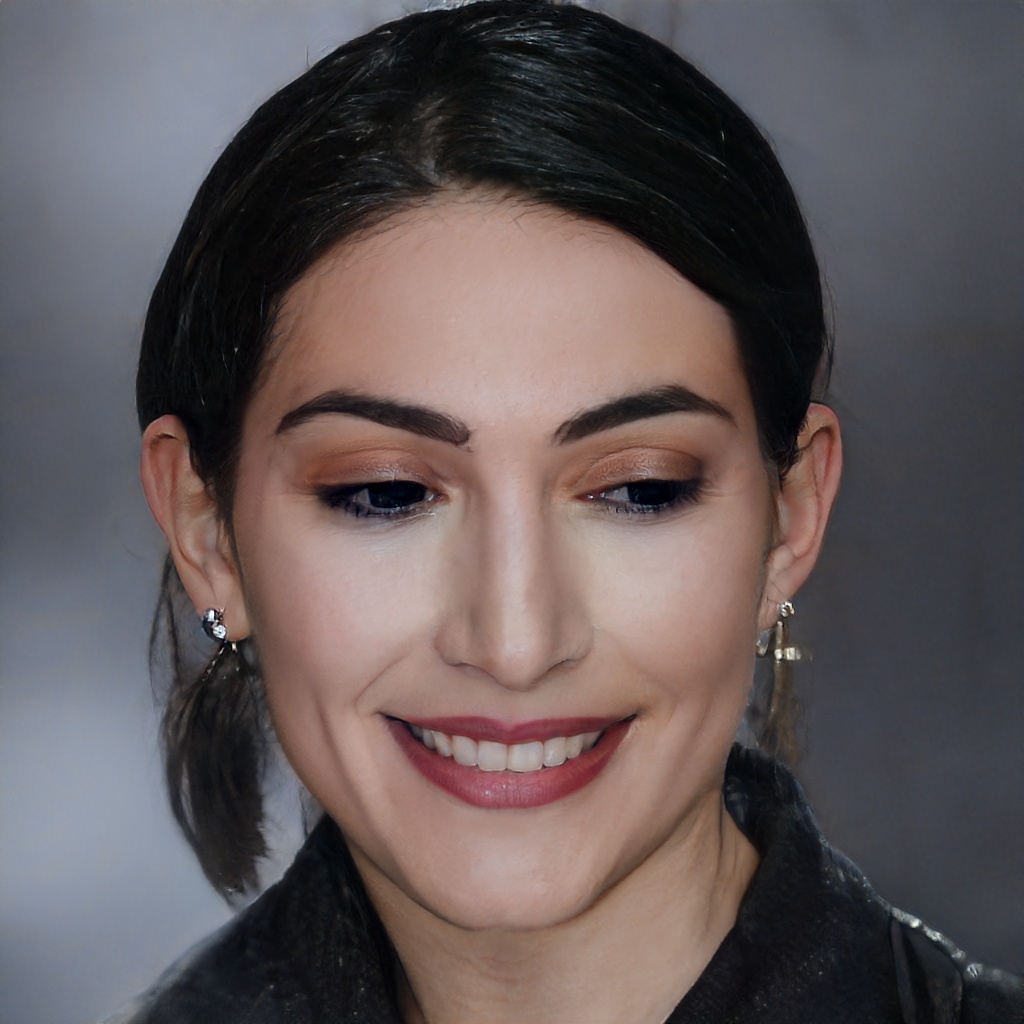}
        \includegraphics[width=1\linewidth]{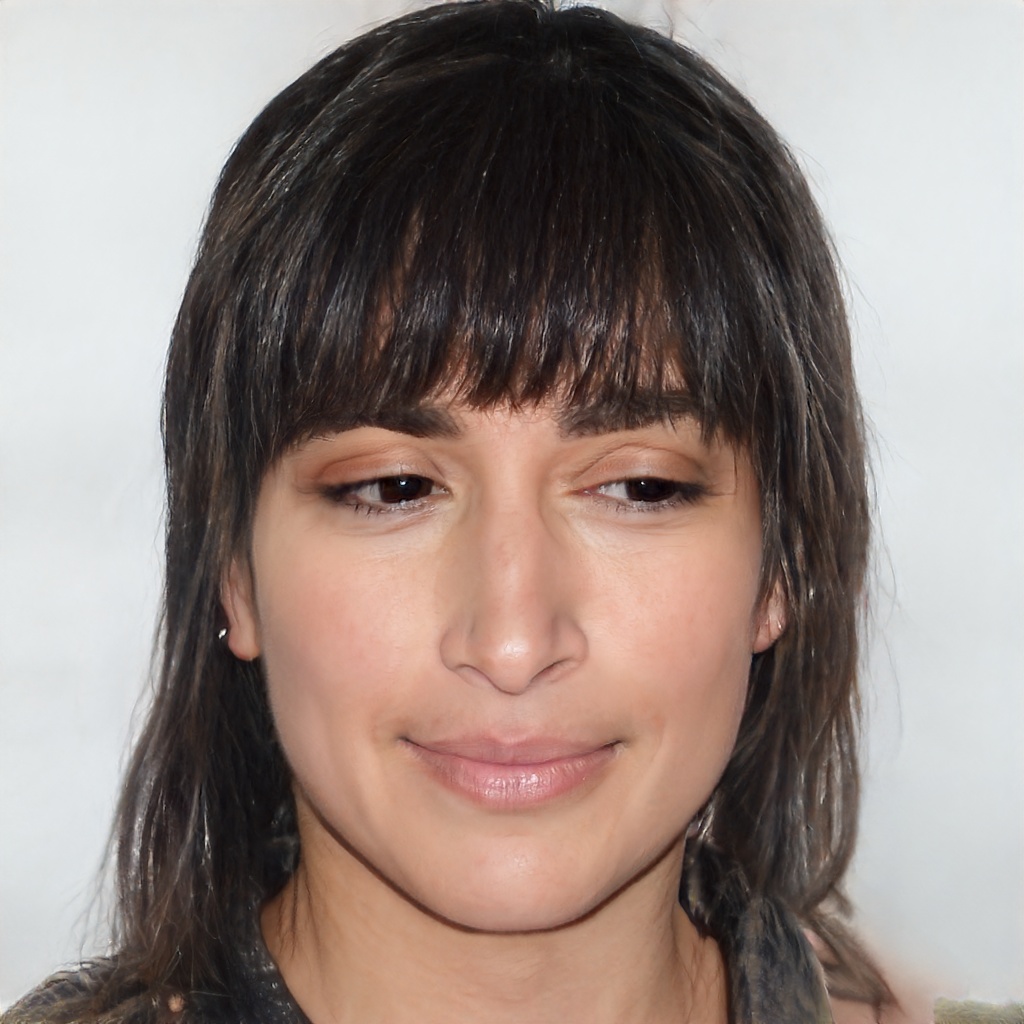}
        \includegraphics[width=1\linewidth]{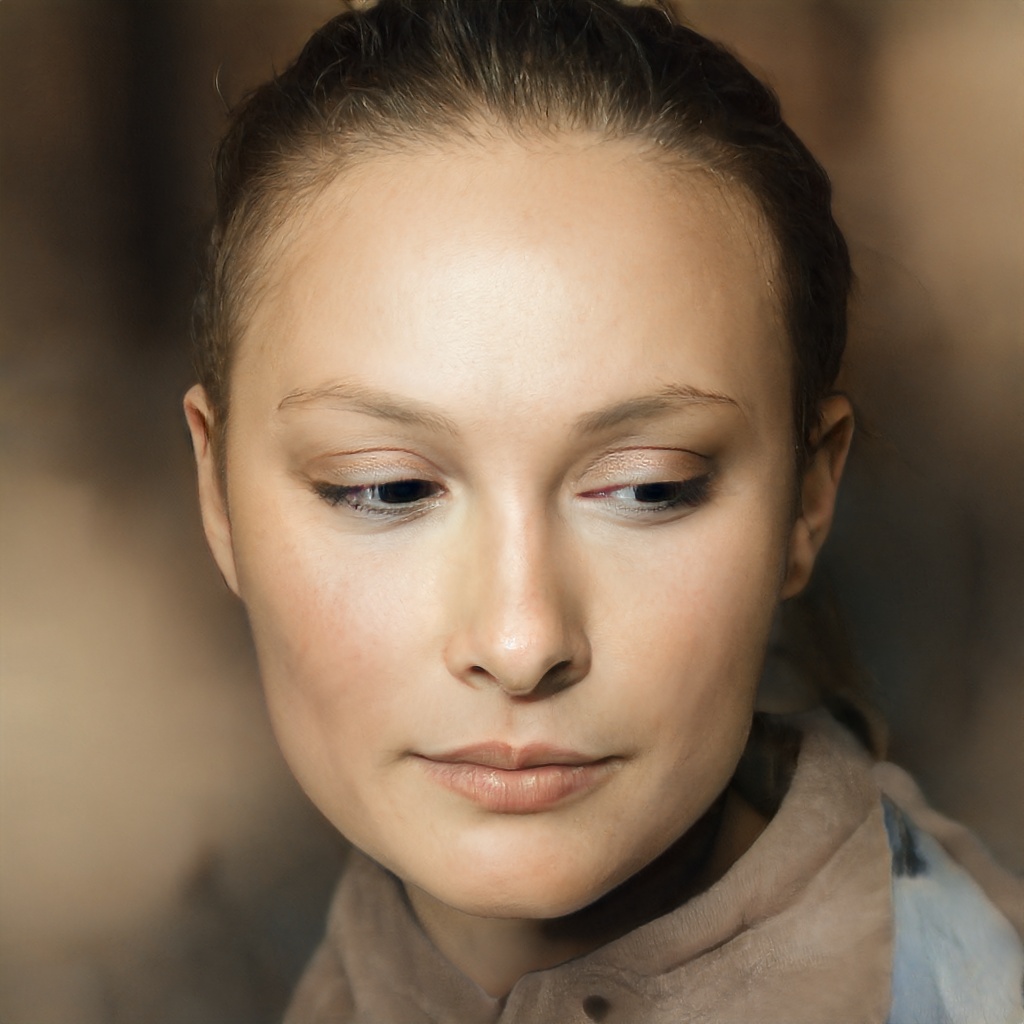}
        \includegraphics[width=1\linewidth]{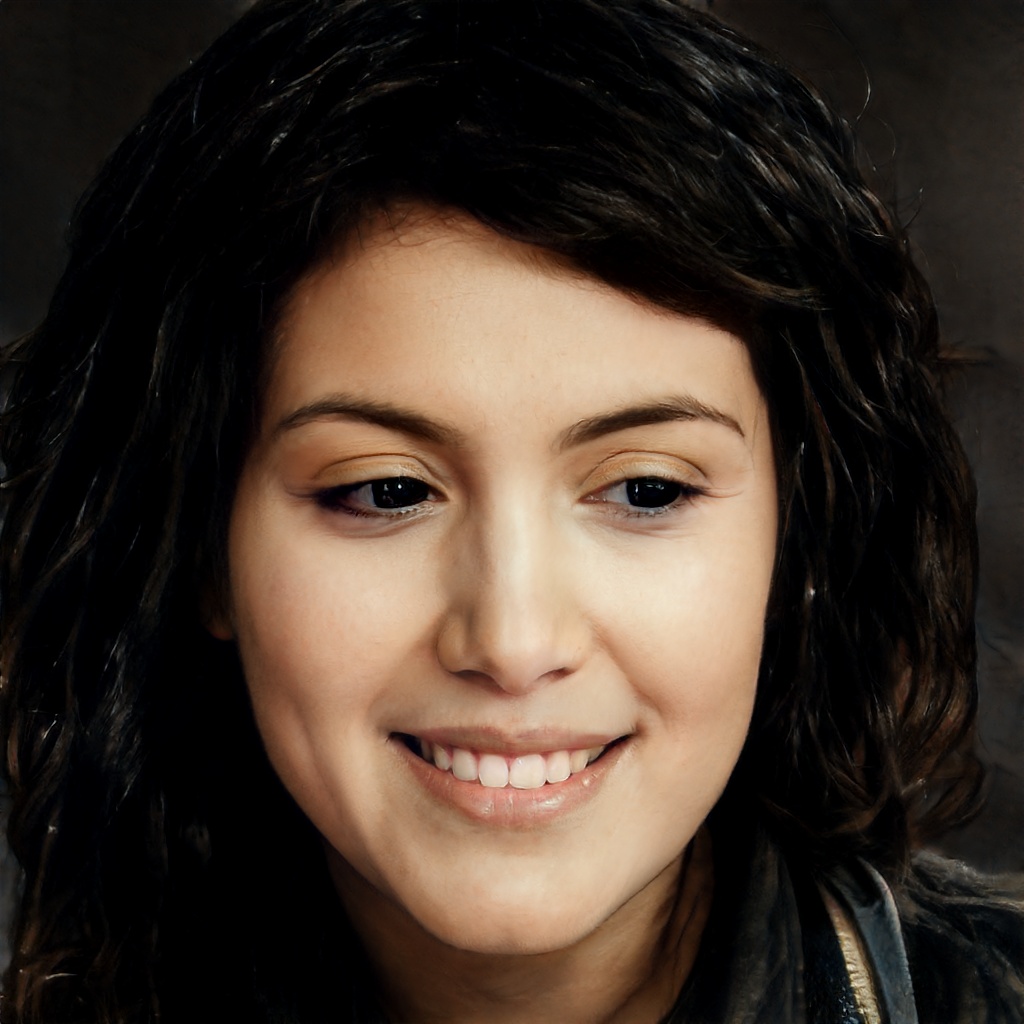}
        \end{minipage}
    \caption{g: ($-30^\circ$, $-20^\circ$) \\ h: (\ \ \ \ \ $0^\circ$, \ \ \ \ \ $0^\circ$)}
    \label{fig:input10}
  \end{subfigure}
  \hspace{-0.01\linewidth}
  \centering
    \begin{subfigure}[t]{0.124\linewidth}
        \begin{minipage}{1\linewidth}
        \includegraphics[width=1\linewidth]{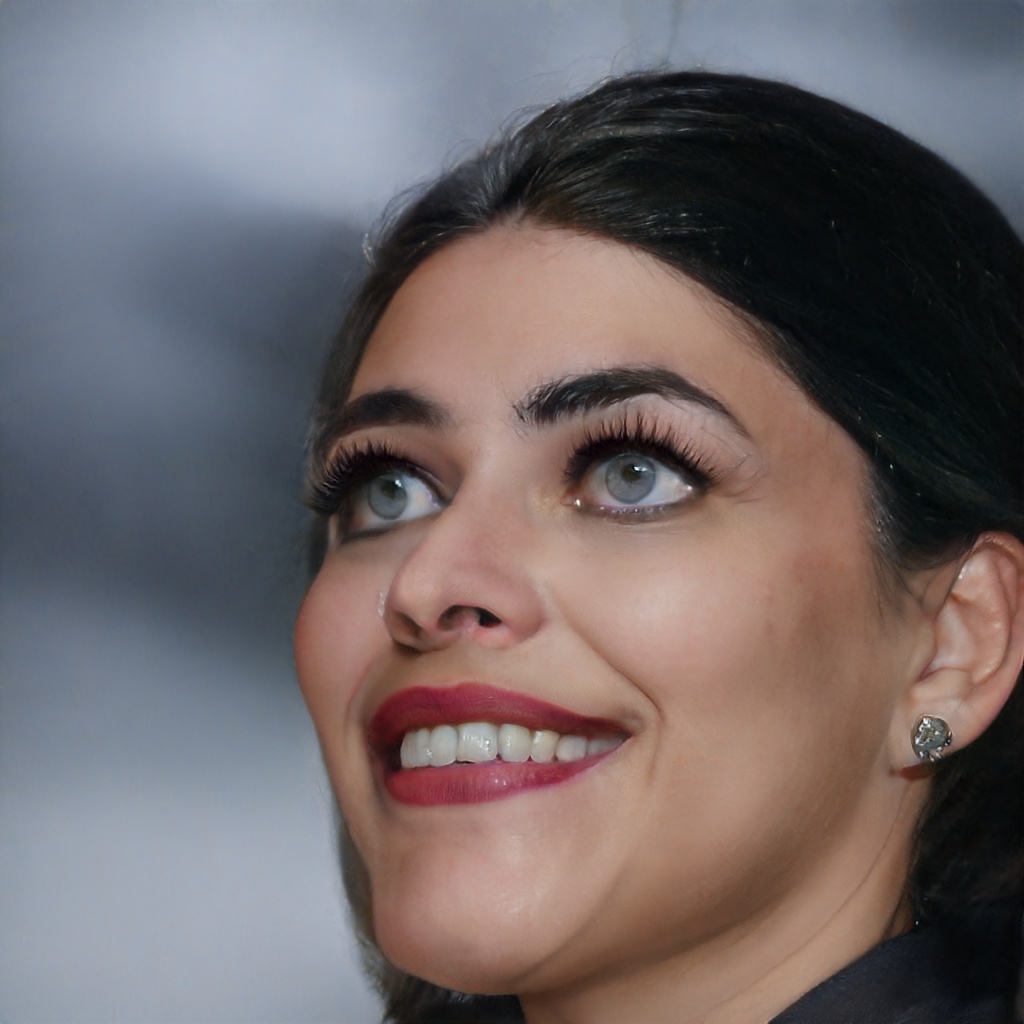}
        \includegraphics[width=1\linewidth]{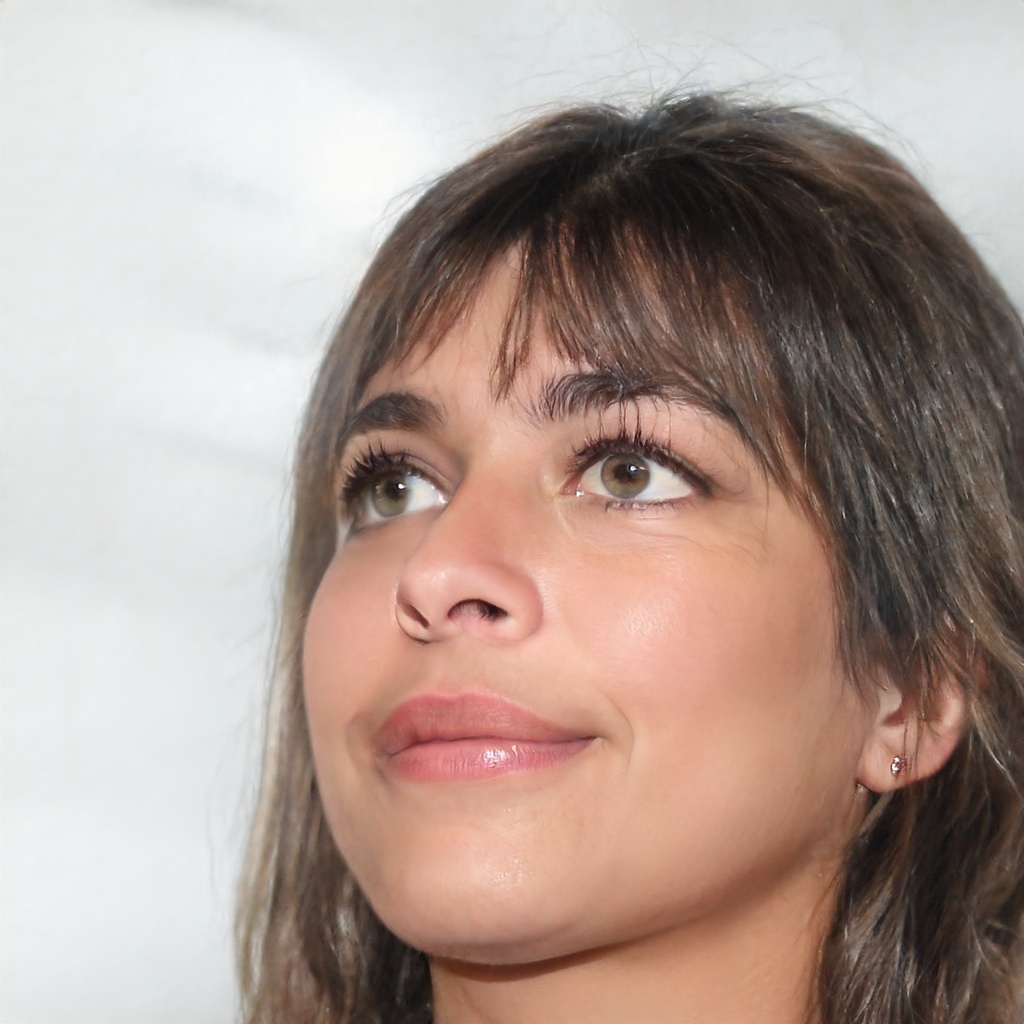}
        \includegraphics[width=1\linewidth]{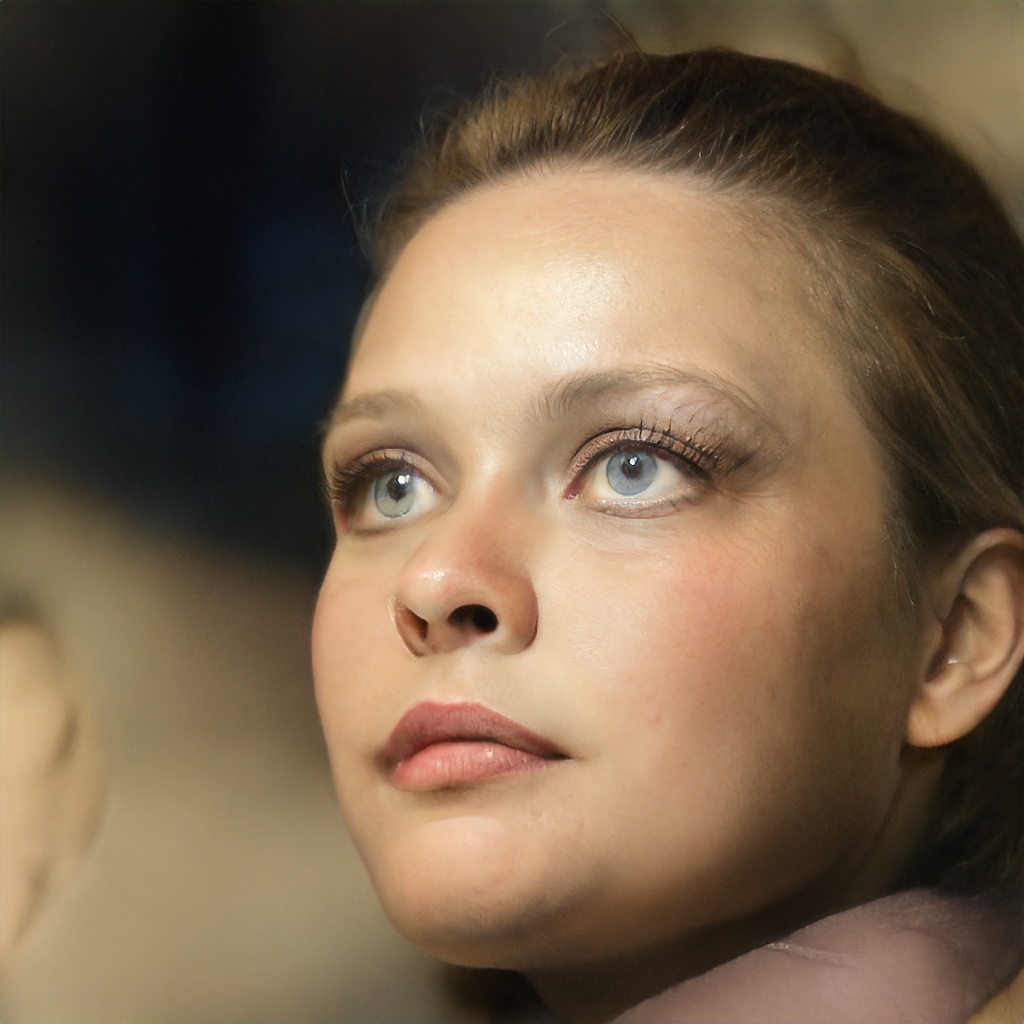}
        \includegraphics[width=1\linewidth]{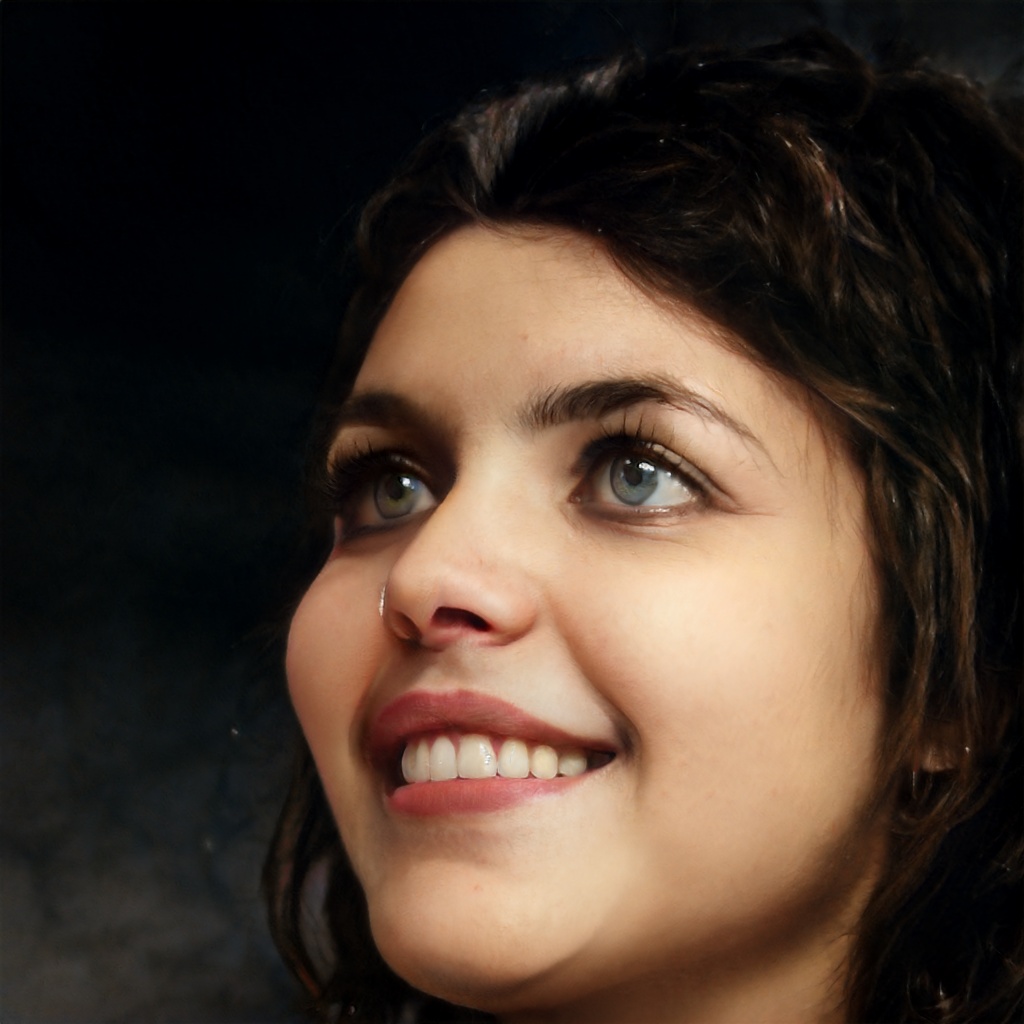}
        \end{minipage}
    \caption{g: ($30^\circ$, $20^\circ$) \\ h: ($30^\circ$, $30^\circ$)}
    \label{fig:input11}
  \end{subfigure}
  \hspace{-0.01\linewidth}
  \centering
    \begin{subfigure}[t]{0.124\linewidth}
        \begin{minipage}{1\linewidth}
        \includegraphics[width=1\linewidth]{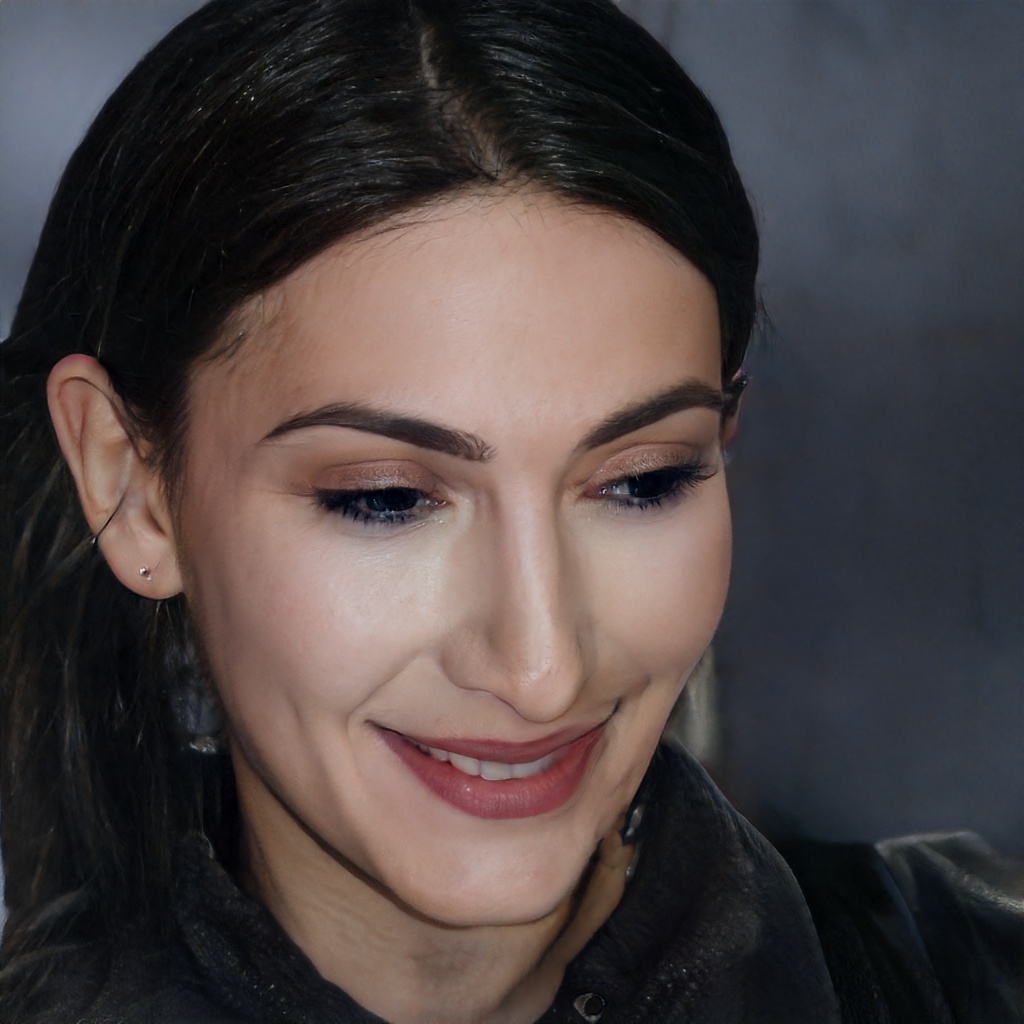}
        \includegraphics[width=1\linewidth]{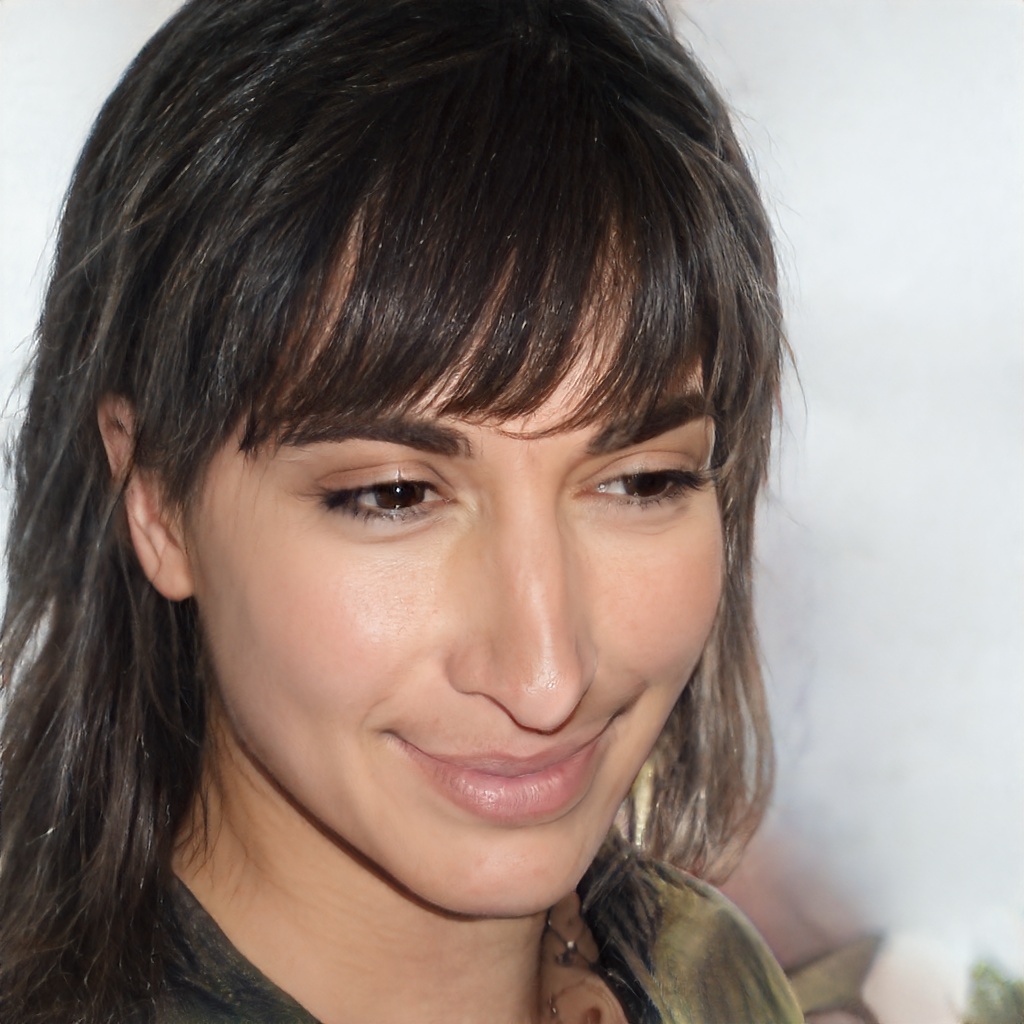}
        \includegraphics[width=1\linewidth]{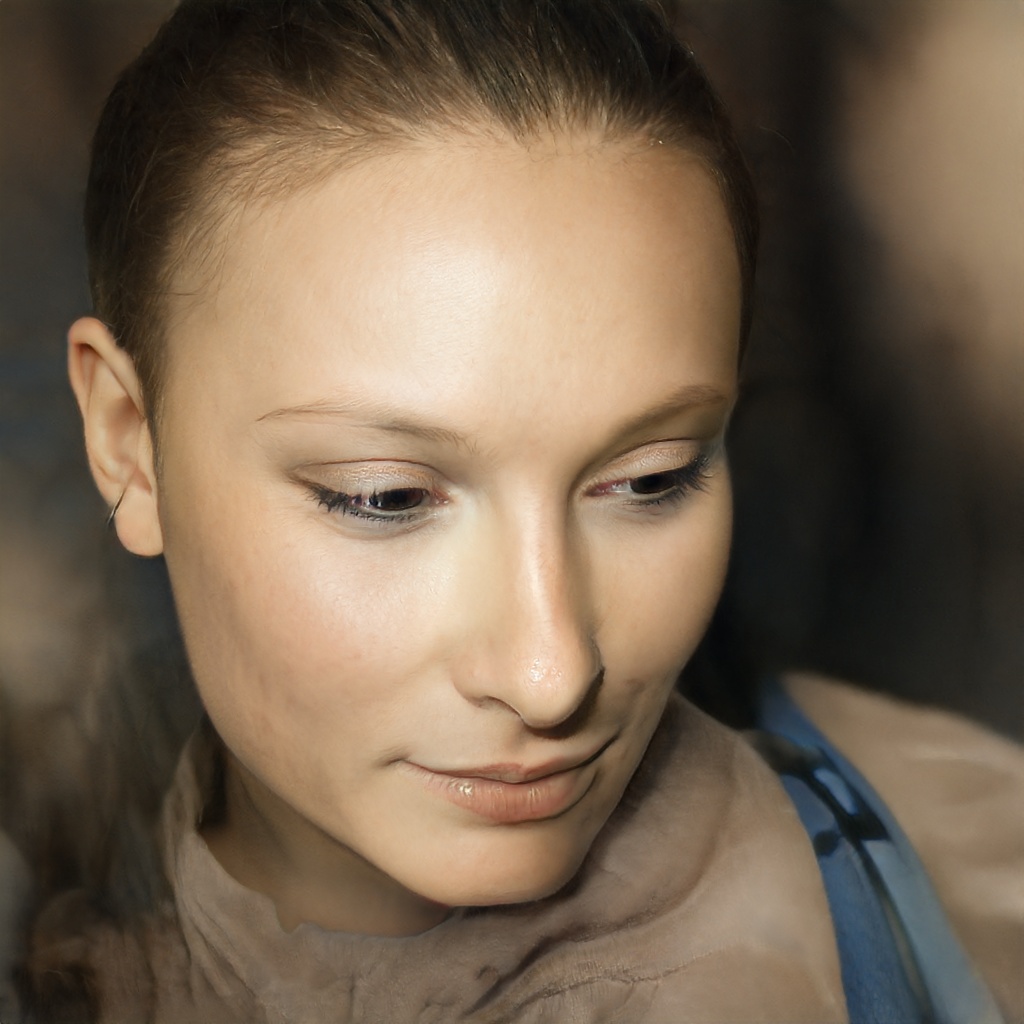}
        \includegraphics[width=1\linewidth]{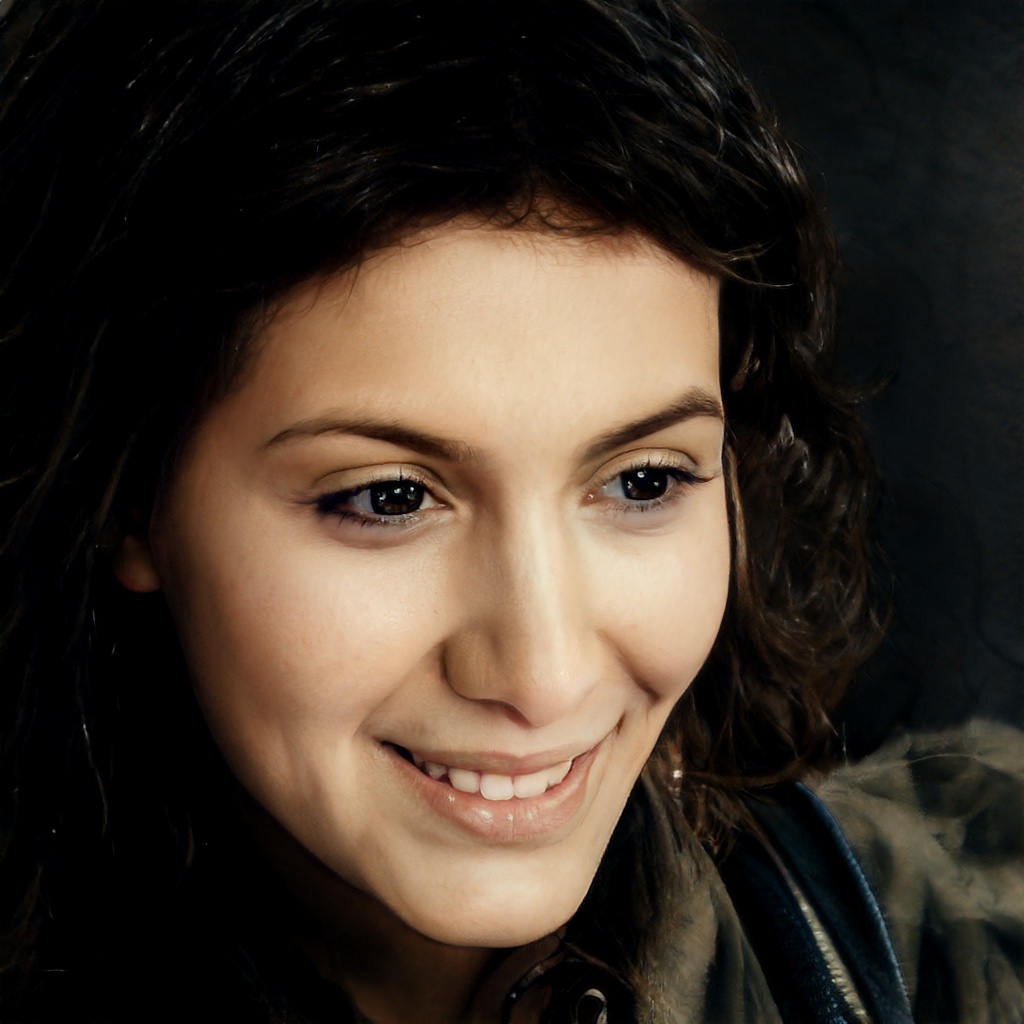}
        \end{minipage}
    \caption{g: ($-30^\circ$, $-20^\circ$) \\ h: ($-15^\circ$, $-15^\circ$)}
    \label{fig:input12}
  \end{subfigure}
  \caption{Redirection results given assigned (pitch, yaw) conditions of gaze directions (g) and head orientations (h). The first two columns are input and inversion results with e4e \cite{tov2021designing} and StyleGAN \cite{karras2020analyzing}. The following columns are redirected samples with assigned redirection values based on the latent code estimated from e4e.} 
  \label{fig_red_wo_target}
\end{figure*}

Except for using the trainable encoder-decoder pair of ST-ED, we also implemented our proposed ReDirTrans within a predefined feature space $W^+$ to achieve redirection task in full face images with desired resolution. 
We utilized e4e \cite{tov2021designing} as the pre-trained encoder, which can transform input images into latent vectors in $W^+$, and we chose StyleGAN2 \cite{karras2020analyzing} as the pre-trained generator to build ReDirTrans-GAN. 
Fig. \ref{fig_red_comparison} shows qualitative comparison between ST-ED and ReDirTrans-GAN in the GazeCapture test subset with providing target images from the same subject. 
ReDirTrans-GAN successfully redirected gaze directions and head orientations to the status provided by target images while maintaining the same appearance patterns with $1024\times1024$ full face images.
Due to the design of ReDirTrans, which maintains unrelated attributes and appearance information in the initial latent space instead of going through the projection-deprojection process, ReDirTrans-GAN keeps more facial attributes such as expressions, mustaches, bangs compared with ST-ED. 
Fig. \ref{fig_red_wo_target} presents qualitative results with assigned conditions (pitch and yaw of gaze directions and head orientations) in CelebA-HQ \cite{karras2017progressive}. 
ReDirTrans-GAN can achieve out-of-domain redirection tasks in predefined feature space while maintaining other facial attributes. 

\begin{table}[t]
\centering
\begin{tabular}{@{}clcclcc@{}}
\toprule
\multirow{2}{*}{\textbf{Q\%}} &  & \multicolumn{2}{c}{GazeCapture} &  & \multicolumn{2}{c}{MPIIFaceGaze} \\ \cmidrule(lr){3-4} \cmidrule(l){6-7} 
   &  & Raw $\downarrow$   & Aug $\downarrow$  &  & Raw $\downarrow$           & Aug $\downarrow$           \\ \midrule
25 &  & 5.875 & \textbf{5.238} &  & 8.607          & \textbf{7.096}          \\
50 &  & 4.741 & \textbf{4.506} &  & 6.787 & \textbf{6.113} \\
75 &  & 4.308 & \textbf{4.200}   &  & 6.165          & \textbf{5.767}          \\ \bottomrule
\end{tabular}
\caption{Learning-based gaze estimation errors (in degrees) in GazeCapture and MPIIFaceGaze with or without redirected data augmentation. $Q\%$ represents percent of labeled data in 10,000 images for training ReDirTrans-GAN. `Raw' or `Aug' mean training the gaze estimator with real data or with real \& redirected data.}
\label{tab_aug}
\end{table}

\subsection{Data Augmentation}
\label{subsec:augmentation}
To solve data scarcity of the downstream task: learning-based gaze estimation, we utilized redirected samples with assigned gaze directions and head orientations to augment training data. 
We randomly chose $10,000$ images from the GazeCapture training subset to retrain ReDirTrans-GAN with using only $Q\%$ ground-truth labels of them. 
The HeadGazeNet $\xi_{hg}(\cdot)$ was also retrained given the same $Q\%$ labeled data and $Q\in\{25, 50, 75\}$. 
Then we utilized ReDirTrans-GAN to generate redirected samples given provided conditions over $Q\%$ labeled real data and combined the real and redirected data as an augmented dataset with size $2\times10,000\times Q\%$ for training a gaze estimator. 
Table \ref{tab_aug} presented within-dataset and cross-dataset performance and demonstrated consistent improvements for the downstream task given redirected samples as data augmentation. 

\begin{figure}[t]
    \centering
    \begin{subfigure}[t]{0.249\linewidth}
        \begin{minipage}[b]{1\linewidth}
        \includegraphics[width=1\linewidth]{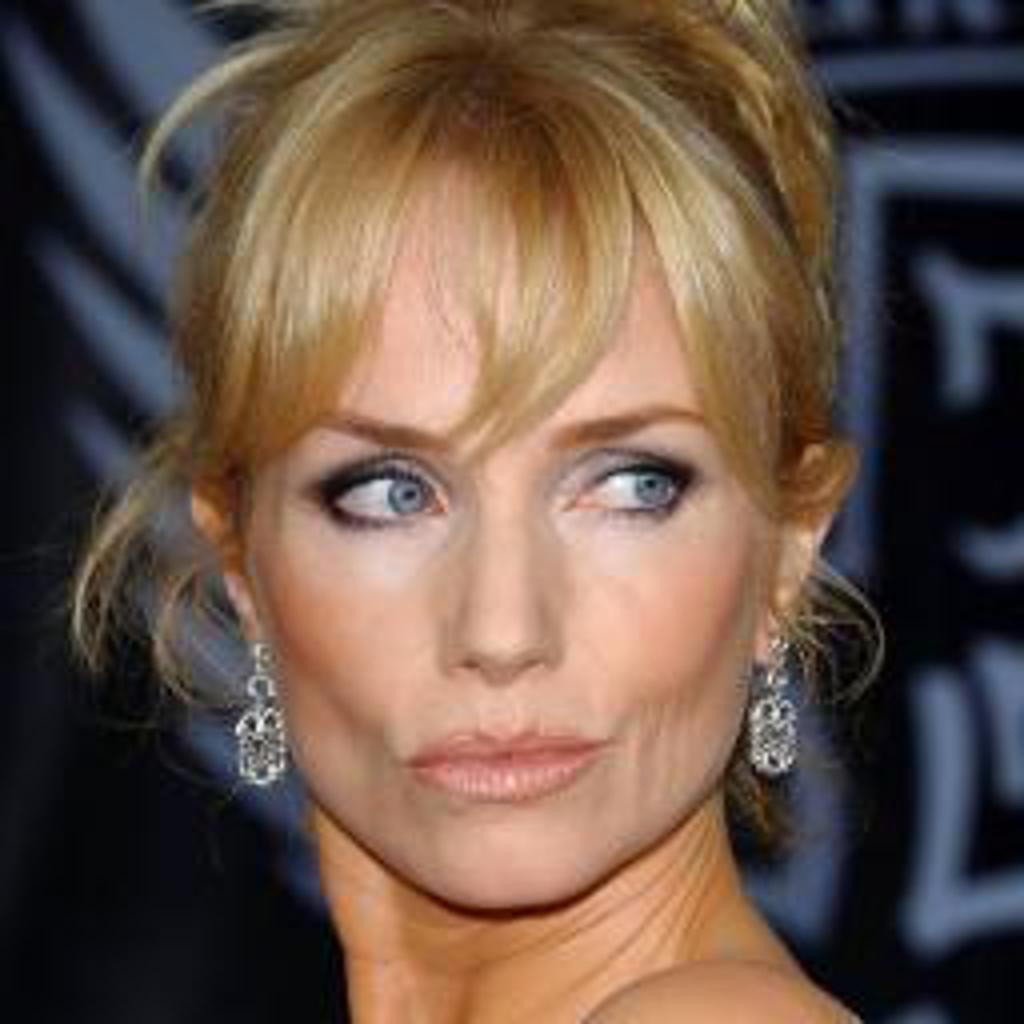}
        \includegraphics[width=1\linewidth]{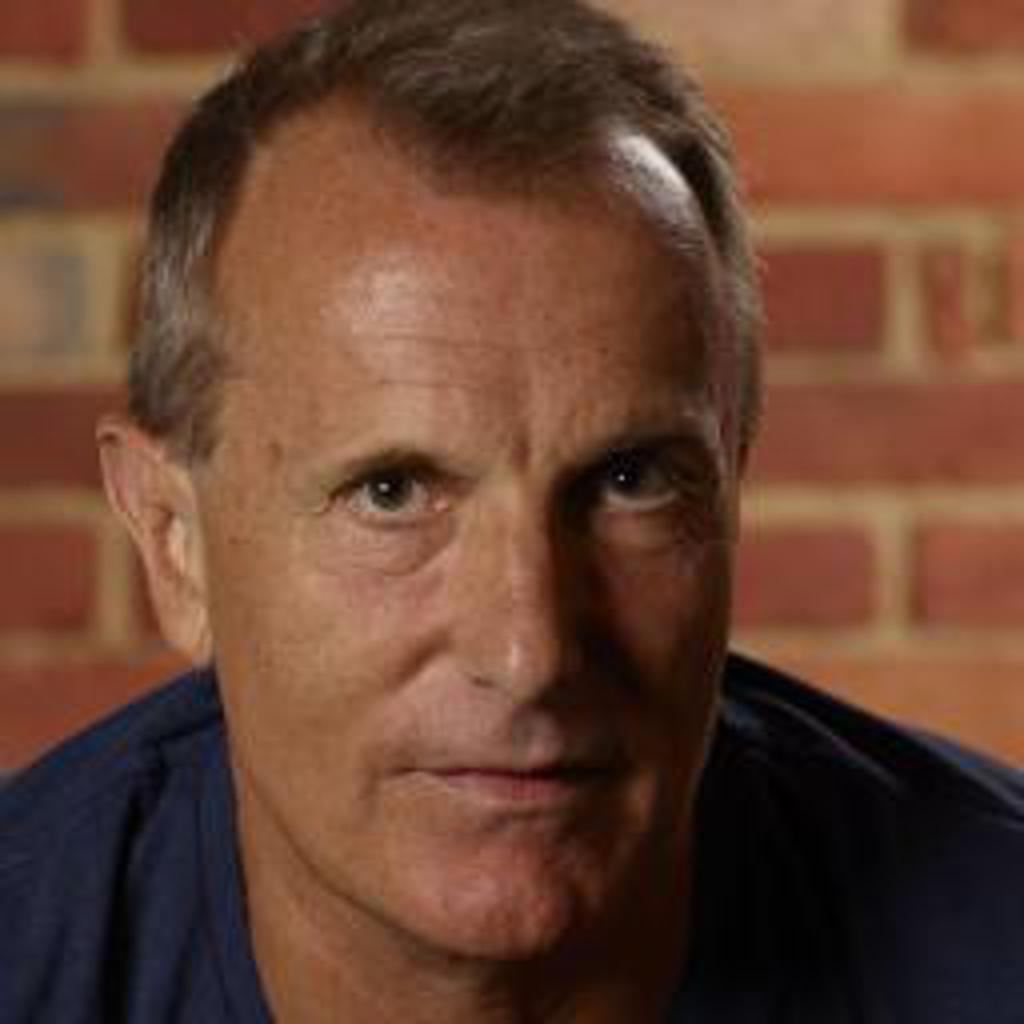}
        \includegraphics[width=1\linewidth]{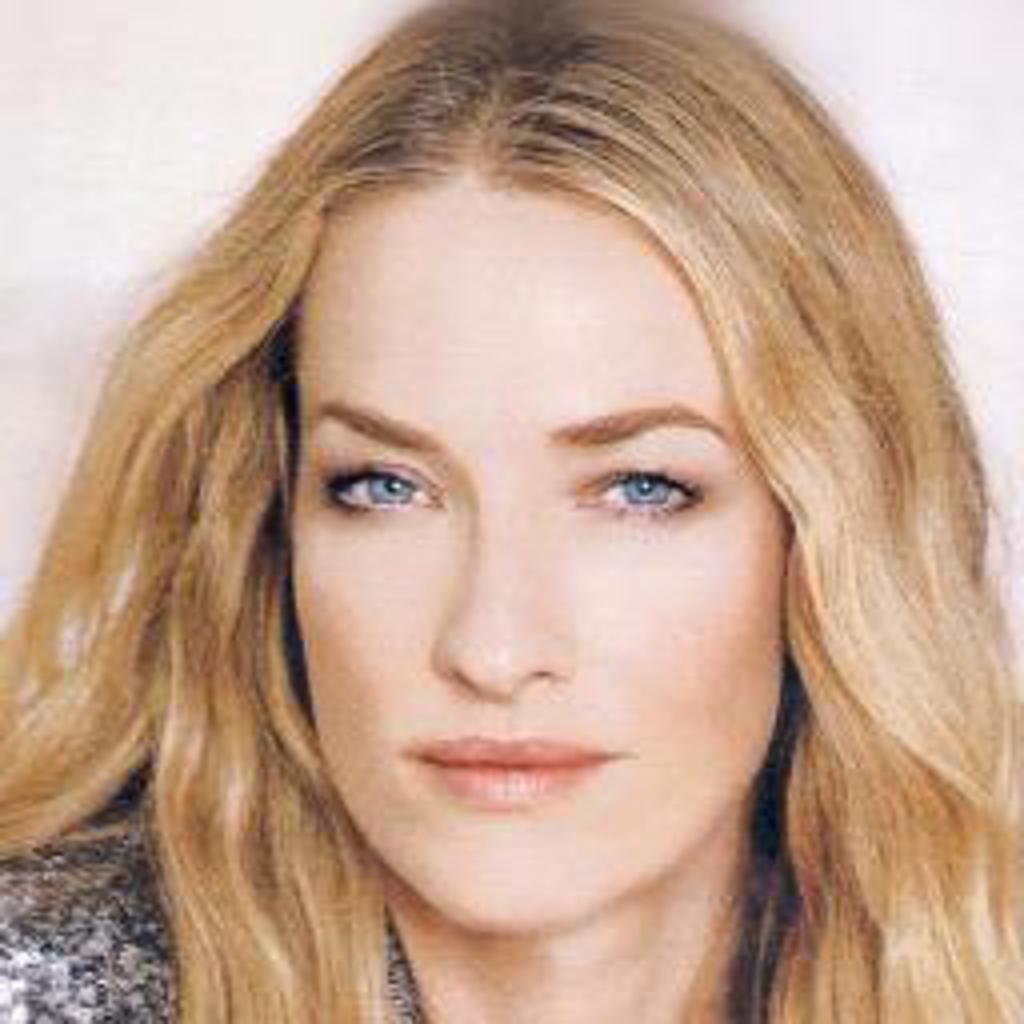}
        \includegraphics[width=1\linewidth]{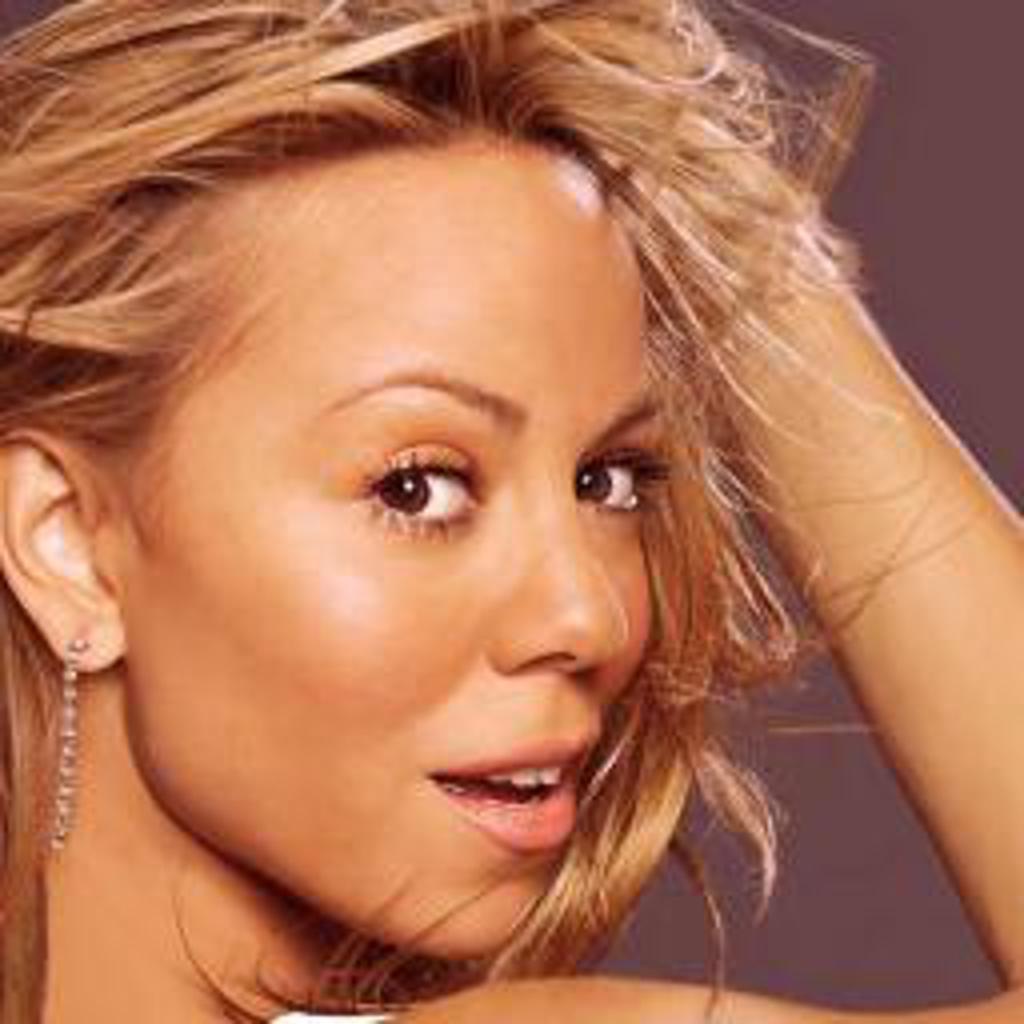}
        \end{minipage}
    \caption{Input}
    \label{fig:input3}
  \end{subfigure}
  \hspace{-0.015\linewidth}
  \centering
    \begin{subfigure}[t]{0.249\linewidth}
        \begin{minipage}[b]{1\linewidth}
        \includegraphics[width=1\linewidth]{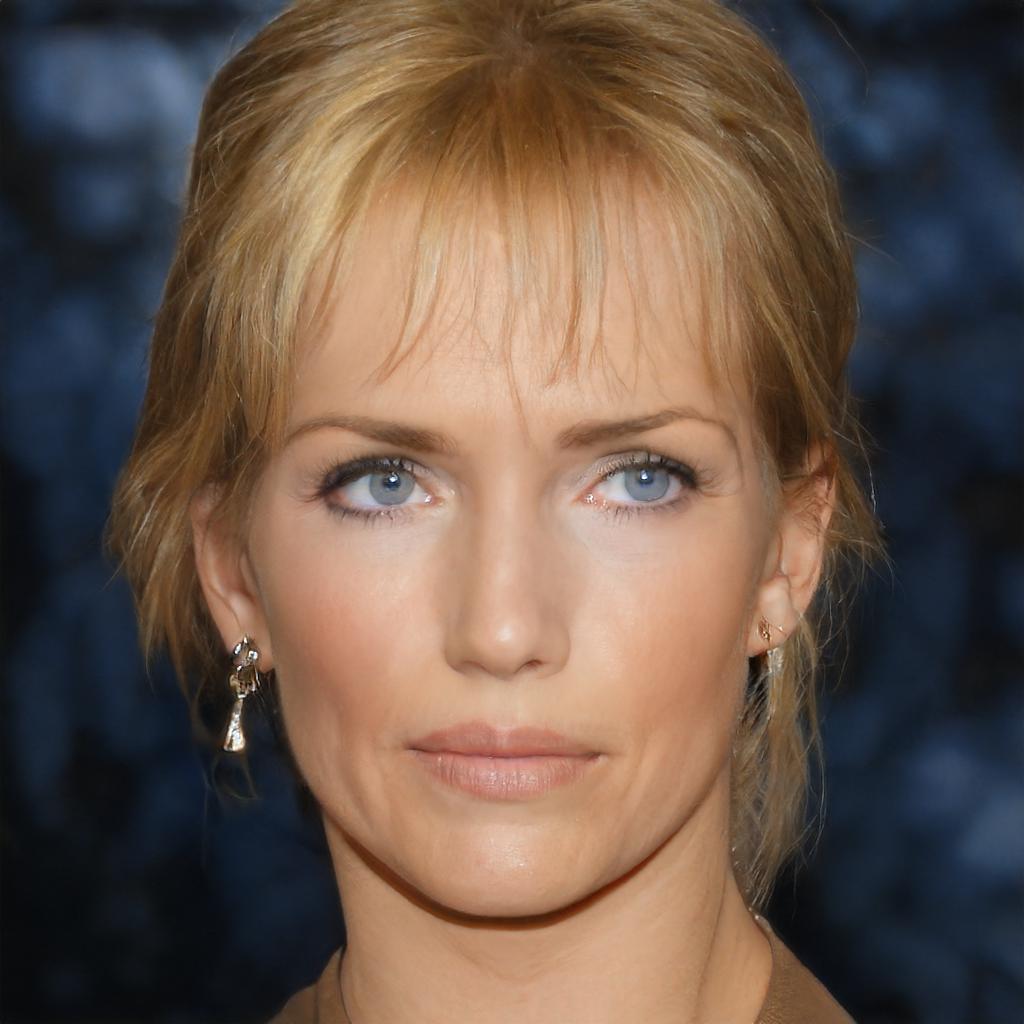}
        \includegraphics[width=1\linewidth]{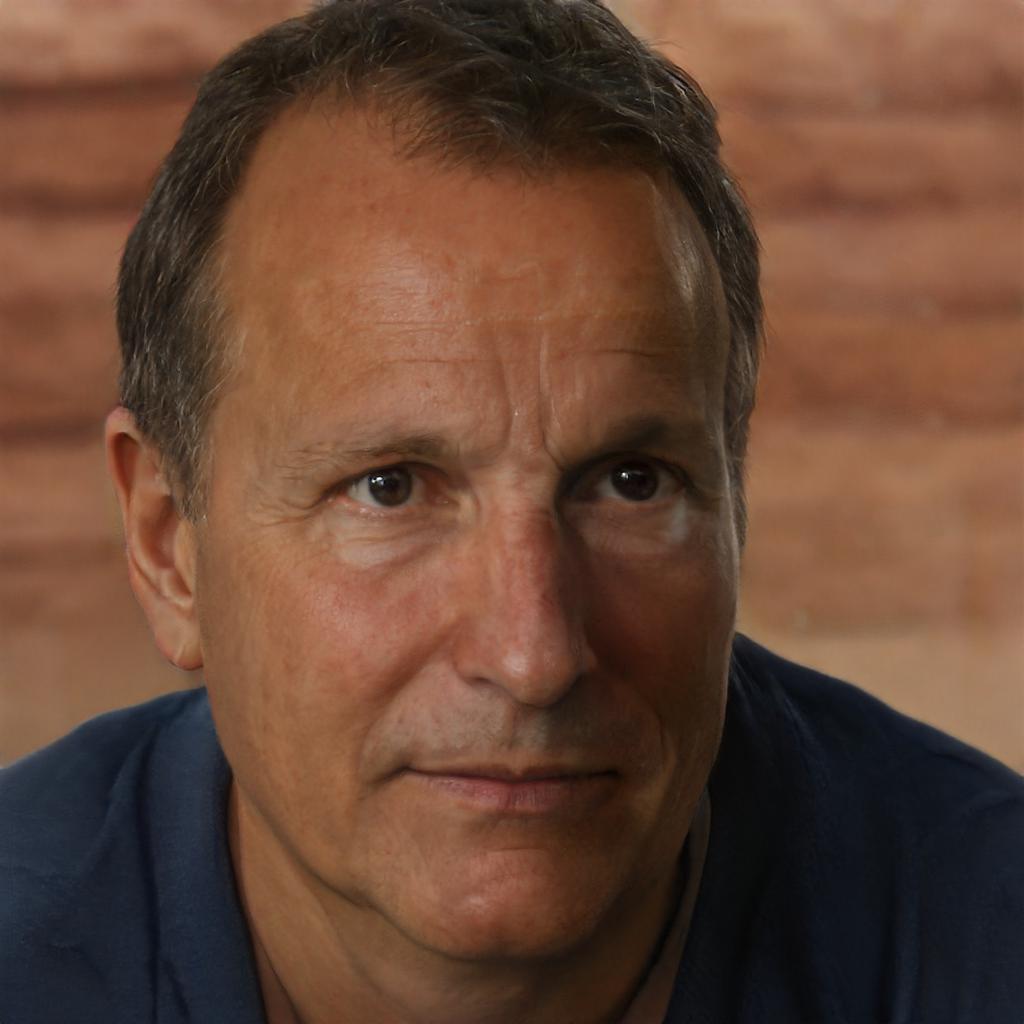}
        \includegraphics[width=1\linewidth]{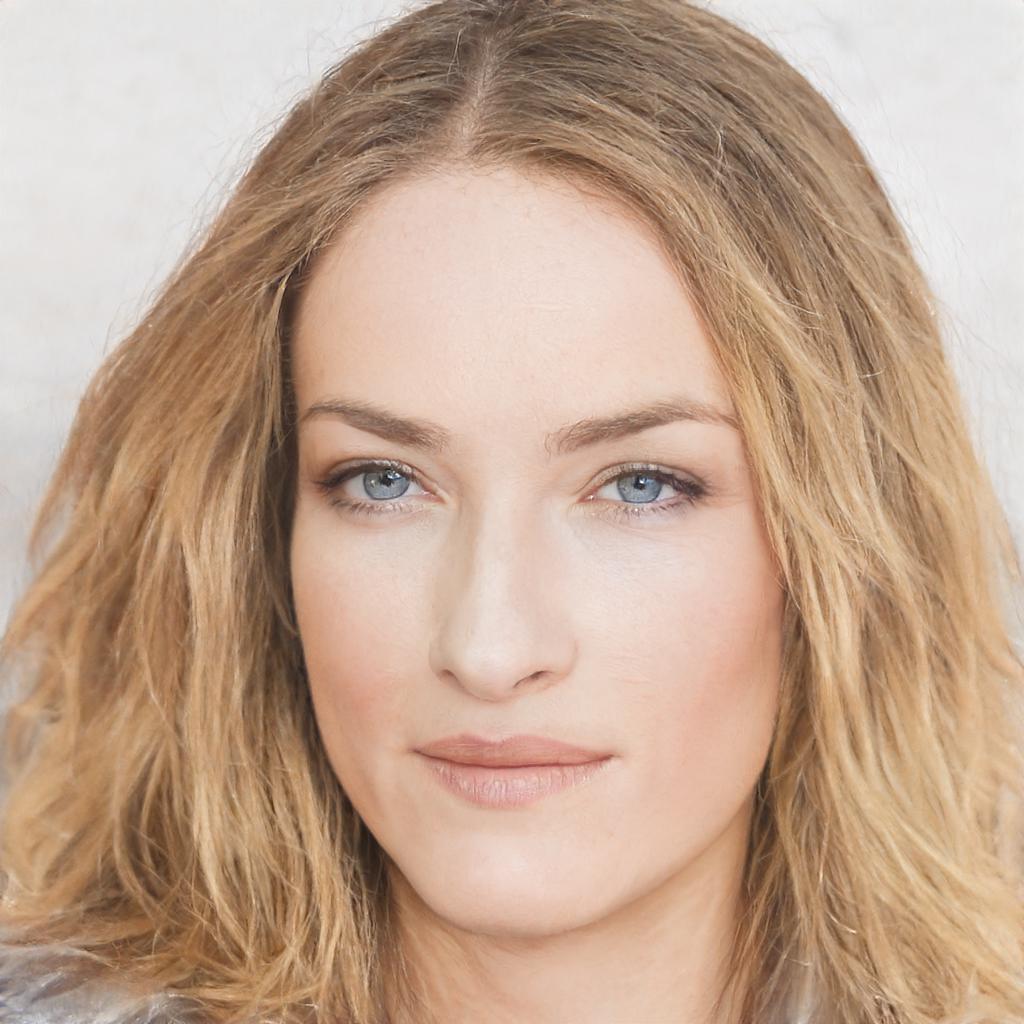}
        \includegraphics[width=1\linewidth]{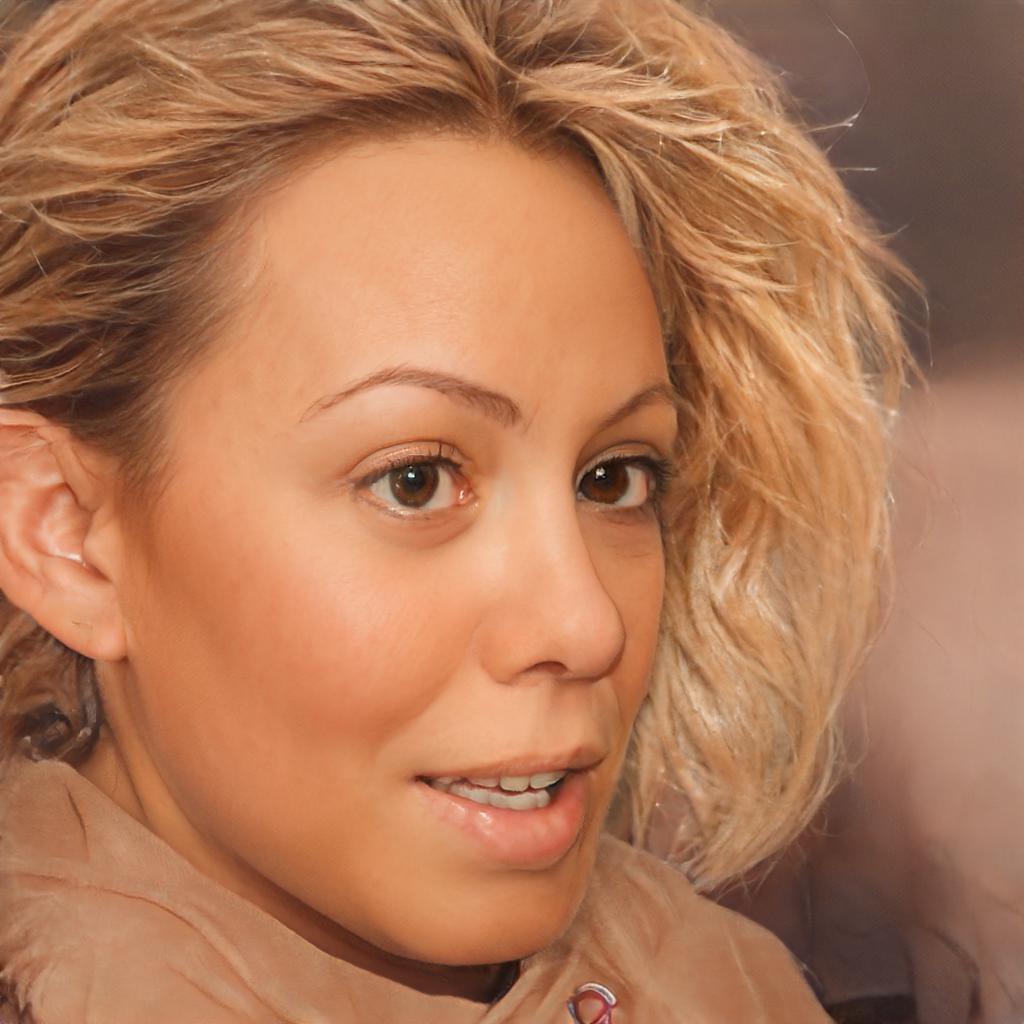}
        \end{minipage}
    \caption{e4e Inversion}
    \label{fig:e4e_inv2}
  \end{subfigure}
  \hspace{-0.021\linewidth}
  \centering
    \begin{subfigure}[t]{0.249\linewidth}
        \begin{minipage}[b]{1\linewidth}
        \includegraphics[width=1\linewidth]{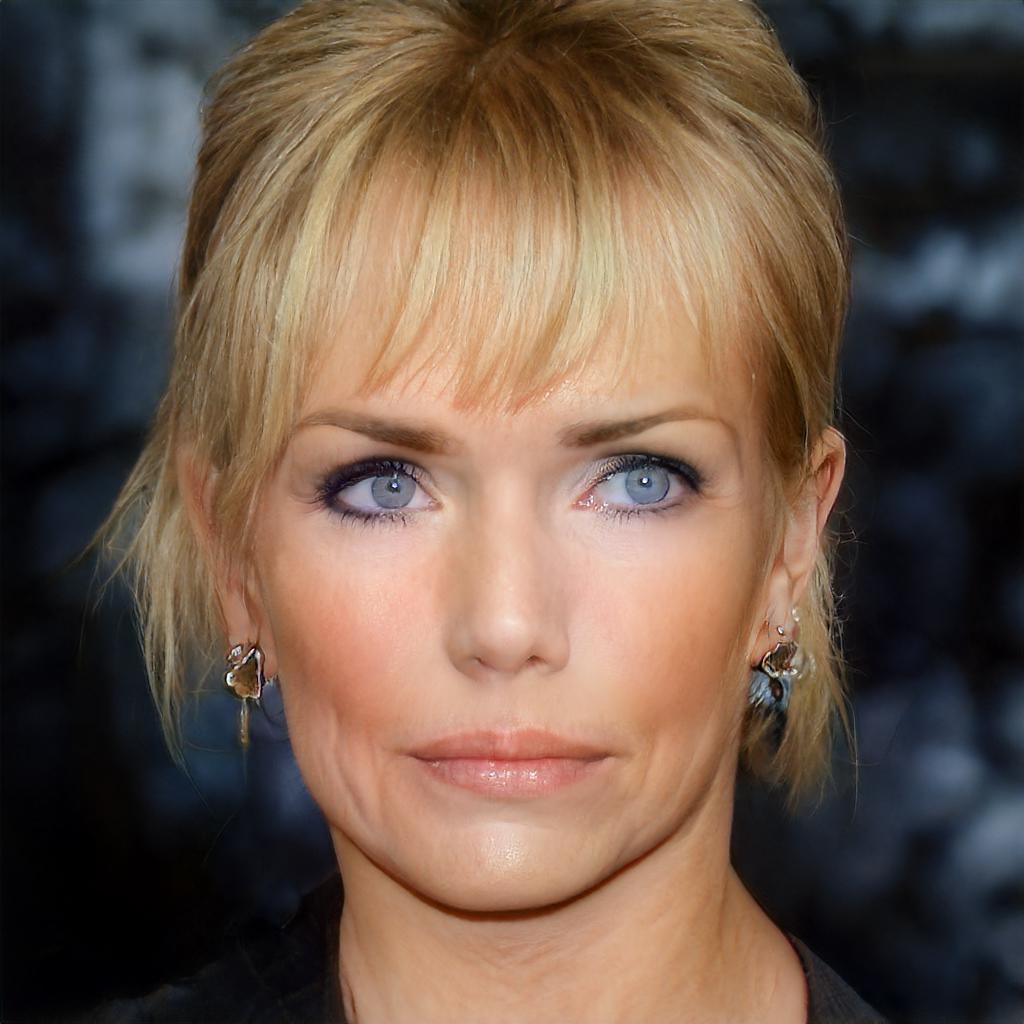}
        \includegraphics[width=1\linewidth]{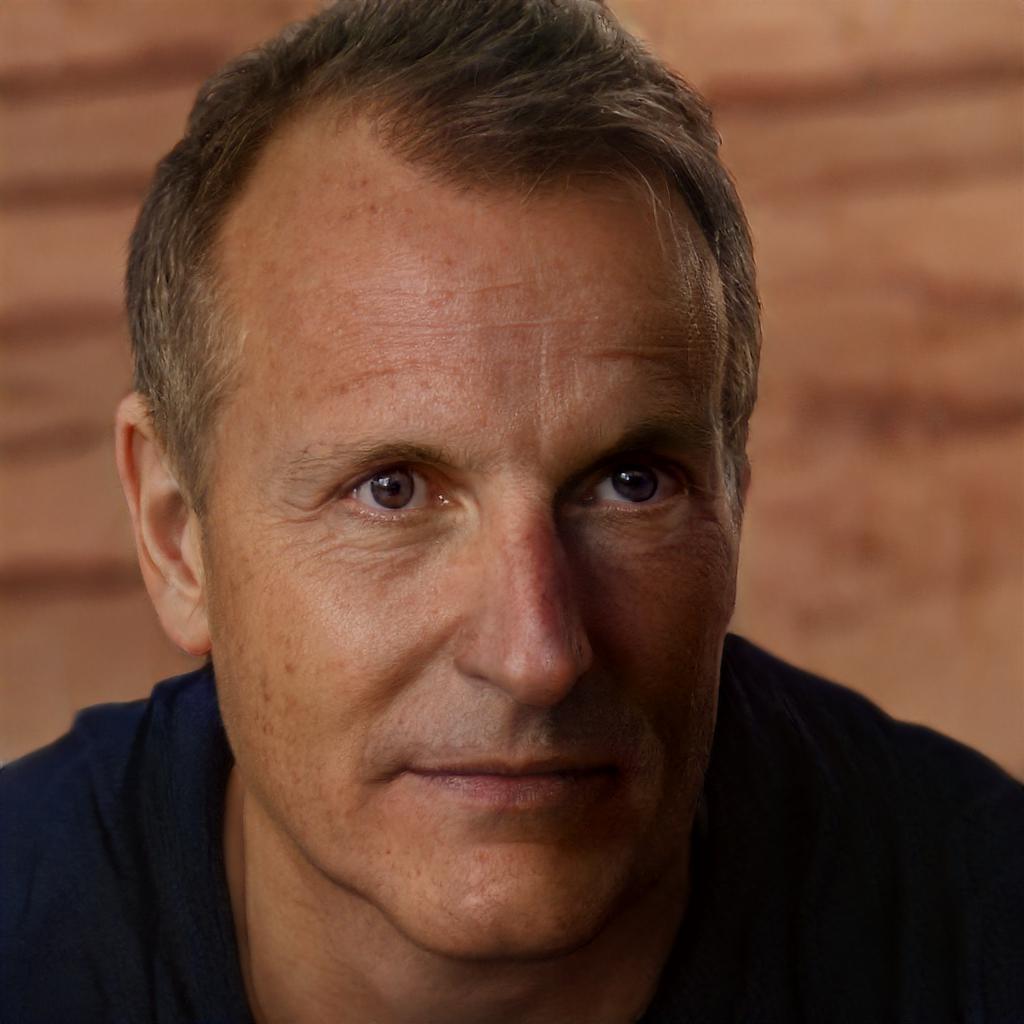}
        \includegraphics[width=1\linewidth]{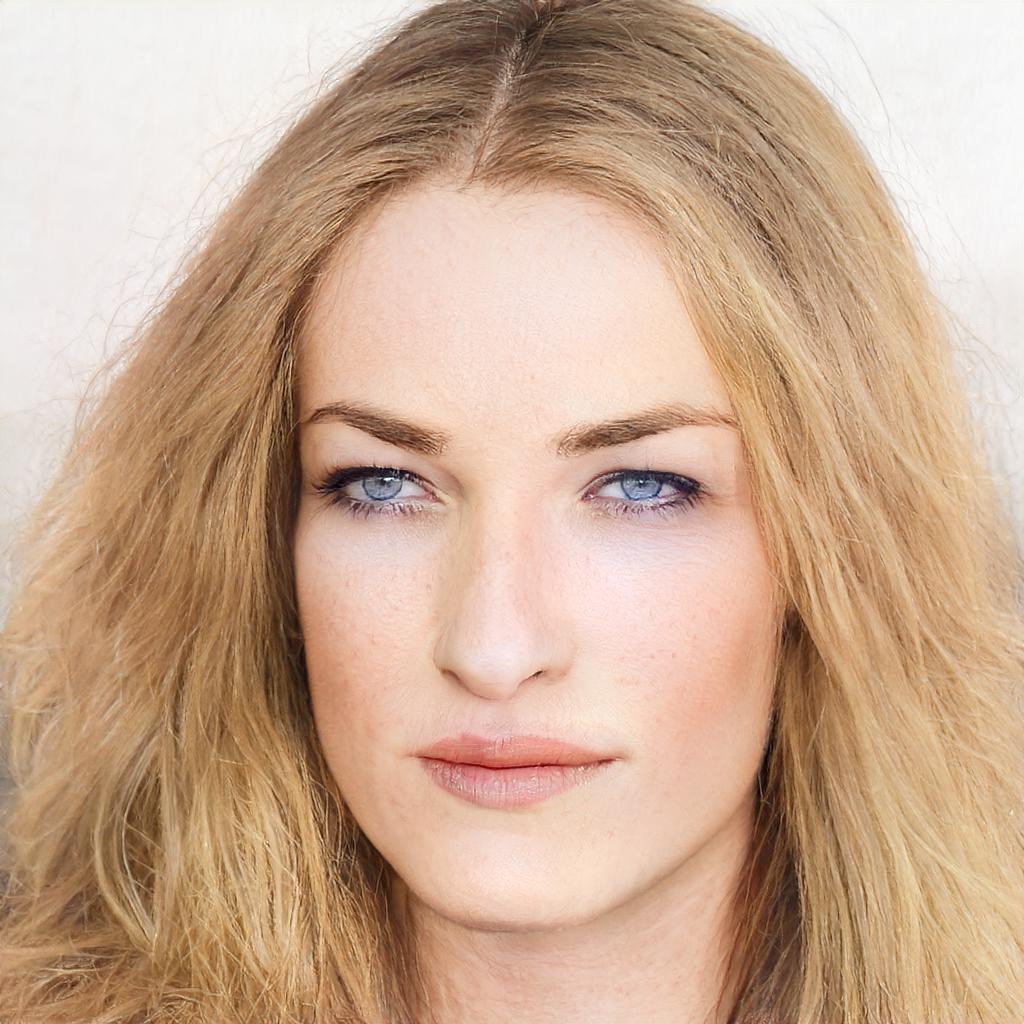}
        \includegraphics[width=1\linewidth]{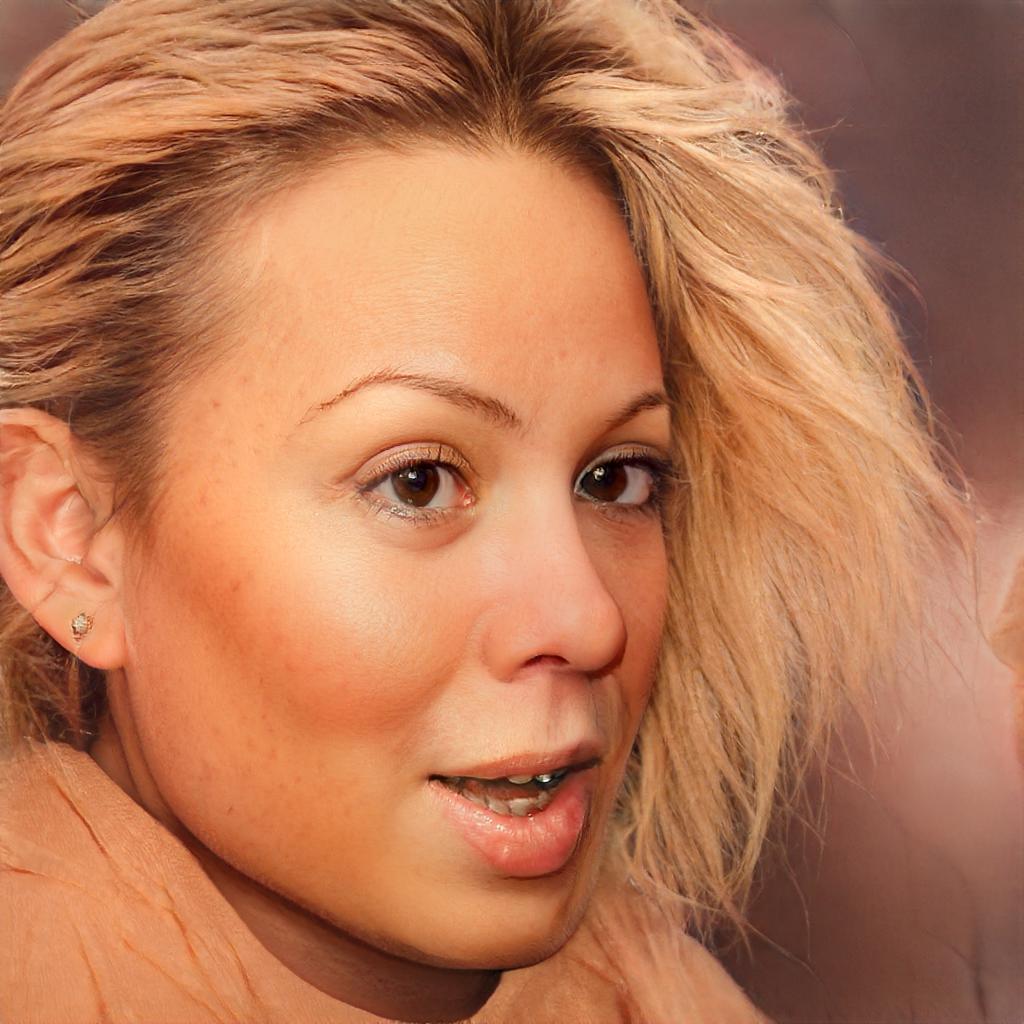}
        \end{minipage}
    \caption{ReStyle-e4e}
    \label{fig:restyle2}
  \end{subfigure}
  \hspace{-0.021\linewidth}
  \centering
    \begin{subfigure}[t]{0.249\linewidth}
        \begin{minipage}[b]{1\linewidth}
        \includegraphics[width=1\linewidth]{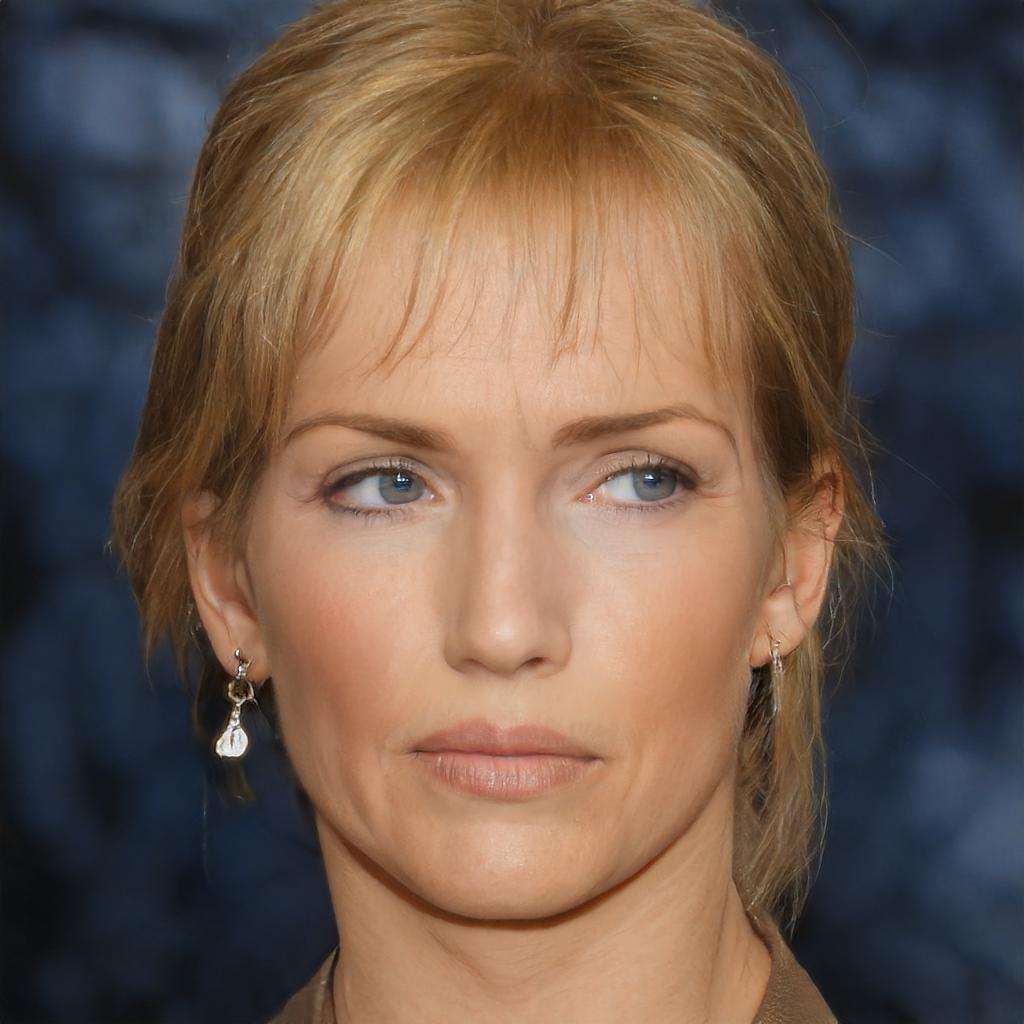}
        \includegraphics[width=1\linewidth]{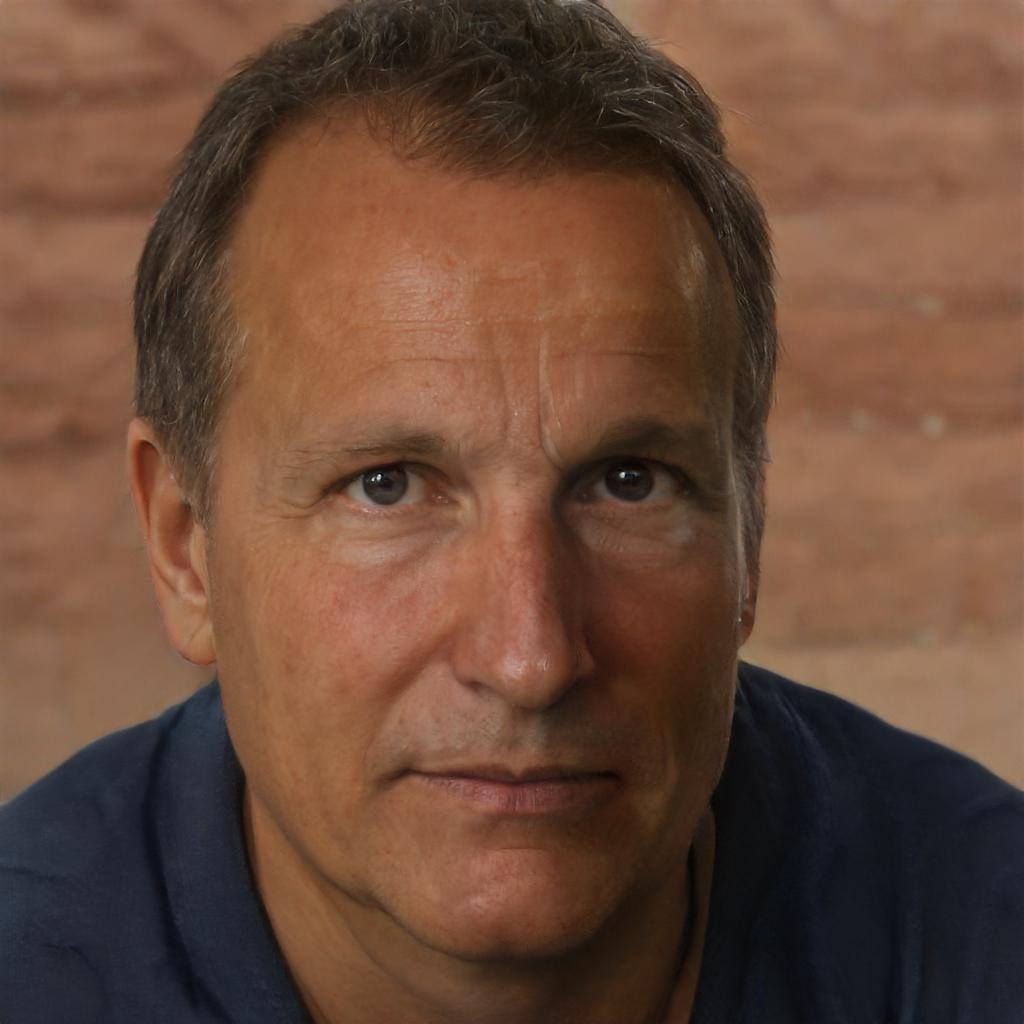}
        \includegraphics[width=1\linewidth]{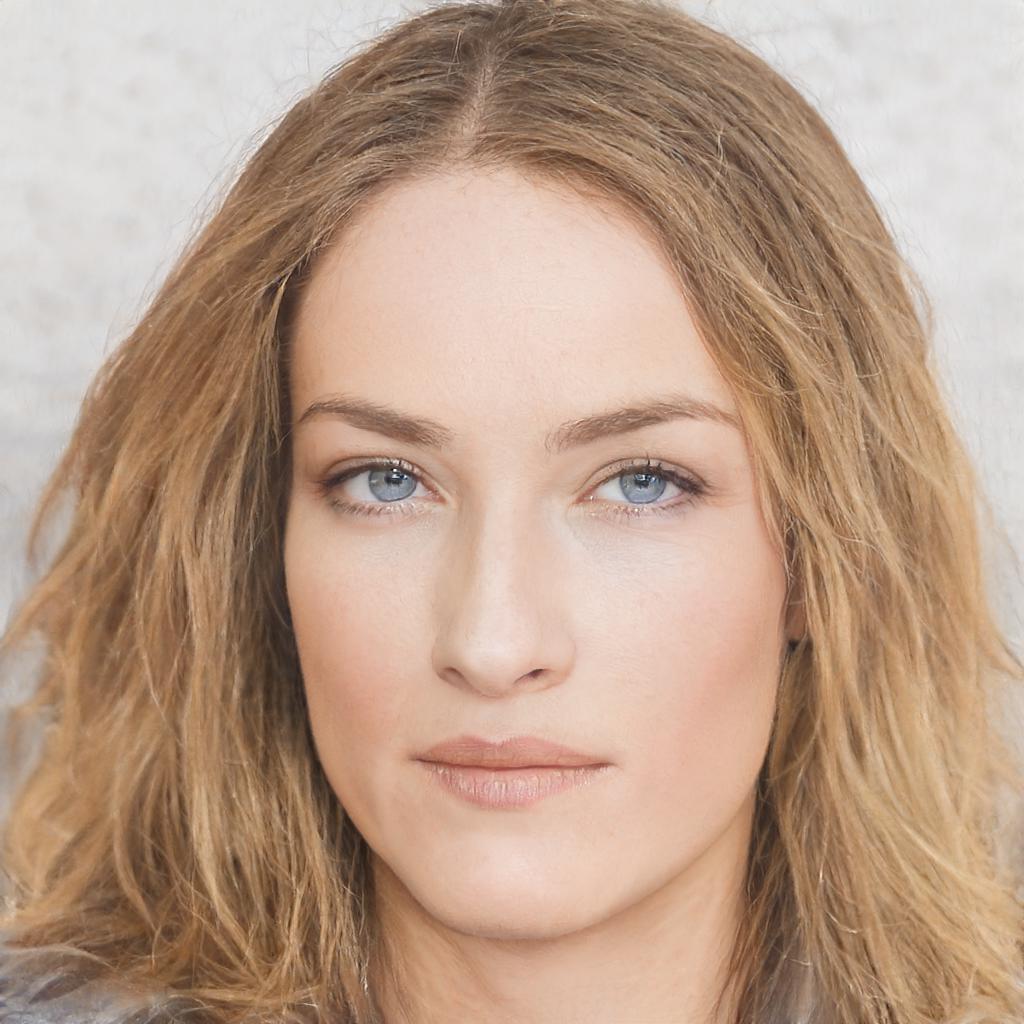}
        \includegraphics[width=1\linewidth]{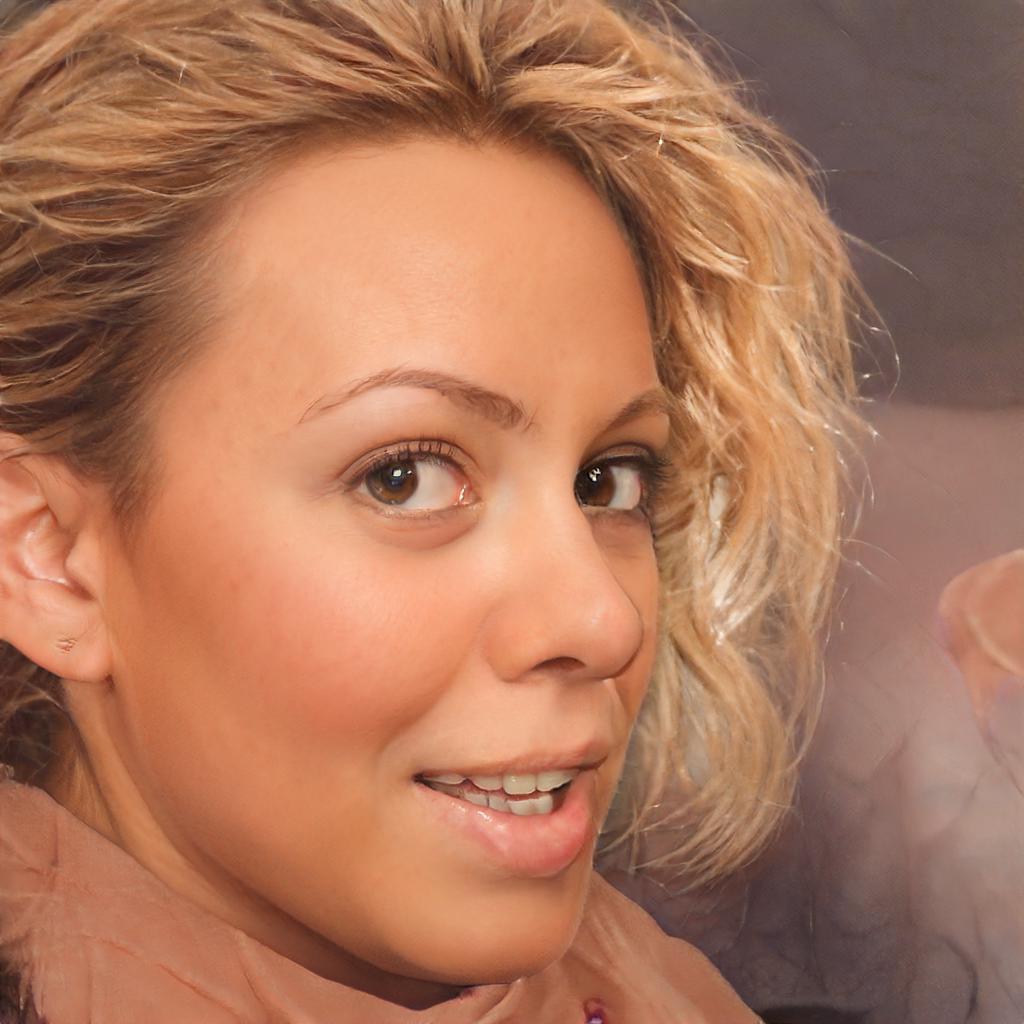}
        \end{minipage}
    \caption{ReDirTrans}
    \label{fig:redir}
  \end{subfigure}
  \caption{Gaze correction in CelebA-HQ by viewing the same image as both the input and target. 
  }
  \label{fig_rec}
\end{figure}
\subsection{Challenge in Predefined Feature Space}
\label{subsec:challeng_predefined}
One challenge for redirection tasks in predefined feature space comes from inconsistency between input and inverted images, mentioned in Sec. \ref{subsec:predefined_space}.  
We can observe that the existing gaze differences between input and inverted images in Fig. \ref{fig_red_comparison}. 
In some cases, the gaze directions are changed after GAN inversion, which means that the encoded latent codes do not necessarily keep the original gaze directions. 
Thus, instead of using provided gaze directions of input images during the training, we utilized estimated gaze directions from inverted results to correctly normalize the gaze and head pose to the canonical status. 
This process ensures correctness when further new directions are added, making the training process more consistent. 

\subsection{Gaze Correction}
\label{subsec:correction}
ReDirTrans can correct gaze directions of inverted results by viewing input images as the target ones. 
e4e guarantees high editability, which is at the cost of inversion performance \cite{tov2021designing}. 
Fig. \ref{fig_rec} shows several samples which failed to maintain input images' gaze directions even by the ReStyle encoder \cite{alaluf2021restyle}, which iteratively updates the latent codes given the differences between the input and inverted results. 
With ReDirTrans-GAN, we can successfully correct the wrong gaze based on inverted results from e4e.
\section{Conclusions}
\label{sec:conclusion}
We introduce ReDirTrans, a novel architecture working in either learnable or predefined latent space for high-accuracy redirection of gaze directions and head orientations. 
ReDirTrans projects input latent vectors into aimed-attribute pseudo labels and embeddings for redirection in an interpretable manner. 
Both the original and redirected embeddings of aimed attributes are deprojected to the initial latent space for modifying the input latent vectors by subtraction and addition. 
This pipeline ensures no compression loss to other facial attributes, including appearance information, which essentially reduces effects on the distribution of input latent vectors in initial latent space. 
Thus we successfully implemented ReDirTrans-GAN in the predefined feature space working with fixed StyleGAN to achieve redirection in high-resolution full-face images, either by assigned values or estimated conditions from target images while maintaining other facial attributes.  
The redirected samples with assigned conditions can be utilized as data augmentation for further improving learning-based gaze estimation performance. 
In future work, instead of a pure 2D solution, 3D data can be included for further improvements. 

{\small
\bibliographystyle{ieee_fullname}
\bibliography{egbib}
}

\end{document}